# Nellie: Automated organelle segmentation, tracking, and hierarchical feature extraction in 2D/3D live-cell microscopy

Austin E. Y. T. Lefebvre[1,*], Gabriel Sturm[1,2], Ting-Yu Lin[1], Emily Stoops[1], Magdalena Preciado López[1], Benjamin Kaufmann-Malaga[1], Kayley Hake[1]

[1]Calico Life Sciences LLC, South San Francisco, CA, 94080
[2]Department of Biochemistry and Biophysics, University of California San Francisco, San Francisco, CA, 94158

*Address correspondence to

Austin E. Y. T. Lefebvre, Ph.D.
Calico Life Sciences
1170 Veterans Blvd
South San Francisco, CA, 94080
Email: austin.e.lefebvre+nellie@gmail.com

**Abstract**

The analysis of dynamic organelles remains a formidable challenge, though key to understanding biological processes. We introduce Nellie, an automated and unbiased user-friendly pipeline for segmentation, tracking, and feature extraction of diverse intracellular structures. Nellie adapts to image metadata, eliminating user input. Nellie's preprocessing pipeline enhances structural contrast on multiple intracellular scales allowing for robust hierarchical segmentation of sub-organellar regions. Internal motion capture markers are generated and tracked via a radius-adaptive pattern matching scheme, and used as guides for sub-voxel flow interpolation. Nellie extracts a plethora of features at multiple hierarchical levels for deep and customizable analysis. Nellie features a point-and-click Napari-based GUI that allows for code-free operation and visualization, while its modular open-source codebase invites extension by experienced users. We demonstrate Nellie's wide variety of use cases with three examples: unmixing multiple organelles from a single channel using feature-based classification, training an unsupervised graph autoencoder on mitochondrial multi-mesh graphs to quantify latent space embedding changes following ionomycin treatment, and performing in-depth characterization and comparison of endoplasmic reticulum networks across different cell types and temporal frames.

**Main**

The complex weave and elaborate dance of organelles lies at the center of cellular physiology and pathology. For example, the alterations and balance of mitochondrial dynamics coregulate its turnover, quality control, mtDNA organization, and bioenergetic output[1–5]. Importantly, organelles can also form contact sites between one another, allowing for the exchange of metabolites, ions, and proteins, and promote autophagic turnover, where dysfunctions in any of these have been correlated with aging and other various diseases[6–9].

The topic of organelles as drivers of physiological dysfunction is clearly important; however, the dynamic morphology and motility of these organelles, coupled with limitations inherent to microscopy such as acquisition speed, the diffraction limit, and tradeoffs between signal and phototoxicity, pose significant challenges in extracting this information. This results in manually involved or organelle-specific pipelines that do not generalize well to broader datasets. Consequently, there is a pressing need for a widely accessible analytical tool capable of providing detailed extraction of spatial and temporal features at multiple organellar scales, but that remains independent of the tool's user or the organelle in question.

Many tools exist for intracellular structural segmentation and tracking[10–28]. However, the pipelines either rely heavily on manual or semi-automated techniques, which are time-consuming, prone to subjective bias, and often infeasible for large and/or spatially 3D datasets. Automated methods, which offer improvements in speed and objectivity, frequently struggle with the complexity and variability inherent in biological imaging data. Common issues of existing tools include the inability to effectively handle multi-scale structures present between or even within datasets, insufficient segmentation accuracy for dim or small objects, and limitations in tracking algorithms, particularly in dense and dynamically complex cellular environments. Additionally, most tools rely on the assumption that an organelle is a single and temporally consistent entity with only occasional and specifically defined merging or splitting events, which limits the quantification of these phenomena to arbitrary metrics. Additionally, deep-learning (DL) based microscopy methods are rapidly advancing, with state of the art segmentation, tracking, and feature extraction tools constantly being released[29]. Most of these tools, however, either lack 3D functionality (mainly due to the complexity of manual 3D annotation leading to the lack of ground truth data for the training or fine-tuning of models), organelle generalizability, or are specifically tailored to electron microscopy organelle datasets[28,30–44]. Furthermore, these DL models inherently contain 'black box' predictions with unexplainable results that are generally hard to interpret[45]. Thus, there remains a significant gap in the development of a comprehensive, automated, and organelle-agnostic pipeline capable of efficiently and accurately processing large-scale and multi-dimensional fluorescence microscopy datasets.

In this paper, we introduce Nellie, short for organellometer, a novel easy-to-use GUI-based, point-and-click image analysis pipeline designed specifically to address these challenges. By incorporating multi-scale structure-enhancing preprocessing methods, Nellie is able to segment and hierarchically divide organelles into logical subcomponents. These subcomponents are interrogated to produce motion capture markers that are compared via local, variable-range feature and pattern matching to create linkages between adjacent frames. These linkages act as beacons for novel temporal interpolation algorithms to provide sub-voxel tracking capabilities. We incorporate and introduce a multitude of both standard and advanced quantification techniques to extract a hierarchical pool of descriptive multi-level spatial and temporal features to choose from.

The whole pipeline can be run on a CPU or accelerated on a GPU, either of which computationally outperforms current SOTA organelle segmentation and tracking tools (Supplementary Note 1, Extended Data Fig. 1). Running the whole pipeline, and analysis of its outputs is made easily accessible via a point-and-click Napari GUI[46]. The Napari GUI prompts the user to select their file or folder to analyze (Fig. 1a). A metadata validation module ensures Nellie has detected proper dimension order and resolutions, and allows the user to correct these parameters if automatic detection has failed (Fig. 1b). If the data contains multiple temporal or channel dimensions, the user can specify which slice(s) to run through Nellie (Fig. 1b). The GUI then prompts the user to run the pipeline, and allows for visualization of intermediate images and tracks (Fig. 1c). Finally, the GUI allows for quick visualization of extracted features (Fig. 1d). Notably, Nellie allows the user to run single frame datasets (without a temporal dimension) for morphology-only analysis, or multi frame datasets for additional motility quantification, and works for both 2D (single-plane) datasets and 3D (volumetric) datasets. We also allow Nellie to find compatible plugins via the 'nellie.plugins' Python entrypoint to allow scientists to integrate their tailored code and pipelines into Nellie's ecosystem.

To showcase the broad range of potential uses that Nellie and its extracted features offer, we present three use cases that we hope will serve to inspire more advanced developments on Nellie's extracted features. First, we show how one can use Nellie's outputs to unmix multiple organelle types from a single channel of a fluorescence timelapse. Second, using Nellie's outputs we develop a novel multi-mesh approach to organelle graph construction, inspired by DeepMind's recent GraphCast paper[47]. We use this multi-mesh to train an unsupervised graph autoencoder, and use the model to compare mitochondrial networks across a complex feature-space. Third, we demonstrate Nellie's capabilities in characterizing and comparing endoplasmic reticulum (ER) networks between different cell types and temporal frames, showcasing its ability to perform in-depth analyses of organelle morphology, motility, and network topology. This case study highlights Nellie's power in extracting meaningful differences between cell types while maintaining consistency within temporal sequences. We hope that this work not only provides a valuable tool for cellular biologists but that it also sets a new standard for automated image analysis as a whole, enabling researchers to gain deeper insights into the complex world of intracellular organization and dynamics.

## Results

**Multi-scale adaptive filters enhances structural features of organelles**

Laser and dye properties can cause the signal-to-noise of organelles to fluctuate widely both between and within datasets. In the preprocessing stage of our pipeline, we account for these fluctuations by implementing a modified version of a multi-scale Frangi filter to enhance the inherent structural contrast of organelles, allowing for segmentation based on local structure rather than fluorescence intensity (Fig. 1e, Supplementary Data Fig. 1)[48]. Our filter is empirically optimized for structures in the size range of typical organelles, and automatically adjusts the filter's effective range based on voxel dimensions to adapt to various magnifications and anisotropies. We modify the traditional Frangi filter to make it generalizable to both tubular and non-tubular structures, which we further make more broadly generalizable via an adaptive and fully automated parameter calculation on a scale-by-scale basis to enhance structures of multiple sizes (Supplementary Data Fig. 2). Our pipeline contrasts with the current state-of-the-art (SOTA) of traditional intracellular segmentation pipelines, which are not adaptive to intrinsic image metadata, and if included, use the same filter parameters across all scales of structural enhancement, limiting the robustness of the filter and subsequent segmentation of variable-scale objects[12,14]. Once filtered, this preprocessed image is subject to standard thresholding techniques for semantic segmentation. Additional details on Nellie's preprocessing pipeline can be found in Supplementary Notes 2 and 3. In this regard, we find that Nellie surpasses current SOTA methods in a diverse range of segmentation tasks across simulated datasets ranging from small round objects to large round objects to small tubular objects to large tubular objects and everything in between across a variety of noise levels (Supplementary Note 4, Extended Data Fig. 2). Furthermore, we show that Nellie generalizes well to various datasets, including those from different microscopes and across various organelles, as compared to custom pre-trained Swin UNETR deep-learning models (Extended Data Fig. 3, Supplementary Note 5)[49].

**Hierarchical deconstruction allows for multi-level organelle instance segmentation**

Before answering how an organelle changes, we must first ask ourselves how an organelle should be represented. Individual organelles are rarely ever individual organelles at all, but rather belong to a complex and continuously evolving organellar landscape. In this regard, it is useful to instead think of and represent organelles as a hierarchical collection of objects at independent frames; the organellar landscape of a cell at a single frame is made up of spatially disconnected organelles, which is in turn made up of numerous subcompartments, which can be broken down into nodes, voxels, or even sub-voxel regions. To capture this representation, Nellie performs several steps to deconstruct our organellar landscape. First, we employ a Minotri threshold on the preprocessed image generating a semantic segmentation mask - our organellar landscape (Supplementary Note 3). We then perform a simple connected-components based labeling scheme to generate instance segmentations of individual objects - our spatially disconnected organelles (Fig. 1f). We then skeletonize these segmentations, and use the skeleton to identify network junction nodes (branching points) within individual components, allowing us to deconstruct the organelle network into individually labeled branches - our organelle subcompartments (Fig. 1g). We further break down these subcompartments into individual skeleton nodes, which hold properties of their radius-dependent surrounding voxels. To maintain a continuous linkage across all levels of our hierarchy, we can generate adjacency maps by iteratively reassigning the semantic segmentation mask voxels via a k-dimensional (k-d) tree of the skeleton nodes and their branch labels, enabling the graph-like traversal of our organellar landscape between any hierarchical level[50]. Additional details on Nellie's segmentation pipeline can be found in Supplementary Note 6.

**Motion capture markers are generated for downstream tracking**

The consistency of object-based segmentations and skeleton networks are notoriously finicky between timepoints, which causes linkage problems when using center-of-mass or skeleton-based tracking approaches. To avoid these problems, we instead generate motion capture (mocap) markers within our organelles, completely independently of our labels, and use these as a basis for linking variable-radius regions of our image across timeframes (Fig. 1h). These mocap markers do not intrinsically contain any biological significance, but instead act as guideposts for downstream flow interpolation. This is in line with representing the organellar landscape as a dynamic entity, rather than tracking specific instance segmentations. Additional details on Nellie's mocap marker generation pipeline can be found in Supplementary Notes 7 and 8. Using these mocap markers to generate tracks, we find that Nellie generalizes well to various datasets, including those from different microscopes and across various organelles, and surpasses current SOTA methods in a diverse range of tracking tasks in simulated datasets (Extended Data Fig. 3, Supplementary Note 9, Extended Data Fig. 4).

**Features for each motion capture marker are gathered via variable-range queries**

To temporally link these mocap markers, a comprehensive feature vector is constructed to encapsulate critical aspects of the organelles' local characteristics and dynamics. At each mocap marker, the distance transformed value, representing the organelle's radius at that point, is multiplied by two, providing the dimensions for the bounding box of each marker (Fig. 2a). Within these bounding boxes, the mean and variance are computed for both the raw image and the preprocessed image. These statistics are collectively termed as the 'stats vector' (Fig. 2b). Furthermore, the first six 2D Hu moment invariants of the raw intensity image and the preprocessed image within the mocap marker's bounding box regions are computed to generate translation, scale, and rotation invariant comparison metrics[51]. In 3D, these 2D Hu moment invariants are calculated for xy, xz, and yz projections of the 3D bounding box region, resulting in what we term as the 'Hu vectors' (Fig. 2c). Finally, to link markers between adjacent frames, a multi-faceted cost matrix is constructed (Fig. 2d). Additional details on Nellie's feature-based cost matrix mocap marker linkage pipeline can be found in Supplementary Note 10.

**Motion capture markers are tracked and used as guides for sub-voxel flow interpolation**

Rather than solving the linear assignment problem by minimizing the global cost of mocap marker linkages, markers from frame t are simply assigned to their best-matched markers in frame t+1 and vice versa, allowing for 1-to-1 matching, 1-to-N matching, and N-to-1 matching (Fig. 2e). Each assignment results in a vector pointing from a marker at time t to another marker at t+1, and vice versa, with an associated cost to that linkage. It is important to emphasize that these motion capture markers and their linkages do not represent the final organelle tracks but rather serve as the beacons that point in the direction of local motion to inform the subsequent sub-voxel flow interpolations that serve as tracks. Once markers are assigned, any arbitrary coordinate or coordinates of interest (CoI) can have its flow vector interpolated, meaning the motion of an entire organellar object's collection of voxels, or a single branch's collection of voxels, or even a single voxel or sub-voxel point within an organelle can be interrogated.

For interpolating flow vectors of CoIs forwards in time from frame t, a k-d tree is first constructed using the coordinates of markers at frame t, facilitating efficient nearest-neighbor searches. All nearby marker coordinates within the maximum travel distance from the CoI are then identified, and the distances between these detected markers and the CoI are calculated by querying the k-d tree. Each detected marker's flow vector for interpolation of the CoI's flow vector is weighted based on a preference for vectors that are closer and have a lower assignment cost value, indicating a better match during mocap marker linkage. The final interpolated vector from frame t to t+1 is the sum of these weighted vectors (Fig. 2f). A similar process for markers at frame t-1 to frame t is performed for interpolation of flow vectors backwards in time. This interpolation process is efficiently executed in parallel for all CoIs, resulting in a list of flow vectors that represent the interpolated motion of the CoIs to the adjacent frame (Fig. 1i, 2g).

Importantly, these flow vectors can be used to match voxels between timepoints, meaning one can determine the fate of each individual voxel coming from an organelle, or any part of its segmentation hierarchy, across all frames of a timelapse (Extended Data Fig. 5, Supplementary Note 11). Additionally, we note that this novel tracking and interpolation method is not limited to organelles or fluorescence microscopy and can easily be adapted to other types of images, including those segmented via, for example, Meta's Segment Anything Model (Extended Data Fig. 6)[52].

**Organellar features are calculated at and between multiple hierarchical levels**

Though segmentation and tracking of organelles is useful for data exploration and visualization, objective interpretation only becomes possible when quantifiable features are available. To this end, Nellie allows for the calculation and export of a plethora of features specific to each level of the segmentation hierarchy, as well as the statistical investigation of inter-level features, such as an organelle's mean branch feature values, or a branch's nodes' mean feature values, etc.

Nellie begins with the calculation of features at the single-voxel level, such as fluorescence and structural intensity values from the raw and preprocessed images, and uses flow interpolation of nearby mocap markers to extract motility metrics at each voxels' coordinates (Fig. 1j, 3a). Next, at the single-node level, which are centered around individual skeleton voxels, Nellie calculates various local morphology and voxel flow patterns (Fig. 1k, 3b). At the single-branch level, Nellie calculates both skeleton-specific and branch-specific morphology features (Fig. 1l, 3c, Supplementary Note 12). Finally, at the single-organelle level, Nellie calculates morphology features of the whole organelle (Fig. 1m, 3d). For all of these metrics, Nellie also outputs a final aggregate dataset, allowing for interpretation of image-wide averages, maximum values, minimum values, variability, etc of the entire organellar landscape (Fig. 3e). Nellie also calculates statistics of aggregate values for features calculated at lower hierarchical levels. For example, a mean aggregation of linear velocities coming from voxels within a node or organelle can be calculated for each node and organelle as a whole (Fig. 3f). A variety of aggregation metrics can be calculated, from those as convoluted as a branch's nodes' mean linear velocity vector magnitude variability, to those as simple as summing the lengths of all the branches within an organelle (Fig. 3g). The values for each of these level-specific and aggregated features can additionally be overlaid as a colormap for each voxel, node, branch, or organelle label, and viewed with the features' corresponding histogram in Nellie's Napari plugin, allowing for easy data exploration and visualization. Additional details on Nellie's hierarchical feature extraction pipeline can be found in Supplementary Note 12, and a full list of exportable features can be found in Supplementary Table 1.

**Case Study 1: Nellie's extensive feature quantification allows for unmixing of multiple organelle types in a single channel**

In cellular microscopy, imaging more than a few organelle types in live cells within a single time series is a formidable challenge, often constrained by the limited availability of imaging channels, dye or fluorophore specificity, and the necessity to minimize phototoxicity and photobleaching. We introduce an innovative methodology that synergizes the advanced feature extraction capabilities of our pipeline, Nellie, with machine learning classification techniques. This approach enables the *post-hoc* demixing of organelles in single-channel images, effectively addressing a critical bottleneck in cellular imaging.

Utilizing multi-channel timelapse fluorescence microscopy, we separately captured images of Golgi apparatus and mitochondria (Fig. 4a). These independent channels were processed through Nellie to extract organelle-specific features. For validation, we generated combined organelle images–comprising both mitochondria and Golgi in a single channel–from the maximum intensity channel-projections of the two channels, and ran this projection image through Nellie to generate instance segmentations and extract their corresponding spatial and temporal features (Fig. 4b). To establish ground truth for the combined channel timelapse, we quantified the overlap of mask voxels between the multi-channel and the combined-channel masks. Organelles were labeled as mitochondria or Golgi based on the predominant overlapping channel.

Employing the multi-channel features from Nellie, we developed three Random Forest classifier models, each trained on either motility features alone, morphology features alone, or a combination of both[53]. The selection of features was grounded both empirically and through the analysis of features showing the most significant fold differences between mitochondria and Golgi (Fig. 4c). We intentionally exclude voxel intensity metrics in the models as intensity is not inherently dependent on structure. The models' efficacies were tested by comparing the models' predicted organelle types to the ground truth organelle types. We captured timelapses of 11 cells and used each cell in an 11-fold leave-one-out cross-validation, leaving one timelapse out of the training set for testing for each cross-validation to evaluate each model, reflecting a realistic experimental scenario with limited sample sizes. Our validation results were promising, with all models surpassing the 0.50 threshold indicative of random guessing in the area under the curve (AUC) of the receiver operating characteristic curve (ROC). The combined model achieved an average AUC of 0.80, followed by 0.79 for morphology-only features, and 0.66 for motility-only features (Fig. 4d). Additional performance metrics also follow similar trends (Extended Data Fig. 7). Moreover, the model allowed us to identify the most impactful features contributing to its performance. These features include a higher aspect ratio and length in mitochondria, owing to its more networked morphology, and a larger radius and solidity in golgi, due to its more spherical shape (Fig. 4e).

This study demonstrates the potential of Nellie in advancing cellular organelle microscopy, especially under conventional imaging constraints. Leveraging standard random forest classification models from features extracted by Nellie, our method adeptly distinguishes complex biological structures in single-channel images, even with a limited dataset size, creating a significantly useful tool in the field of cellular imaging.

**Case Study 2: Nellie's hierarchical organelle segmentation allows for the learning of comparable variable-range latent space representations of organelle graphs**

In cellular microscopy, the intricate task of analyzing organelle networks demands innovative approaches, particularly when examining dynamic alterations in organelle organization. Here, we introduce a novel method that employs graph-based latent space representations to interpret changes in organellar networks. By transforming skeletonized networks of organelle segmentation masks into graph structures and utilizing a attention-based graph autoencoder to transform Nellie's extensive feature outputs into a comparable representation, we decode subtle shifts in organellar arrangements.

We first define the nodes of our graph as skeleton voxels underlying the organelle segmentation masks, with each node encapsulating features of the adjacent organelle voxels. Utilizing the distance-transformed image, we determine the radius representative of each node. The features of surrounding voxels within this radius and the semantic segmentation mask—including raw intensity, structural enhancement, and motility features—are aggregated to form a comprehensive feature set for each node.

Inspired by DeepMind's GraphCast multi-mesh, our method constructs a multi-level graph network to efficiently facilitate message passing at multiple distances within the organelle graph[47]. Intuitively, this graph represents intra-organellar feature dependence, where each node depends, in part, on all other nodes' features within the organelle. The adjacency matrix's construction begins with the selection of a tip node, connecting nodes at increasing powers of two to establish a multi-level mesh (Fig. 5a). A detailed explanation of the multi-mesh creation scheme can be found in Supplementary Note 13.

The graph autoencoder architecture is central to our methodology (Fig. 5b)[54]. We use an initial multilayer perceptron (MLP) to transform the inputs from our original feature set dimension to 512 dimensions, followed by a sigmoid linear unit (SiLU), or Swish, activation function, and layer normalization[55]. The encoder uses an independently weighted 16-layer graph neural network (GNN) with a linear MLP, again followed by Swish activation and layer normalization, with a mean aggregation message passing step across the multi-mesh and a final residual connection for each layer[56,57]. The decoder is similarly composed of 16 independently weighted layers, but instead uses a graph attention network (GAT) operator with a 20% dropout, followed again by a Swish activation function, layer normalization, and a residual connection[58]. Finally, we transform the 512 features back to the original feature set's dimensions. A more detailed explanation of the model can be found in Supplementary Note 14.

Our case study focuses on examining mitochondrial networks in cells treated with Ionomycin, a calcium channel depolarizer known to induce fission-like events in mitochondria[59]. Before treatment, we capture 20 volumes of fluorescently tagged mitochondria at a frequency of 1Hz (1 volume per second), leading to 18 pre-treatment graph embeddings (Fig. 5c). We then treat the cells with 4 uM of Ionomycin, and begin imaging for up to 120 volumes, again at 1Hz. We run the dataset through feature extraction with Nellie, construct its multi-mesh graph, and normalize the nodes' features to zero mean and unit standard deviation. To train the model, we use a 70/30 train/test split of our data and use an MSE loss to compare the reconstructed features to the original normalized features during training. We run the training with an Adam optimizer and a learning rate of 0.01 until validation loss stops decreasing for more than 10 epochs[60]. We use the model with the lowest validation loss as our final model, which was achieved after only 40 epochs.

Post-training, we deployed the model's encoder to obtain latent space representations of each node across different timepoints. These embeddings allowed us to geometrically compare graphs between temporal frames via cosine distances, revealing distinct phases in the mitochondrial network's response to Ionomycin treatment. A control graph embedding was established by averaging the 18 temporal frames' pre-treatment graph embeddings, serving as a baseline for comparison. The cosine distance to the control embedding delineated a consistent period of alteration and gradual recovery post-treatment (Fig. 5d). We see that, compared to the control graph, Ionomycin shows a quick rise (~60s to the peak) followed by a recovery nicely modeled by an exponential decay (tau of ~63s) (Fig. 5e). Using t-SNE for dimensionality reduction, we visualize the divergence and eventual convergence of post-treatment graph embeddings towards the pre-treatment group, whose 18 points are essentially in the same position on the t-SNE plot (Fig. 5f)[61]. Surprisingly, we identify oscillatory patterns in the graph embeddings post-treatment, absent in pre-treatment embeddings (Fig. 5g). These oscillations, discerned via frequency-based bandpass filtering and Fourier analysis, suggest intriguing mitochondrial responses to Ionomycin treatment that are ripe for exploration, but whose analyses and interpretations lie outside the scope of this paper.

This study demonstrates a novel method for analyzing organellar organization and motility using graph-based latent space representations. The approach offers rich, exploratory insights into cellular dynamics, akin to cell-painting strategies in drug discovery. The potential applications of this technique are vast, ranging from rare event detection in single organelles, to detailed organelle network studies, to broader systemic analyses in various model systems, and can be expanded to graph-representations of other structures at both larger and smaller resolutions. We hope that this case study will inspire further innovative research in organellar microscopy, leveraging the power of graph-based analyses to uncover new dimensions in cellular biology.

**Case Study 3: Nellie's branch segmentation and skeletonization allows for in-depth characterization and comparisons of ER morphology, motility, and network topology.**

In cellular biology, characterizing and comparing complex organelle networks is crucial for understanding their structure and function. However, this task often presents significant challenges due to the intricate and dynamic nature of these networks. Here, we demonstrate how Nellie's segmentation and feature extraction capabilities can be leveraged to perform detailed analyses of endoplasmic reticulum networks, enabling both comprehensive characterization and nuanced comparisons between different cells.

To showcase Nellie's ability to characterize complex organelle networks, we analyzed a 3D lightsheet volume of the full ER network in a primary human fibroblast (hFB) cell (Fig. 6a, b). Using Nellie's segmentation algorithms, we isolated the largest connected network component and constructed a graph representation, with the component's network's skeleton branch junctions serving as nodes and branches as edges between nodes (Fig. 6c). This graph-based approach allowed us to apply concepts from graph theory to investigate the ER network topology. We examined features such as node degree (the number of other nodes connected to each node) and node betweenness centrality–how frequently a node is traversed when traveling between two nodes via their shortest path (Fig. 6a, d, e). We found that the distribution of node degrees in our ER network closely aligns with existing data from ER-specific quantification tools like ERnet, with a majority of nodes (56.83%) having a degree of 3 (Fig. 6f)[32].

Beyond characterization, researchers often need to compare organelle networks between different conditions, such as wild type versus treated cells, cells derived from different patients, distinct cell lines, etc. To demonstrate Nellie's comparative capabilities, we analyzed three sequential frames from both an hFB cell and a U2OS cell line. We compared ER network graph topologies between the two cell types and across temporal frames. As expected, we found minimal variability in average node degree, betweenness centrality, and normalized cyclomatic number between the three frames within each cell. Interestingly, however, significant differences were observed between the hFB and U2OS cell lines for these metrics (Fig. 6g-i). This suggests that while the ER network topology remains relatively stable over short time periods within a cell, there are distinct differences in ER organization between these two cells.

Nellie's branch segmentation capabilities allowed us to compare ER branch morphology and motility between cells and across frames. We examined features such as branch length and linear velocities. Similar to the topology analysis, we observed little variability in these features between frames of the same cell line, but significant differences between the hFB and U2OS cells (Fig. 6k-l).

To perform a more comprehensive comparison, we employed tensor decomposition techniques. This approach allowed us to extract weights of different features, cells, and temporal frames for the first component of the decomposition (Fig. 6m-o). The results provide a holistic view of the most influential factors distinguishing the ER networks between cell types and across time.

To further validate the distinctiveness of ER characteristics between cells and the consistency within temporal frames, we built random forest classifiers to predict the origin of a branch object. When classifying branches between hFB and U2OS cells, the model achieved high accuracy with a receiver operating characteristic (ROC) area under the curve (AUC) of 0.72 (Fig. 6p-q). In contrast, when attempting to classify branches between two separate temporal frames from the same cell, the model performed essentially no better than random chance, with an AUC of 0.51 and 0.52 for hFB and U2OS cells, respectively.

These classification results further underscore the intrinsic differences in ER branch features between the two cells while confirming the consistency of Nellie's outputs between temporal frames of the same cell. This consistency is crucial, as we would expect similarity between close temporal frames of the same cell under stable conditions.

In conclusion, this case study demonstrates Nellie's powerful capabilities in characterizing and comparing complex organelle networks. By leveraging advanced segmentation, feature extraction, and analytical techniques, Nellie enables researchers to gain deep insights into organelle structure and dynamics, even complex endomembrane organelles such as the ER. The ability to perform robust comparisons between different cell types or conditions opens up new avenues for understanding the role of organelle organization in cellular function and disease states.

**Discussion**

Nellie is an unbiased and efficient pipeline for accurately and automatically analyzing spatial and temporal features of organelles, while staying agnostic to the organelle or substructure in question. We use intracellularly optimized adaptive preprocessing to allow for multi-scale segmentation, and adaptive local peak detection to generate multi-scale markers for tracking. Our motion capture marker linkage method uses the most important regions in our data as waypoints, independent of instance segmentations, while our flow interpolation method allows for tracking of sub-voxel coordinates. Finally, our multi-level feature extraction allows cross-sectional and hierarchical analyses both within and between scales of interest. This allows users to interpret their data from a sub-voxel level, all the way up to an image-wide level. We fully automate parameter selection for every step of the pipeline in a way that adapts to the image context to ensure no tailoring of features is required by the user. Conveniently, we package our methods in a Napari plugin GUI to allow for visualization and point-and-click functionality for ease-of-use, while keeping the codebase modular to allow flexibility and extensibility for more advanced users.

The case studies we presented demonstrate Nellie's broad range of applications across multiple aspects of cellular biology. First, we demonstrated Nellie's ability to capture extensive metrics in both Golgi and mitochondria by using object and branch-based morphology and motility features to train random forest models and predict organelle type. These useful results set the stage for performing multi-organellar fluorescence imaging in single channels, extending the information available from microscopy datasets.

Second, we were able to construct multi-level mesh-like graph networks of mitochondria, which we used to train an unsupervised graph autoencoder, allowing for efficient variable-range message passing between nodes. From this trained autoencoder, we were able to compare latent space embeddings of the graph nodes to investigate the effect of ionomycin on mitochondrial networks, such as intrinsic feature oscillations and post-treatment effect-and-recovery dynamics. To our knowledge, this case study also establishes the first organelle-based graph autoencoder, a model that would not be possible without high quality segmentation and tracking, and a diverse and comparable feature set, both of which were only possible through Nellie. We predict this organelle graph autoencoder type of model to have a broad and useful range of applications, from treatment-based clustering and comparisons of organelles, reminiscent of Cell Painting methods, to local morphology and motility predictions of organelles, reminiscent of GraphCast's weather prediction, but at intracellular scales[47,62].

Third, we showcased Nellie's capabilities in characterizing and comparing ER networks across different cell types and temporal frames. By leveraging Nellie's advanced segmentation and skeletonization techniques, we were able to perform in-depth analyses of ER morphology, motility, and network topology. This case study highlighted Nellie's ability to not only characterize complex organelle networks using graph theory concepts but also to detect meaningful differences between cells while maintaining consistency across temporal frames within the same cell. The application of tensor decomposition and machine learning classification further demonstrated the power of Nellie in extracting and utilizing multi-dimensional feature sets for holistic comparisons of organelle networks.

We hope Nellie will promote imaging-based approaches for analyzing organelles and their perturbation-mediated disruptions. Additionally, we hope Nellie's ease-of-use will encourage open-access science by providing a simple way to share intracellular feature data in a comparable manner. We present Nellie as a catalyst for a new wave of scientific inquiry, where the complex weave and elaborate dance of organelles is not just observed, but deeply understood, and where the mysteries of the cell are unlocked, one pixel at a time.

## Methods

**Cell culture**

U2OS osteosarcoma cells used in the representative figures throughout the main text, and in the multi-mesh GNN case study were cultured following standard procedure at 37°C with 5% $CO_2$ in DMEM (Thermo #10567014) supplemented with 10% fetal bovine serum (Gibco #26140-079). Cell passaging occurred every 3-5 days, and downstream assays were performed before reaching 20 passages of growth. For the organelle unmixing case study, U2OS cells were cultured in DMEM supplemented with 10% FBS, 1X antibiotic-antimycotic solution and 1X GlutaMAX (Gibco).

Primary human fibroblasts (male, LifeLine Cell Technologies, FC-0024 Lot #03099) were cultured in DMEM with 5.5mM glucose (ThermoFisher #10567022) supplemented with 10% FBS. Fibroblasts were split every 7 days and all experiments were performed between 3-5 passages after thawing.

**Fluorescent labels**

To establish a stable cell line expressing a fluorescently labeled version of COX8A, a mitochondrial matrix protein, for representative images throughout the main text and the multi-mesh GNN case study, U2OS cells at passage #8 (one million cells) were transfected with 1ug of DNA (Davidson-COX8A-mEmerald construct, Addgene code 54160) using the SE Cell Line Nucleofector Kit (Lonza #V4XC-1032) following the manufacturer's protocols. Transfected cells were cultured on collagen-coated plates to facilitate the recovery process. After two days, cells were selected with 1mg/ml

Geneticin Selective Antibiotic (G418 Sulfate, ThermoFisher #10131035) for a week. Fluorescence signals were continuously monitored during the selection process. Upon recovery, G418 at a concentration of 0.5 mg/ml was used for stable cell line maintenance. FACS-sorting for moderate-expression cells was performed to enhance the homogeneity of cells containing labeled mitochondria. Lysosome staining was performed via a 30 min incubation of 1µM SiR-Lysosome (Cytoskeleton #CY-SC012) and washed out 2x with warm media.

To generate U2OS cells expressing three genetically encoded fluorescent markers targeted to organelles for the organelle unmixing case study, a plasmid was generated containing mEGFP targeted to the mitochondrial matrix with a human COX8 presequence, ECFP-tagged H2B (not used in this paper), and the first 82 residues of B4GALT1 tagged with mScarlet. MluI and AsiSI sites were used to insert left and right homology arms, respectively, for the CLYBL safe harbor locus flanking the coding sequence. 2 homozygous knock-in clones were generated in U2OS cells (ATCC #HTB96) using CRISPR editing. Clones were validated using Sanger sequencing of genomic DNA.

To fluorescently label the endoplasmic reticulum, fibroblasts and U2OS cells were plated in 8-well glass-bottom chamber slides (CellVis #C8-1.5H-N), pre-coated with 1:100 fibronectin (Sigma #F1141). After 4 hours of plating cells were transfected overnight with 10ppc of CellLight™ ER-RFP, BacMam 2.0 (ThermoFisher #C10591), as per the manufacturer's protocol. No other transfection reagent was used due to the active viral capacity of the CellLight formula. Fresh media was added to cells an hour prior to imaging.

**Image acquisition**

Cells were plated on fibronectin coated (Sigma #F1141-1mg) 8-well glass-bottom chamber slides (CellVis #C8-1.5H-N) and incubated for 4 hours (main text figures and multi-mesh GNN case study) or 24 hours (organelle unmixing case study and ER network imaging). Imaging was then performed on an in-house single objective light sheet microscope with a stage-top incubator maintaining a temperature of 37°C with 5% $CO_2$ throughout[63]. Videos were acquired of individual cells at a frequency of 1 3D volume/second using 5% laser power and 1ms exposure, calculated to be 2.62 uJ per volume on the sample.

Imaging of the ER was performed on a Nikon W1 spinning-disk confocal system with a Plan Apo VC 100x/1.4 oil objective, using a 561 nm excitation laser line and zET405/488/561/635m quad filter, and a stage-top incubator set at 37 C with 5% $CO_2$. Full 3D z-stacks were acquired with 250 nm slices for a total of 5.25 μm depth across the cell. Imaging 2D time series videos were collected at 1.6 frames $sec^{-1}$.

**Treatments**

For calcium-ionophore treatment experiments, Ionomycin (4μM, Thermo #I24222) in DMSO was manually injected into the media of the cell imaging dish.

**Random forest classifier**

For our random forest classifier in Case Study 1, we use the same hyperparameters across all three models - motility only, morphology only, and motility and morphology combined - to discriminate against mitochondria and Golgi within our organelle demixing case study. The random forest classifier is implemented via scikit-learn's RandomForestClassifier class, with the following hyperparameters: n_estimators=300, criterion='gini', max_depth=None, min_samples_split=2, min_samples_leaf=1, min_weight_fraction_leaf=0.0, max_features='sqrt', max_leaf_nodes=None, min_impurity_decrease=0.0, bootstrap=True, oob_score=False, n_jobs=-1, random_state=42, verbose=0, warm_start=False, class_weight=None, ccp_alpha=0.0, max_samples=None, monotonic_cst=None.

For the motility only model, the median, maximum, minimum, and standard deviation values of the following features are used as inputs: the angular velocity magnitude w.r.t. the branch's pivot point between t1 and t2 (rel_ang_vel_mag_12), the angular acceleration magnitude w.r.t. the branch's pivot point (rel_ang_acc_mag), the linear velocity magnitude w.r.t. the branch's pivot point between t1 and t2 (rel_lin_vel_mag_12), the linear acceleration magnitude w.r.t. the branch's pivot point (rel_lin_acc_mag), the linear velocity magnitude of the branch's pivot point between t1 and t2 (ref_lin_vel_mag_12), the linear acceleration magnitude of the branch's pivot point (ref_lin_acc_mag), the directionality of the branch w.r.t. the center of mass of the fluorescence intensity between t1 and t2 (com_directionality_12), and the rate of change of the directionality w.r.t. the center of mass of the fluorescence intensity (com_directionality_acceleration).

For the morphology only model, the following features are used as inputs: area, extent, solidity, the minimum, median, and maximum of the object's inertia tensor eigenvalues (inertia_tensor_eig_sorted_min, inertia_tensor_eig_sorted_mid, intertia_tensor_eig_sorted_max), the total length (branch_lengths), the average radius (branch_radius), the average tortuosity (branch_tortuosity), and the average aspect ratio (branch_aspect_ratio).

For the morphology and motility combined model, all the features from both the morphology and motility models were aggregated.

For both training and testing, all feature columns were first standardized (mean of 0, standard deviation of 1), then used as inputs to the random forest classifier.

# Figures

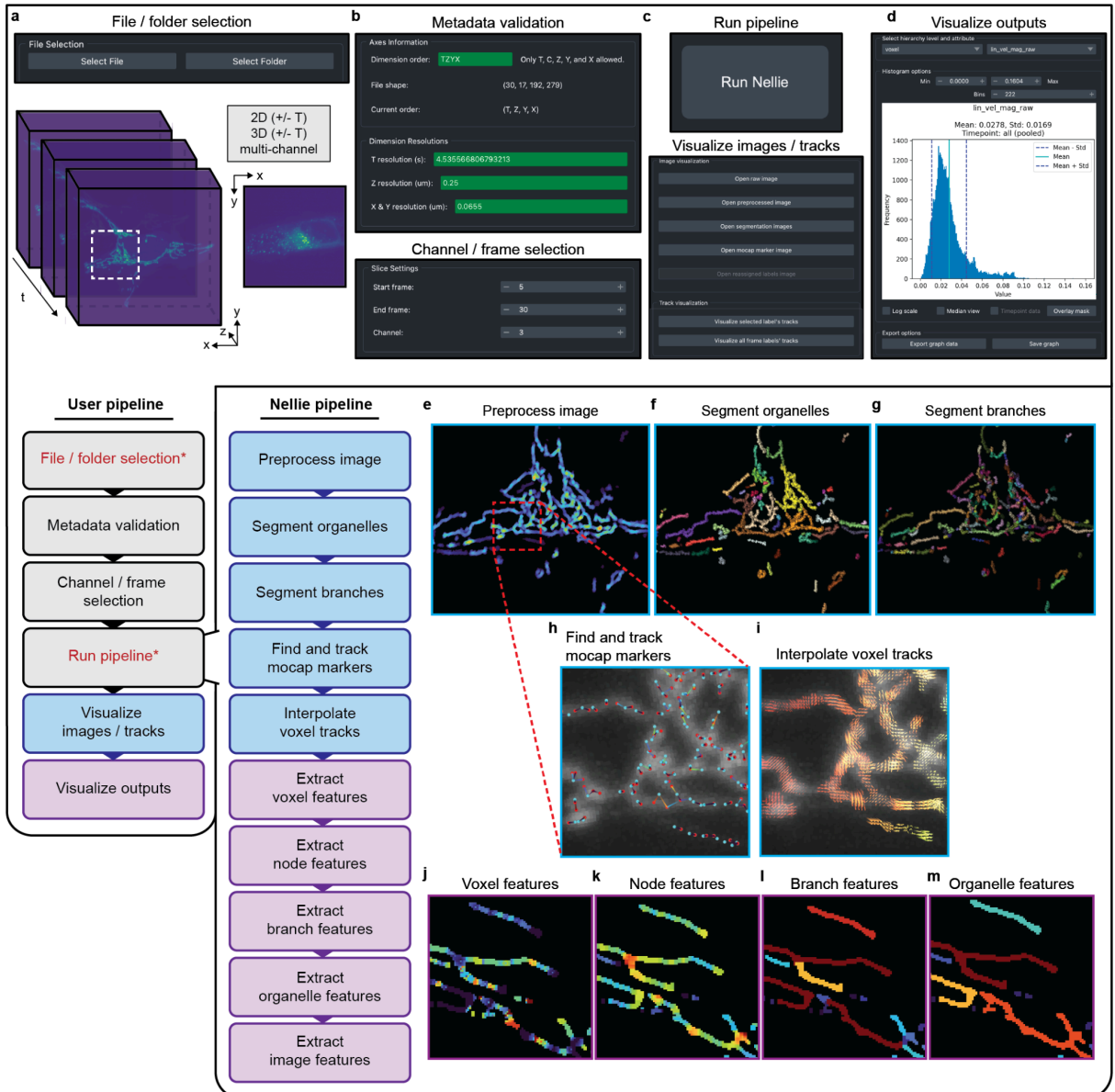

**Fig. 1: Nellie's user workflow and internal workflow**
The user pipeline (**a-d**) requires (red text, star) the selection and confirmation of data to process, and optionally (black) allows for correction of file metadata, data slice selection, and visualization of intermediate images and extracted features. **a**, The file / folder selection menu, capable of accepting single plane (2D, XY) or volumetric (3D, XYZ) data with or without multiple timepoints (T) and channels (Z). **b**, The validation menu, which automatically populates dimension order and dimension resolutions based on the file's metadata (top), and allows the user to select a specific channel or a range of temporal frames to run (bottom). **c**, The processing tab (top) starts Nellie's pipeline, and the visualization tab (bottom) allows the user to concurrently visualize image outputs and tracks (**e-i**) during the pipeline's run. **d**, The analysis tab, which allows the user to export and visualize specific extracted features for the different hierarchical levels and overlay those features on the original image (**j-m**). After data confirmation, the Nellie pipeline runs through preprocessing (**e**), segmentation of organelles (**f**) and branches (**g**), mocap marker detection and tracking (**h**), voxel-level track interpolation (**i**), and extraction of features at the hierarchical voxel (**j**), node (**k**), branch (**l**), and organelle (**m**) levels.

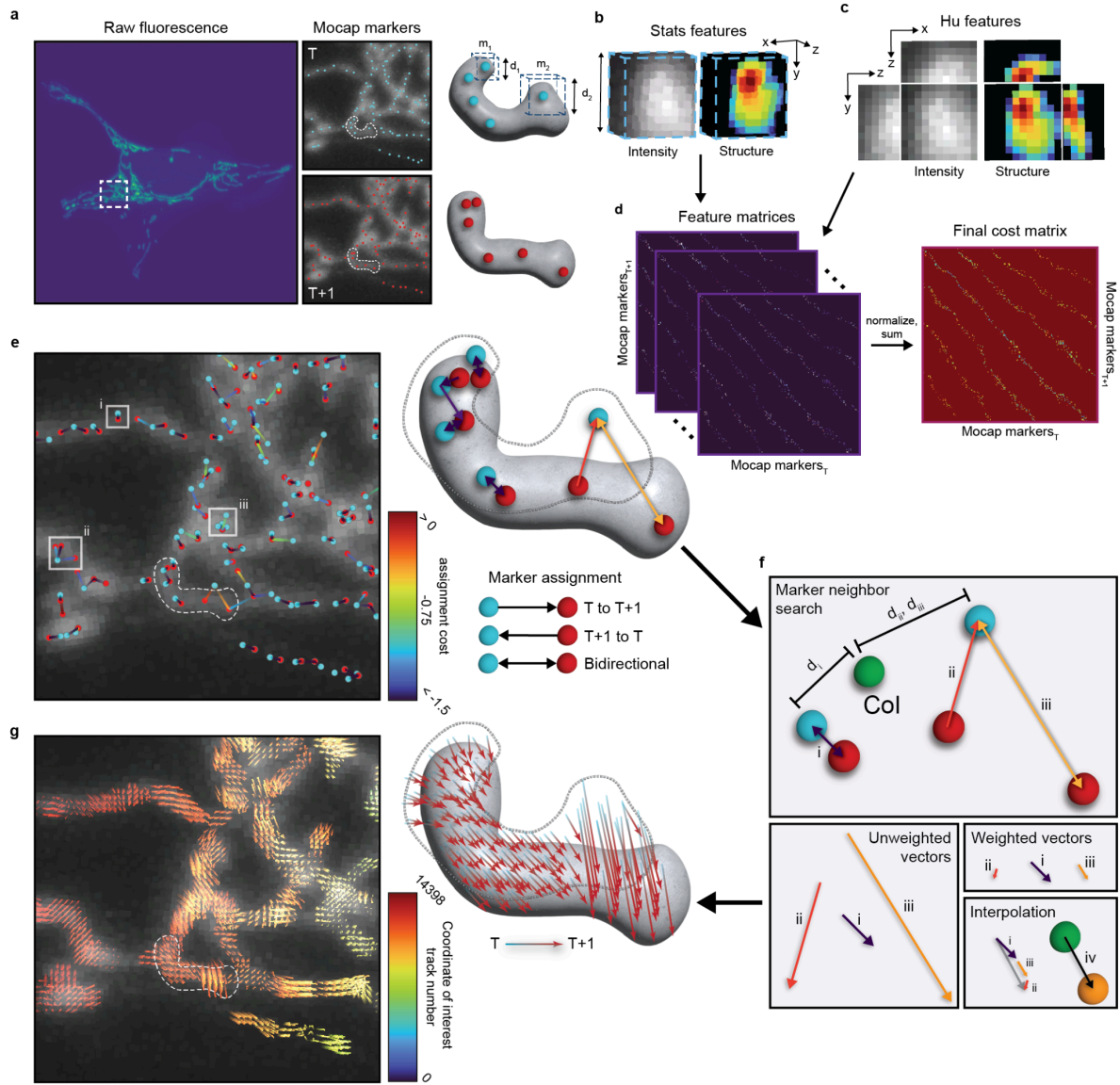

**Fig. 2: Linking motion capture markers and interpolating sub-voxel movements of organelles.**
**a**, Fluorescently labeled mitochondria (left) and their respective mocap markers for time T (top right, blue dots) and T+1 (bottom right, red dots). Feature search bounding boxes are marked by dotted lines for two mocap markers (m1, m2) with different radii (d1, d2) at time T. **b**, 3D search bounding box raw intensity values (left) and post-structural enhancement values (right) of m2, which are used for calculating stats features. **c**, 2d orthogonal max projections in xy, xz, and yz of raw intensity values (left) and post-structural enhancement (right) of mocap marker 2, which are used for calculating Hu features. **d**, Difference matrices of distance, Hu features, and stats features are calculated between mocap markers in frame T (columns) and T+1 (rows) to create weighted feature matrices (left), which are then z-score normalized and summed to create the final cost matrix for marker linkage (right). **e**, Mocap markers from frame T (blue dots) are linked to their best mocap marker match in T+1 (red dots) based on assignment cost (line colors), and vice-versa for T+1 to T, resulting in 1-to-1 (i), 1-to-N (ii), or N-to-1 (iii) matches. **f**, A coordinate of interest (CoI, green sphere) not corresponding to a mocap marker (red, blue spheres) has a flow vector interpolated via distance-weighted and cost-weighted vector summation of nearby assigned mocap marker linkages to a new coordinate at T+1 (orange sphere). **g**, Interpolated flow vectors for all voxels in the image (left) and flow vector representations (arrows) for all voxels within a mitochondrion (right) between time T (blue) and T+1 (red).

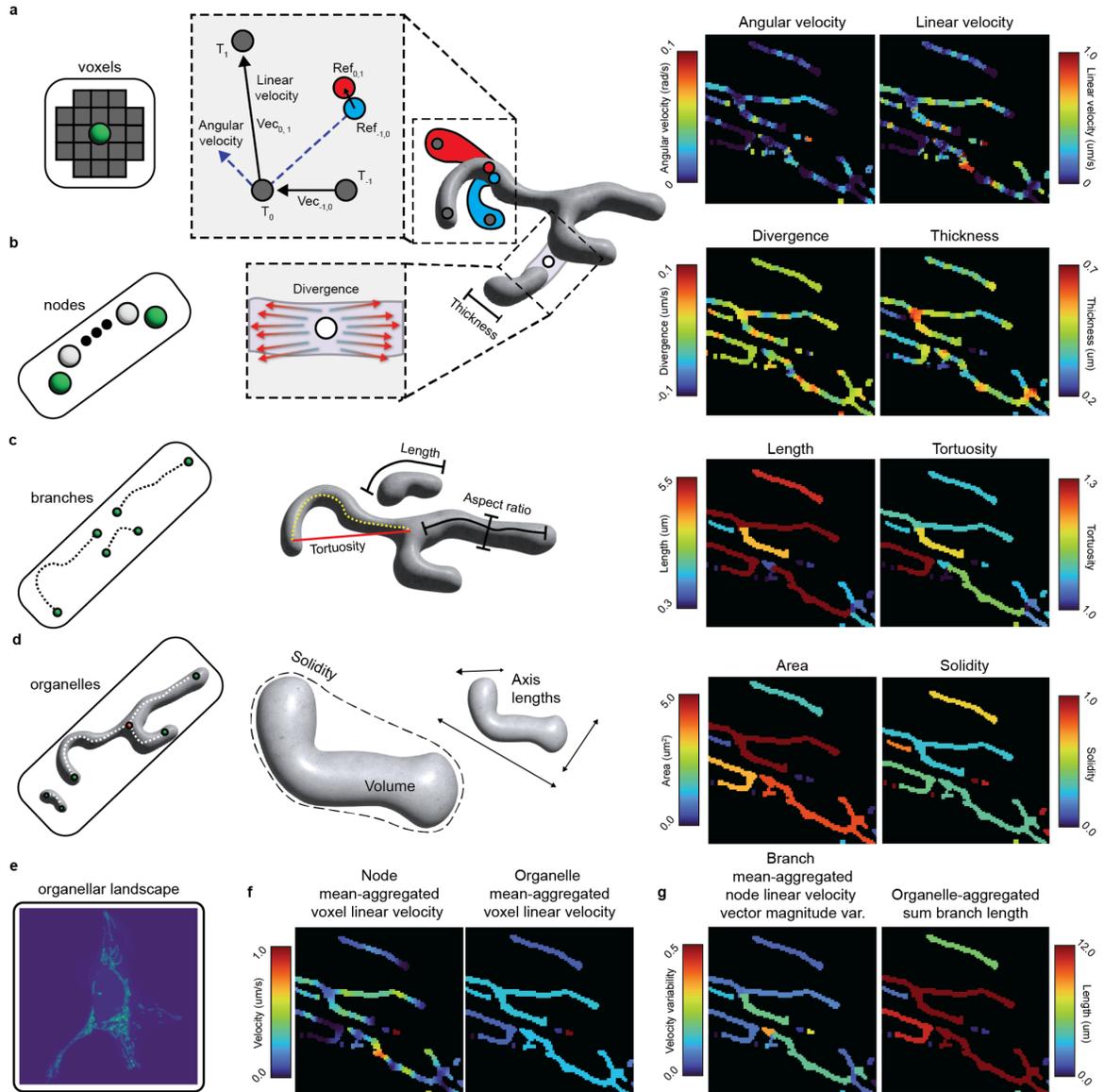

**Fig. 3: Extraction of spatial and temporal features of organelles at multiple hierarchical levels.**
**a**, Individual voxels represent the lowest hierarchical resolution of our organelle, containing feature information such as voxel intensities and motility metrics. Shown here is a subset of motility features extracted for one tracked voxel (gray dots) through time T-1 (blue structure) to T+1 (red structure). The tracked reference point for each time point (blue dot, red dot) is shown, and is used to calculate reference-adjusted linear and angular movements. **b**, Individual skeleton nodes represent the next level of the organellar hierarchy, encapsulating voxels within a radius corresponding to that node's border distance, containing features such as divergence / convergence of surrounding flow vectors, thickness, and more. **c**, Individual branches represent the next level of the organellar hierarchy, containing information on the curviness (tortuosity), length, aspect ratio, and more. **d**, Organelles represent the next level of the organellar hierarchy, spatially disconnected components in the image, containing information about volume, solidity, axis lengths, and more. **e**, The organellar landscape as a whole represents the highest level of our hierarchy, containing aggregate information from all levels below it. **f**, Each hierarchical level can aggregate metrics from its lower level's components, such as a node's or organelle's voxels' mean linear velocity. **g**, Other metrics as complex as a branch's nodes' mean linear velocity vector magnitude variability, or as simple as an organelle's branches' sum branch length can be calculated as well. All images were color-mapped via Nellie's Napari plugin.

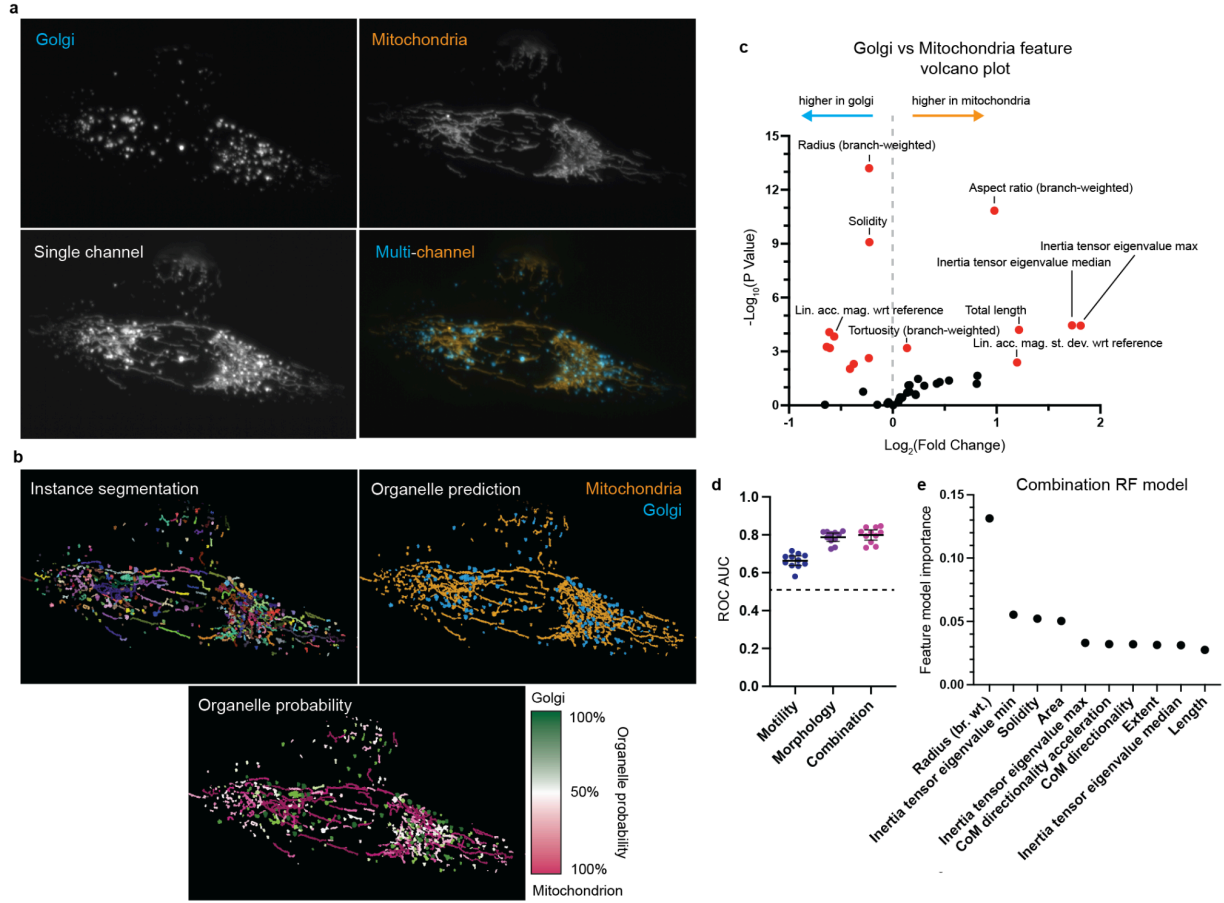

**Fig. 4: Single-channel multi-organelle unmixing using features extracted by Nellie.**
**a**, Raw intensity images of fluorescently labeled Golgi and mitochondria (top). A single channel max-intensity projection (bottom left) over the channel dimension, combining the fluorescence signal from both channels (bottom right) into one. **b**, The branch-based instance segmentation output derived from Nellie of the single channel image (top left). The binary organelle prediction from a trained random forest classifier (top right) and their corresponding probabilities (bottom). **c**, A volcano plot showing features upregulated (positive) or downregulated (negative) in mitochondria compared to Golgi, with significantly different (p<0.05) features colored in red. **d**, Areas under the receiver operating characteristic curve for N=11 leave-one-out random forest classifier models with only motility features, only morphology features, or a combination of both. Bars are mean +/- standard deviation. **e**, Random forest model feature-importance for the classification of the representative image's organelles.

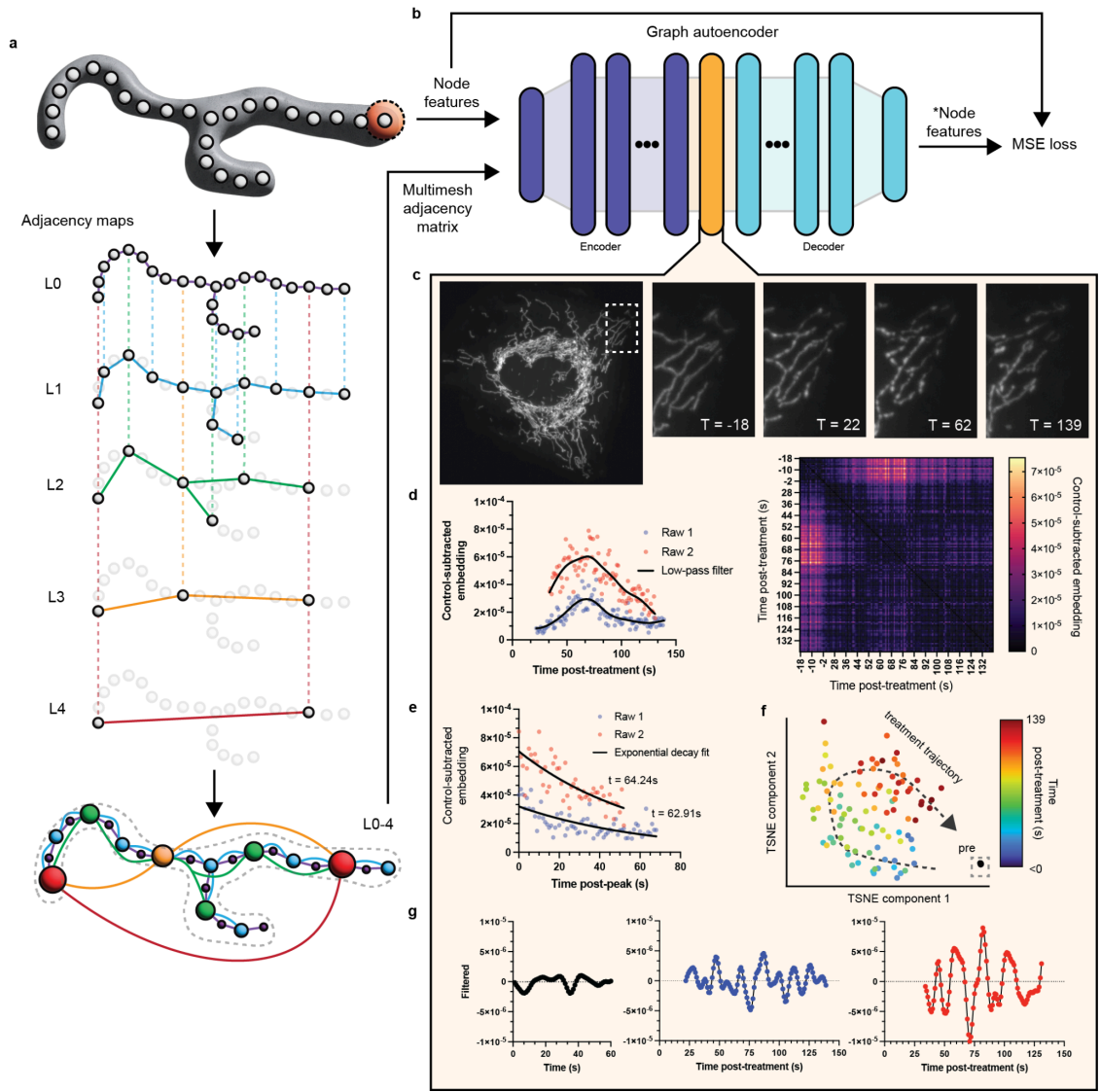

**Fig. 5: Quantification of the evolution of Ionomycin-treated mitochondria multi-mesh graphs via comparisons of graph autoencoder latent space embeddings.**
**a**, Example construction of a multi-mesh adjacency network of a mitochondrion, where each level (L) corresponds to the power-of-2 node-jump distance used for node linkage at that level, allowing for efficient message-passing at variable ranges within one organelle. **b**, The multi-mesh adjacency matrix (node edges) and node features are used as the inputs to a graph autoencoder. The GNN uses an MSE loss derived from the comparisons on the differences of the input and output node features for unsupervised training. After training, the latent space representation outputs (orange) from the encoder can be used for vector-based similarity comparisons. **c**, Fluorescently labeled mitochondria with representative images throughout their responses to Ionomycin treatment. **d**, Cosine distances of latent space embeddings to the average of the 18 pre-treatment latent space embeddings for two independent samples (left), or to all latent space embeddings (right) of mitochondria multi-mesh graphs after Ionomycin treatment. Raw 1 corresponds to the representative images in c. **e**, Cosine distances of latent space embeddings to the average of the 18 pre-treatment latent space embeddings post-peak, fit to exponential decay curves. **f**, t-SNE dimensionality reduction of latent space embeddings colored by time post-treatment. **g**, Band-pass filtered cosine distances of latent space embeddings of untreated mitochondria to the first frame's mitochondrial latent space embeddings (left) or of Ionomycin treated mitochondria to the first pre-treatment frame's mitochondrial latent space embeddings (middle and right).

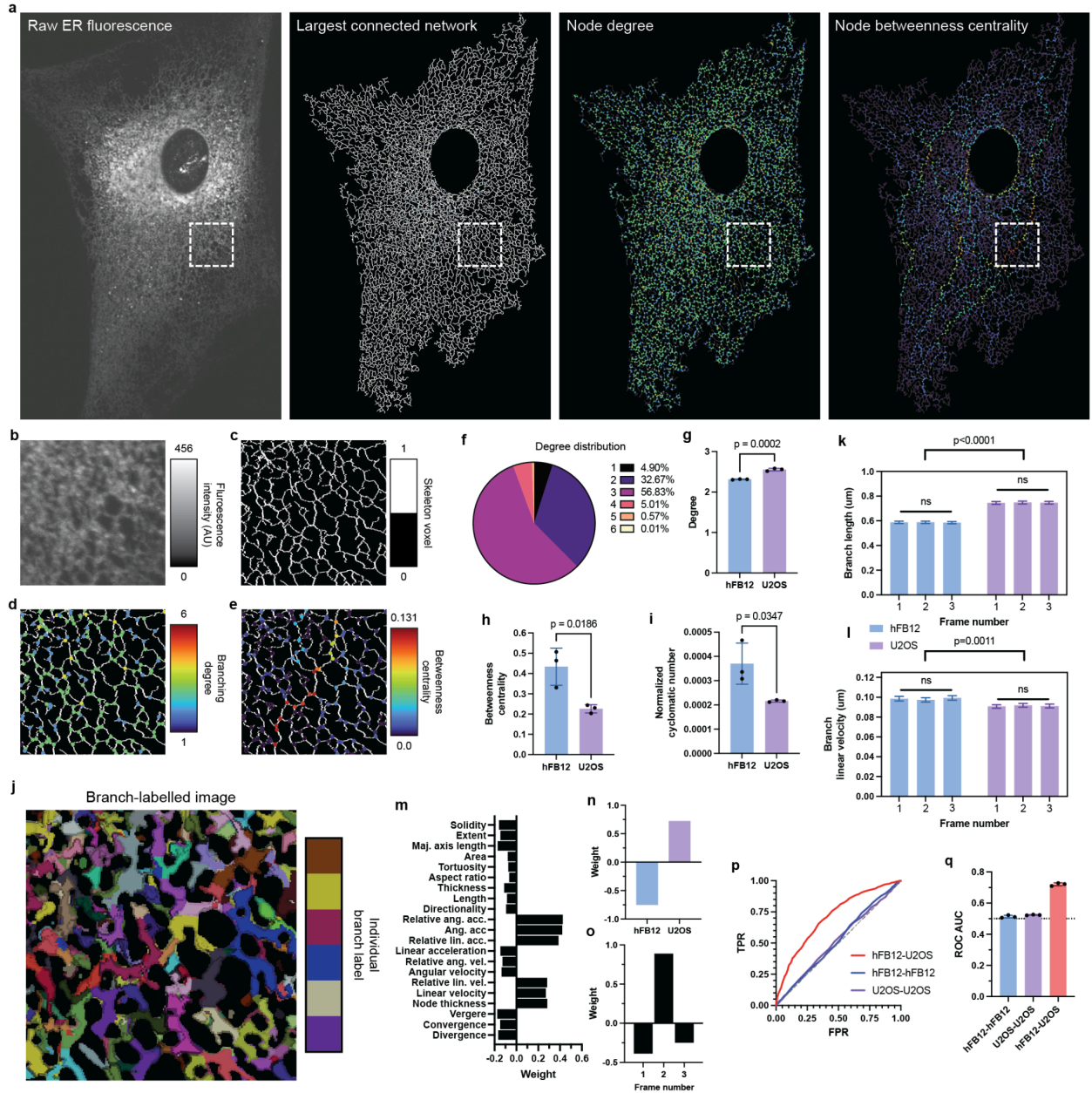

**Fig. 6: Nellie allows for in-depth characterization and comparisons of ER morphology, motility, and network topology.**

**a**, Raw fluorescence of a 3D endoplasmic reticulum (ER) network from a primary human fibroblast cell (hFB12) labeled with CellLight-ER. Shown also is the ER network's largest connected component skeleton, and its corresponding connectivity graph with nodes color coded based on node degree and node betweenness centrality. Also shown are close-ups of the regions denoted by the white bounding boxes in the raw (**b**), skeleton (**c**), degree color coded (**d**), and betweenness centrality color coded (**e**) images. **f**, The distribution of node degrees in the ER network from **a**. The mean degree (**g**), betweenness centrality (**h**) and normalized cyclomatic number (**i**) in 3 sequential 2D temporal frames of either a hFB12 or U2OS cell. Bars are mean +/- standard deviation. p-value calculated via an unpaired two-tailed t-test. N=3 temporal frames. **j**, The instance segmentation of individual ER branches in the same close-up region as **b-e**, randomly color coded by label number. Branch lengths (**k**) and linear velocities (**l**) of 3 sequential 2D temporal frames of either a hFB12 (blue) or U2OS (purple) cell. Bars are mean +/- 95% confidence interval. Intra-cell p-value calculated via Kruskal-Wallis test with a corrected Dunn's multiple

comparisons test. N=6109, 6242, 6229 branches for hFB12 frames 1-3, respectively, and N=6319, 6507, 6526 branches for U2OS frames 1-3 respectively. Inter-cell p-value calculated via an unpaired two-tailed t-test. N=18590, 19352 hFB12 and U2OS branches, respectively. Tensor decomposition weights of the first component for branch features (**m**), cell types (**n**), and temporal frame number (**o**). Receiver operating curves (ROC) (**p**) and corresponding areas under the curve (AUC) of the ROC (**q**) for random forest models to classify between ER branches between hFB12 and U2OS cells (red), two separate temporal frames of hFB12 cells (blue) or of U2OS cells (purple). N=3 random forest classifiers for each group.

**Nellie: Automated organelle segmentation, tracking,
and hierarchical feature extraction in 2D/3D live-cell microscopy**

**Supplementary Materials**

Austin E. Y. T. Lefebvre[1,*], Gabriel Sturm[1,2], Ting-Yu Lin[1], Emily Stoops[1], Magdalena Preciado López[1], Benjamin Kaufmann-Malaga[1], Kayley Hake[1]

[1]Calico Life Sciences LLC, South San Francisco, CA, 94080
[2]Department of Biochemistry and Biophysics, University of California San Francisco, San Francisco, CA, 94158

*Address correspondence to

Austin E. Y. T. Lefebvre, Ph.D.
Calico Life Sciences
1170 Veterans Blvd
South San Francisco, CA, 94080
Email: austin.e.lefebvre+nellie@gmail.com

**Supplementary Note 1: Runtime benchmarking of intracellular segmentation and tracking algorithms**

Here we test segmentation and tracking runtimes of Nellie and other SOTA segmentation and tracking tools on datasets of increasing sizes (Extended Fig. 1a, b).

All segmentation pipelines were executed as described in Supplementary Note 4, with consistent methodologies applied across the various tools for a fair comparison. The runtime performance of each segmentation algorithm—MitoGraph, Mitometer, Nellie (CPU), and Nellie (GPU)—was tested on increasingly larger datasets, ranging from 3.7 MB to 1.87 GB, to evaluate their scalability and efficiency.

For Nellie, the segmentation process comprises multiple stages, including pre-processing, semantic segmentation, instance segmentation, skeletonization, and branch relabeling. The combined runtime of these stages is reported as the total segmentation time. To test the performance on larger datasets, we used two temporal frames of a dataset representing a single yeast cell with fluorescently labeled mitochondria and mirrored the dataset laterally to double its size with each subsequent run (Extended Fig. 1c, d). This methodology ensured a fair assessment of how each tool handles progressively larger 3D datasets, starting from the initial 3.7 MB and scaling up to 1.87 GB.

The results show that Nellie, running on either the CPU or GPU vastly outperforms both MitoGraph and Mitometer in terms of segmentation speed. Notably, Nellie exhibits a near-linear increase in segmentation time as dataset size grows, demonstrating excellent scalability and computational efficiency. For the smallest dataset (3.7 MB), for both CPU and GPU runs, Nellie completes segmentation in under 5 seconds, while MitoGraph requires nearly 40 seconds, and Mitometer requires approximately 10 seconds. This difference becomes even more pronounced as the dataset size increases. At 233.1 MB, Nellie (GPU) finishes the segmentation task in roughly 1 minute and Nellie (CPU) finishes in just over 7 minutes, whereas MitoGraph takes close to 45 minutes and Mitometer requires close to 10 minutes. At the largest dataset size tested, 1.87 GB, Nellie (GPU) completes segmentation in 9 minutes and Nellie (CPU) in just under an hour, maintaining a steady performance across larger datasets. In stark contrast, MitoGraph requires over 17 hours to process the same dataset, and Mitometer takes just over an hour. These results highlight the large performance gains provided by Nellie when using GPU acceleration, particularly for large datasets, where the other tools struggle to maintain reasonable runtimes.

Similarly, the runtime performance of the different tools for tracking tasks follows a similar experimental setup. The tracking pipelines were executed as described in Supplementary Note 9, with MitoTNT relying on segmentation outputs from MitoGraph, due to the specific file format output from MitoGraph being required by MitoTNT, and Nellie and Mitometer using Nellie's segmentation outputs for tracking mitochondria from frame 0 to frame 1 of the two temporal frames of the 3D dataset. For Nellie, tracking time reflects the combined runtime of the motion capture marking and matching pipelines.

Again, the results show that Nellie (GPU) vastly outperforms the other tools in terms of tracking time, particularly for larger datasets. For the smallest dataset (3.7 MB), Nellie completes tracking in less than 1 second over both the GPU and CPU, while MitoTNT requires nearly twice that time, and Mitometer about five times that time. As the dataset size increases to 233.1 MB, Nellie (GPU) completes the tracking task in about 11 seconds and Nellie (CPU) in roughly 1 minute, compared to nearly 20 minutes for MitoTNT and more than 5 minutes for Mitometer. For the largest dataset (1.87 GB), Nellie (GPU) completes the tracking task in less than 3 minutes and Nellie (CPU) in 14 minutes, whereas MitoTNT could not complete the tracking of the two largest datasets in less than a day (our cutoff), and Mitometer required just under 1 hour. These results further underscore the significant advantages of Nellie in terms of both segmentation and tracking efficiency, especially when deployed on GPU hardware.

These results demonstrate that Nellie, whether running on CPU or GPU, computationally outperforms MitoGraph and Mitometer in segmentation tasks, and MitoTNT and Mitometer in tracking tasks across a wide range of dataset sizes. The use of GPU acceleration provides an especially dramatic speedup, allowing Nellie to handle even the largest datasets tested in a fraction of the time required by the other tools. This makes Nellie a powerful and scalable tool for high-throughput organelle segmentation and tracking in 3D microscopy datasets, enabling researchers to process and analyze large volumes of data quickly and accurately.

**Supplementary Note 2: Nellie's extended preprocessing pipeline**

**Optimization of multi-scale Gaussian filtering for anisotropic images in cellular fluorescence microscopy datasets**

In our study, the establishment of minimum and maximum radii for filtering operations is guided by the specific parameters of the imaging metadata, ensuring an alignment with the physical dimensions represented within the dataset. The lower bound for the radius is determined by selecting the greater of two values: a predefined threshold of 0.20 micrometers, which approximates the diffraction limit of light, or the pixel size in the X dimension of the image. This approach ensures that the filtering scale remains relevant to the resolution characteristics of the dataset. Conversely, the upper bound is uniformly set at 1 micrometer, marking the maximal scale of interest for our analysis. These radii are subsequently translated into pixel/voxel dimensions. Mathematically, we can express this as:

$$r_{min} = max(0.2um,\ pixel\ size_X)$$

$$r_{max} = 1.00\ um$$

$$r_{min\ px} = \frac{r_{min}}{pixl\ size_X}$$

$$r_{max\ px} = \frac{r_{max}}{pixel\ size_X}$$

For the Gaussian filtering process, we derive the minimum and maximum sigma values directly from these pixel-based radii, applying reduction factors of one-half and one-third, respectively. This strategy is designed to calibrate the filter's sensitivity to features of varying intracellular sizes:

$$\sigma_{min} = min\left(\frac{r_{min\,px}}{2}, \frac{r_{max\,px}}{3}\right)$$

$$\sigma_{max} = max\left(\frac{r_{min\,px}}{2}, \frac{r_{max\,px}}{3}\right)$$

To span the range of feature scales, we implement a series of five sigma values, ensuring comprehensive coverage across the standard deviation spectrum. The inter-sigma step size is calculated to bridge the desired feature scales effectively, imposing a minimum threshold on step size to prevent redundantly small increments that yield negligible changes in filtering outcomes:

$$N_{\sigma} = 5$$

$$step\ size_{\sigma} = max\left(0.2, \frac{\sigma_{max} - \sigma_{min}}{N_{\sigma}}\right)$$

$$\sigma_{array} = [\sigma_{min}, \sigma_{min} + step\ size_{\sigma}, ..., \sigma_{max}]$$

This structured approach yields a discrete set of sigma values, each corresponding to a specific standard deviation within the Gaussian filtering framework, thus enabling the filter to adaptively respond to biological features of different scales.

In addressing the challenge of anisotropic datasets prevalent in microscopy imaging – characterized by equivalent resolutions in the X and Y axes but reduced resolution along the Z axis – it is imperative to adjust the kernel values of 3D convolutional filters. This adjustment is critical within our preprocessing framework, which utilizes Gaussian filters across a multi-scale parameter space in both Frangi and Laplacian of Gaussian filtering applications. To counteract anisotropy, we construct a sigma vector for each scale, maintaining uniform standard deviations for the X and Y dimensions while adapting the Z dimension's standard deviation in accordance with the voxel resolution ratio (Z/X). This methodology ensures a consistent filtering effect across the dataset, irrespective of inherent resolution variances. Mathematically, for a given sigma value σ, we define the sigma vector as:

$$\sigma_{vec3} = \left(\sigma \frac{res_X}{res_Z}, \sigma, \sigma\right)$$

In 2D datasets, which are typically (and for the use of Nellie, we assume to be) exempt from anisotropic concerns, a singular sigma value is employed without modifications for anisotropy:

$$\sigma_{vec2} = (\sigma, \sigma)$$

**Automation of the Frangi filter for optimized contrast enhancement of intracellular structures**

For each sigma value of our Gaussian filter, several steps are undertaken to both automate and optimize the Frangi filter to enhance the contrast of structures at the intracellular scale. First, the gamma parameter for the Frangi filter is derived from the minimum of the triangle and Otsu threshold values – termed henceforth as the Minotri threshold – of the Gaussian filtered image, providing a size-adaptive approach to contrast enhancement of structural features (Supplementary Note 3)[64,65]:

$$\gamma_\sigma = min\left(T_{tri}(I_\sigma), T_{Otsu}(I_\sigma)\right)$$

where $I_\sigma$ is the Gaussian filtered image at scale σ, and $T_{tri}$ and $T_{Otsu}$ are the triangle and Otsu thresholding functions, respectively.

Next, the Hessian matrix for each sigma level is computed:

$$H_\sigma = \begin{bmatrix} \left[\dfrac{\delta^2 I_\sigma}{\delta x^2}, \dfrac{\delta^2 I_\sigma}{\delta y \delta x}, \dfrac{\delta^2 I_\sigma}{\delta y \delta z}\right] \\ \left[\dfrac{\delta^2 I_\sigma}{\delta y \delta x}, \dfrac{\delta^2 I_\sigma}{\delta y^2}, \dfrac{\delta^2 I_\sigma}{\delta y \delta z}\right] \\ \left[\dfrac{\delta^2 I_\sigma}{\delta z \delta x}, \dfrac{\delta^2 I_\sigma}{\delta z \delta y}, \dfrac{\delta^2 I_\sigma}{\delta z^2}\right] \end{bmatrix}$$

The Hessian matrix is rescaled and its Frobenius norm is calculated:

$$H_{\sigma\,rescaled} = \dfrac{H_\sigma}{max(|H_\sigma|)}$$

$$F_\sigma = \|H_{\sigma\,rescaled}\|$$

The Hessian matrix is masked using the Minotri threshold of the Frobenius norm:

$$Mask_\sigma = F_\sigma > min(T_{tri}(F_\sigma), T_{Otsu}(F_\sigma))$$

$$H_{\sigma\,masked} = H_\sigma \cdot Mask_\sigma$$

which is a modified approach to MitoGraph's masking method[12]. This approach not only significantly improves computational and memory efficiency, which is particularly crucial for eigenvalue calculations of large datasets, but also allows for completely automatic parameter selection. The eigenvalues of the vectorized Hessian matrix are calculated in memory-efficient chunks and sorted by their absolute values within individual voxels:

$$\lambda_{1,\sigma}, \lambda_{2,\sigma}, \lambda_{3,\sigma} = sort(|eigvals(H_{\sigma\,masked})|)$$

where $|\lambda_{1,\sigma}| \leq |\lambda_{2,\sigma}| \leq |\lambda_{3,\sigma}|$

Finally, Frangi dissimilarity metrics are calculated:

$$R_{A,\sigma} = \frac{|\lambda_{2,\sigma}|}{|\lambda_{3,\sigma}|}$$

$$R_{B,\sigma} = \frac{|\lambda_{1,\sigma}|}{\sqrt{|\lambda_{2,\sigma} \cdot \lambda_{3,\sigma}|}}$$

$$S_\sigma = \sqrt{\lambda_{1,\sigma}^2 + \lambda_{2,\sigma}^2 + \lambda_{3,\sigma}^2}$$

The vesselness measure $V_\sigma$ at scale $\sigma$ is computed:

$$V_\sigma = \left(1 - e^{\frac{-R_{A,\sigma}^2}{2\alpha^2}}\right) \cdot \left(e^{\frac{-R_{B,\sigma}^2}{2\beta^2}}\right) \cdot \left(1 - e^{\frac{-S_\sigma^2}{2\gamma_\sigma^2}}\right)$$

where α and β are shape-balanced parameters described in detail in the next section, and γ is the scale-adaptive parameter derived above. These metrics as a whole are used to calculate the inherent structural measure of the organelles at that specific scale, rather than the traditional approach of using the same metrics across all scales. The final multi-scale structure-enhanced filtered image – termed henceforth as the preprocessed image – is constructed by compiling the maximum voxel-wise value across all scales, ensuring that the most prominent features at each scale are captured in the final image (Fig. 1e, Supplementary Data Fig. 1):

$$I_{pre} = \max_\sigma (V_\sigma)$$

**Automated selection of the Frangi filter parameters**

Though the Frangi filter, which uses the eigenvalues of the Hessian matrix of the image as attributes, is historically used to accentuate vessel structures within images, it also works exceptionally well at accentuating structural components in general, as long as proper Frangi parameters and scale ranges over which to run the filter are chosen. As organelle shapes vary dramatically both between and within different types of organelles, we must carefully balance the Frangi filter's parameters and scale range in order to properly capture this diversity without accentuating any specific type of organelle. In this regard, we explicitly and specifically modify the Frangi filter to allow for automatic adaptation to specific image contexts, including and beyond scale variations.

The alpha parameter controls the sensitivity of the filter to the deviation from a blob-like structure. In the context of structure detection, it helps in distinguishing between disc-like, blob-like, and tube-like structures based on the eigenvalues of the Hessian. In practice, a lower alpha value makes the filter more sensitive to disc-like structures and less sensitive to tube-like structures, while increasing alpha will do the opposite, making the filter less responsive to blob-like and disc-like structures and more focused on tube-like features. We set:

$$\alpha^2 = 0.5$$

This setting provides a moderate sensitivity to blob-like structures (Supplementary Data Fig. 2). This allows the filter to be balanced in its response to tube-like structures and blob-like structures. This setting means the filter will enhance tubular structures while still allowing for some degree of response to blob-like features. It won't be overly discriminative against structures that slightly deviate from the traditionally Frangi-directed tubular shape.

The beta parameter controls the filter's sensitivity to the deviation from a disc-like structure. It is directly involved in the suppression of the response of the filter to disc-like and blob-like structures. In practice, adjusting beta alters the filter's response to different types of structures. A lower beta value increases the sensitivity of the filter to disc-like structures, while a higher beta value makes the filter more selective for tubular structures. We also set:

$$\beta^2 = 0.5$$

This setting provides a moderate level of suppression for disc-like structures (Supplementary Data Fig. 2). This prevents the filter from being overly aggressive in suppressing disc-like features.

We modify the calculation of R_B, which measures the deviation from a line-like structure:

$$R_{B,\sigma} = \frac{|\lambda_{2,\sigma}|}{\sqrt{|\lambda_{2,\sigma} \cdot \lambda_{3,\sigma}|}}$$

where $\lambda_{1,\sigma}, \lambda_{2,\sigma}, \lambda_{3,\sigma}$ are the eigenvalues of the Hessian matrix, sorted such that

$$|\lambda_{1,\sigma}| \leq |\lambda_{2,\sigma}| \leq |\lambda_{3,\sigma}|.$$

This results in the filter deemphasizing purely line-like and tube-like structures and emphasizing both line and blob-like structures. Running this change through our segmentation validation comparison studies showed slightly improved F1 and IoU scores on the multi grid test, and no changes in the pixel size and separation tests.

The gamma parameter is a scaling parameter that controls the sensitivity of the filter to the overall magnitude of the eigenvalues. This parameter essentially acts as a normalization factor in the function. In practice, changing gamma affects the filter's sensitivity to the background noise and contrast of the image. A higher gamma value can suppress the response of the filter to areas with low contrast, reducing the likelihood of detecting false positives due to noise. Conversely, a lower gamma would increase sensitivity to lower contrast regions but would also increase vulnerability to noise. This is the most important parameter to set properly, as poor gamma selection will lead to failure in correctly thresholding regions of interest post-filtration. By automating the gamma selection process by deriving it from the Minotri threshold, we're linking gamma to the inherent contrast characteristics of the image at that scale. This is a novel way to make the filter's response adaptive to each specific scale's contrast profile. This is particularly useful where samples can vary greatly in their staining, illumination, and intrinsic structural contrast. Additionally, by adapting gamma using our Minotri threshold method at each scale independently, we automatically adjust gamma to enhance subtler structures in low-contrast images, while in high-contrast images, it prevents over-enhancement and preserves detail. This ensures that the filter's sensitivity to the eigenvalue magnitudes is tailored not just to the image's overall contrast profile, but also to the contrast profile relevant to structures at each specific scale. This means for finer scales (smaller Gaussian sigma) which targets smaller structures, the gamma value will adapt to the contrast nuances at that scale, and similarly for larger scales.

**Preprocessing refinement techniques for structural contrast enhancement**

Following the initial multi-scale structural enhancement of organelles, the pipeline incorporates further post-preprocessing techniques to refine the output images. The refined post-preprocessing begins with the application of a masking technique to the preprocessed image, aiming to eliminate regions with minimal structural information. This is achieved by setting a threshold that masks regions falling below the 1st percentile of non-zero pixel values:

$$T = percentile(I_{pre}, 1)$$

where percentile(A, p) returns the pth percentile of the non-zero values in A.

From this threshold, the initial mask is created:

$$M(x, y, z) = \{1 \text{ if } I_{pre}(x, y, z) > T \text{ else } 0\}$$

For 2D images, this simplifies to:

$$M(x, y) = \{1 \text{ if } I_{pre}(x, y) > T \text{ else } 0\}$$

This ensures a focus on regions rich in meaningful content. Subsequently, a binary opening operation is performed on the mask to smooth out noise within the structures. This morphological operation, involving erosion followed by dilation, refines the mask's boundaries, and removes spurious pixels that could detract from the integrity of the analysis:

Let S be the structuring element. For 3D images, S is a 2x2x2 cube, while for 2D images, S is a 2x2 square.

For erosion:

$$E = M \ominus S = \{z | S_z \subseteq M\}$$

where $S_z$ is the translation of S by vector z.

For dilation:

$$D = E \oplus S = \{z | \widehat{S}_z \cap E \neq \emptyset\}$$

where $\widehat{S}_z$ is the reflection of S.

The final refined mask is the result of the binary opening:

$$M_{refined} = (M \ominus S) \oplus S$$

The final refined preprocessed image I_refined is obtained by applying the refined mask to the original preprocessed image:

$$I_{refined}(x, y, z) = I_{pre}(x, y, z) \cdot M_{refined}(x, y, z)$$

For 2D images:

$$I_{refined}(x, y) = I_{pre}(x, y) \cdot M_{refined}(x, y)$$

These post-preprocessing steps collectively ensure that the pipeline produces refined output images. This approach not only enhances the visibility of organelles but also prepares the images for accurate segmentations of regions of genuine structural significance.

2D datasets typically have reduced structural noise compared to 3D datasets, which can be attributed to several potential factors, including the reduced complexity of organelles that can be captured in a single plane, compared to the three-dimensional complexity of organelles in ZYX datasets. Furthermore, anisotropy present in 3D datasets - where differences in resolution between the XY plane and the Z-axis lead to variability in structural clarity - typically necessitates more complex filtering approaches to achieve uniform enhancement across all dimensions.

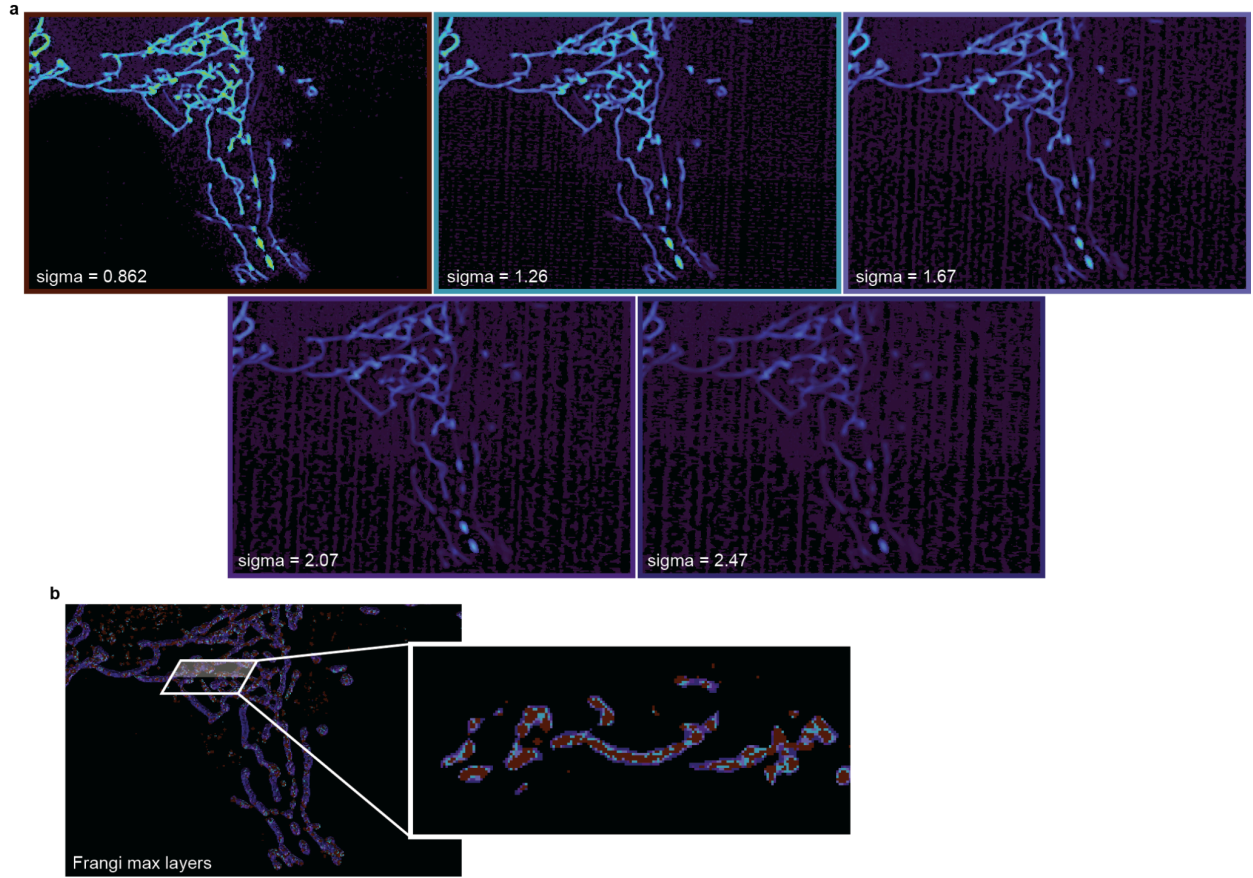

**Supplementary Data Fig. 1: Contributions of preprocessing multi-scale intermediates**
**a**, A 3D fluorescently mitochondrial-labeled cell run through a Frangi filter at 5 different scales, undergoing Gaussian filtration with respective sigma values. **b**, An integer-label image where the voxel color represents the corresponding Frangi filter scale contributor, as denoted by the border color.

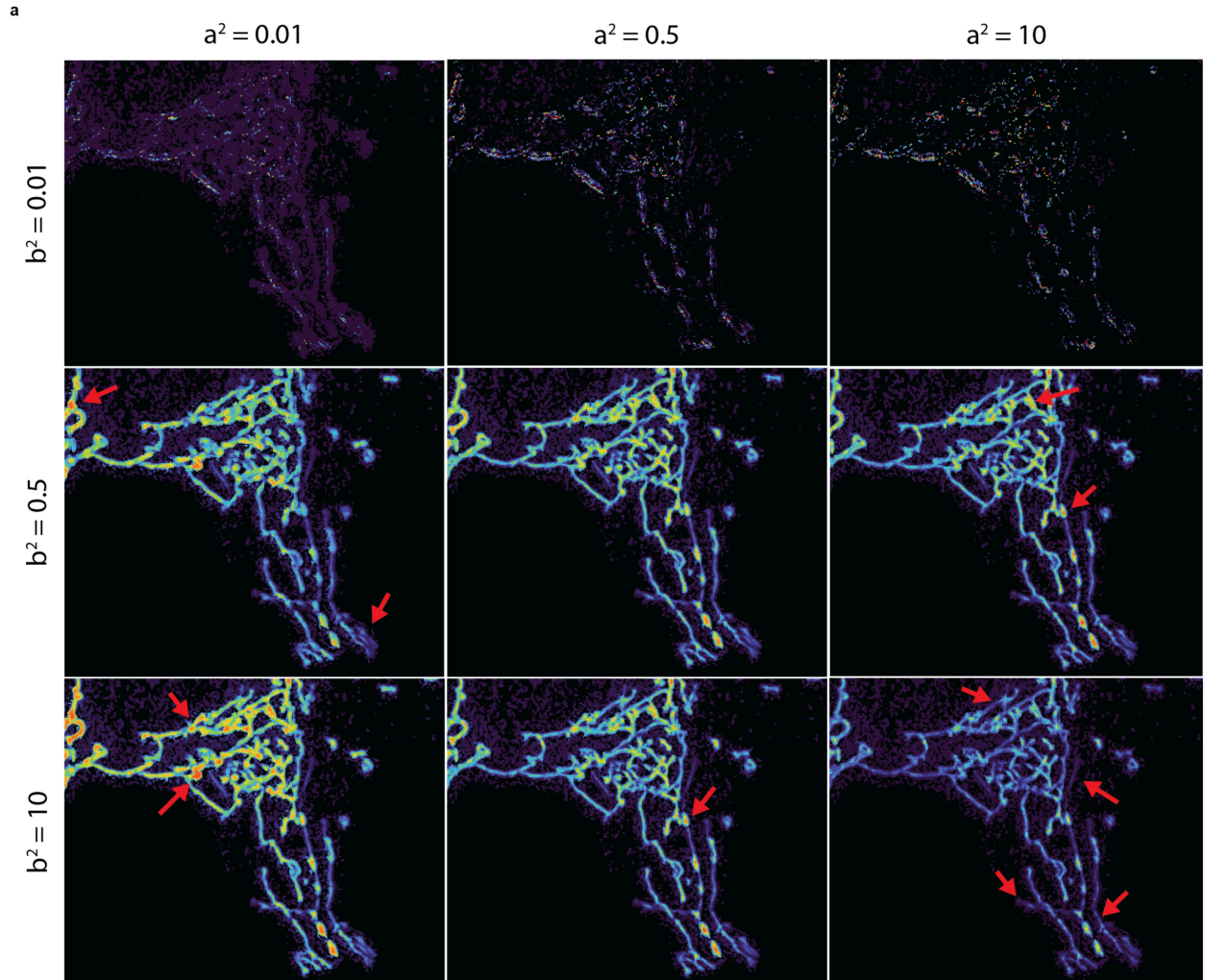

**Supplementary Data Fig. 2: Frangi filter parameter effects on 3D structural contrast enhancement**
**a**, A 3D fluorescently mitochondrial-labeled cell showcasing regions of tubular and blob-like structures. The outputs of the structural enhancement preprocessing using alpha and beta-squared Frangi filter parameters with alpha-squared ranging from 0.01 to 10 (columns), and beta-squared ranging from 0.01 to 10 (rows). The middle panel (0.5, 0.5) represents the hard-coded parameter set used in Nellie's pipeline. Red filled-in arrows show problematic regions.

**Supplementary Note 3: The Minotri threshold**

In this study, we introduce the Minotri threshold approach - a fusion of the Otsu and Triangle methods, adapted to the unique challenges of automated organelle segmentation. The Minotri threshold is defined as:

$$T_{minotri} = min(T_{tri}, T_{Otsu})$$

where $T_{tri}$ is the Triangle threshold and $T_{Otsu}$ is the Otsu threshold.

The Otsu method is well known for its capability to segment images based on a bimodal intensity distribution. It accomplishes this by calculating a threshold that maximizes the between-class variance, effectively distinguishing between the foreground (organelles) and the background:

Let the image have L gray levels [1, 2, ..., L]. The number of pixels at level i is denoted by n_i and the total number of pixels by N = n_1 + n_2 + ... + n_L.

Compute histogram and probabilities of each intensity level:

$$p_i = \frac{n_i}{N}$$

Compute cumulative sums for k = 0, 1, 2, ..., L-1:

$$P_1(k) = \sum_{i=1}^{k} p_i$$

Compute cumulative means:

$$m(k) = \sum_{i=1}^{k} \frac{i \cdot p_i}{P_1(k)}$$

Compute global intensity mean:

$$m_G = \sum_{i=1}^{L} i \cdot p_i$$

Compute between-class variance:

$$\sigma_B^2(k) = \frac{[m_G P_1(k) - m(k)]^2}{P_1(k)[1 - P_1(k)]}$$

The Otsu threshold is the value that maximizes between-class variance:

$$k^* = \text{argmax}_k \left( \sigma_B^2(k) \right)$$

Otsu's algorithm excels in scenarios where the histogram exhibits two distinct peaks, corresponding to these two classes. In such cases, it tends to favor the class with the larger peak, thus ensuring the majority of the pixels from this peak are classified correctly. However, its performance may be less optimal in images with skewed histograms or those lacking a clear bimodal distribution.

On the other hand, the Triangle method approaches thresholding from a geometric perspective. It identifies the threshold by drawing a line from the peak of the histogram to the farthest end and then finding the point on the histogram that is furthest from this line:

Compute the histogram of the image.

Find the peak $b_{max}$ of the histogram $h$.

Find the leftmost $b_{min}$ and rightmost $b_{max}$ non-zero bins of the histogram.

Compute the line equation passing through $(b_{min}, h[b_{min}])$ and $(b_{max}, h[b_{max}])$:

$$y = mx + c$$

where

$$m = \frac{h[b_{max}] - h[b_{min}]}{b_{max} - b_{min}}$$

$$c = h[b_{min}] - m \cdot b_{min}$$

For each bin i, compute the distance $d_i$ from $(i, h[i])$ to the line:

$$d_i = \frac{|h[i]| - (mi + c)}{\sqrt{m^2 + 1}}$$

The Triangle threshold is the bin i that maximizes $d_i$:

$$i^\star = argmax_i(d_i)$$

This method is particularly effective in dealing with histograms that are not bimodal or are skewed, such as in cases where there is a smooth transition from foreground intensity to background intensity, or when there is a large amount of diffuse background signal, whose intensity lies within a distribution higher than the background but lower than the foreground, making it a versatile tool for images where Otsu's method might struggle.

Our Minotri threshold approach combines these two methodologies by adopting the minimum (min) of the thresholds determined by the Otsu (ot) and Triangle (tri) methods. This approach ensures a more nuanced and adaptable segmentation, particularly useful in microscopy images where intensity distributions can vary significantly. By choosing the lower threshold, we err on the side of inclusivity, capturing a broader range of (both photon density-based and structurally based) intensities that represent the organelles, while still maintaining a rigorous standard for segmentation quality. This strategy is especially beneficial in cases where one method may yield a threshold that is too high, potentially excluding relevant structural information.

By integrating the Minotri threshold into our segmentation workflow, we achieve a balance between sensitivity and precision. This approach allows for the robust detection of structures across a wide range of intensity distributions, thereby enhancing the accuracy and reliability of our semantic segmentation process. This combined thresholding technique offers a versatile and effective tool for the detailed study of diverse organelles.

# Supplementary Note 4: Segmentation comparison and benchmarking against state-of-the-art organelle segmentation algorithms

|  | Nellie (ours) | Mitometer | MitoGraph v3 |
|---|---|---|---|
| **2D** | Yes | Yes | Yes |
| **3D** | Yes | Yes | Yes |
| **Temporal frames** | Optional | Required | Incompatible |
| **Metadata detection** | Yes | No | No |
| **Automated parameters** | Yes | Yes | No |
| **Node morphometrics** | Yes | No | Limited |
| **Branch morphometrics** | Yes | No | Limited |
| **Organelle morphometrics** | Yes | Yes | Limited |
| **Image-wide morphometrics** | Yes | No | No |
| **Mask output** | Yes | Yes | Limited |
| **Intermediates output** | Yes | Yes | Limited |
| **GUI** | Yes | Yes | No |
| **GPU-accelerated** | Yes | No | No |
| **Language** | Python | MATLAB | C++ |
| **Compatible OS** | Mac, Windows, Linux | Mac, Windows, Linux | Mac |

**Nellie, 2024** (https://github.com/aelefebv/nellie)

The subject of this paper, Nellie is an automated segmentation, tracking, and hierarchical feature extraction pipeline for organelles in both 2D and 3D live-cell microscopy. Nellie optionally takes a time series, though the segmentation does not require it. Nellie automatically detects the metadata of the image if it is present, but allows the user to modify the metadata if needed. Nellie returns morphometrics at the node, branch, organelle, and image-wide level in .csv format. Nellie also returns all intermediates, including the preprocessed frangi-filtered image, the skeleton mask, the tip, junction, and edge image, the branch-labeled skeleton image, the branch-labeled segmentation image, and the organelle-labeled segmentation image. Nellie has a Napari-based GUI, and uses GPU acceleration for segmentation. Nellie is fully written in Python, and is compatible with all major operating systems.

**Mitometer, 2021** (https://github.com/aelefebv/Mitometer)

Mitometer is an automated segmentation and tracking pipeline for mitochondria in both 2D and 3D live-cell microscopy, though still performs well for processing and analysis of other organelles. Mitometer requires temporal frames to perform segmentation, as it uses an inter-frame variability metric to determine smoothing and thresholding parameters. It requires the user to input image metadata such as the pixel resolution and the frame rate, and adjusts its segmentation pipeline accordingly. Mitometer's segmentation outputs are limited to connected-component analysis, but returns a wide variety of metrics in a .txt file. Mitometer also returns a time-series of the preprocessed diffuse background-removed image, and instance segmentations of individual mitochondria, as well as a masked version of the original fluorescence intensity image. Mitometer has a GUI for both single-file and batch-file processing, but does not use GPU acceleration. It uses MATLAB, which is not open source, but is compatible with all major operating systems. For the sake of reproducibility in running benchmarking and comparisons, we have rewritten Mitometer's MATLAB segmentation algorithm in Python, with GPU acceleration. The rewritten code is accessible in the supplemental Nellie repository (https://github.com/aelefebv/nellie-supplemental).

**MitoGraph v3, 2018** (https://github.com/vianamp/MitoGraph)

MitoGraph is an automated segmentation and analysis tool for 2D (since v3) and 3D mitochondrial images, though it works well for other organelles as well. Though the link to the code on the original webpage (https://rafelski.com/susanne/MitoGraph) no longer works, we assume that the first author's GitHub repository is a valid replacement. MitoGraph accepts, but does not segment images with multiple temporal frames. It also does not detect metadata automatically, and does not adjust the segmentation to the user-defined metadata (though it does scale the outputs accordingly). MitoGraph outputs a limited number of node, branch, and organelle morphometrics across various .mitograph, .cc, .txt, .gnet, and .coo formats. MitoGraph also outputs a 2D max projection of the post-segmentation image in .png format, as well as node, skeleton, and surface image intermediates in .vtk format. MitoGraph is written in C++, runs via the command line, and does not use GPU acceleration for segmentation. It is only compatible with MacOS. For the sake of reproducibility in running benchmarking and comparisons, we have written a batch conversion and processing script, adapted from

MitoTNT's scripts, as well as a method to reconstruct organellar objects by producing a sphere at each node position with its corresponding thickness, as we were unable to extract usable object masks for IoU and F1 segmentation quantification calculation without it. The scripts and methods are accessible in the supplemental Nellie repository (https://github.com/aelefebv/nellie-supplemental).

For the benchmarking and comparisons of the aforementioned segmentation algorithms, we chose to compare and contrast F1 scores to quantify accuracy in the detection of the correct objects, and the intersection over the union (IoU) to quantify the overlap of the algorithms' segmentation mask outputs and the ground truth object masks. As generating ground truth masks of 3D microscopy datasets is non-trivial and subject to user bias, we instead opt to build a suite of simulation tools to generate ground truth organelle-like objects. All of the simulation tools are made available in the supplemental Nellie repository (https://github.com/aelefebv/nellie-supplemental).

The first segmentation comparison tests each algorithm's ability to properly detect and segment objects of varying length, width, and intensities within the same image, considering organelles, even of the same type, may vary in size, length, and be tagged with intensity-varying functional readout markers. (Extended Data Fig. 2). To do this, we simulate a 3x510x510x510 (TZYX) OME-TIFF file. The dataset consists of 3 independently noise-sampled replicates of 10x10x10 simulated objects, which varies in thickness along Z, length along Y, and intensity along X. To simulate the object, we extend a single pixel line across Z, while simulating curvature in the Y and X axes. We then perform a binary dilation, followed by a skeletonization to ensure a proper skeleton object is formed. Finally, we create a sphere with a diameter of the corresponding thickness at each skeleton pixel coordinate to create the mask of the object, and populate that mask with the corresponding intensity. Each object is contained within a 50x50x50 voxel cube, all of which get stitched together into the final test dataset. We generate these datasets with increasing background noise values, generated via both additive Gaussian and Poisonian distributions (Extended Data Fig. 2c,d). We hard code a resolution of 0.1 um/pixel. We use as input the full 3x510x510x510 dataset consisting of 1000 objects. We use Nellie's "skel_relabelled" image, MitoGraph's reconstructed image, and Mitometer's post-size exclusion mask as outputs. We additionally include both an Otsu and Triangle thresholded output for baseline. To calculate the F1 and IoU scores, we separate the output and ground truth datasets into the corresponding single-object 50x50x50 voxel cubes. For each single-object cube, we compare each connected component to the ground truth object and keep as the final IoU the maximum IoU found for each individual object. If the IoU is above the minimum IoU threshold (0.1) we calculate F1 using the number of connected components in the output image. We then aggregate the scores for all 1000 objects to calculate our final scores. We find that Nellie outperforms all other methods in IoU score across all noise levels, and is similar to Mitometer's F1 score at low noise levels, but outperforms at mid to high noise levels (Extended Data Fig. 2a,b).

The second segmentation comparison tests each algorithm's ability to segment and detect objects across varying voxel resolutions, considering microscopy datasets are often acquired with widely varying parameters and microscope settings (Extended Data Fig. 2e,f). Objects are simulated as in the first test, except each dataset consists of only a single object, rather than a grid of objects. We generate and compare objects at um/px resolutions of 0.04, 0.06, 0.08, 0.1, 0.15, 0.2, 0.25, 0.3. For each pixel resolution, 4 objects are generated, a small (thickness of 0.4 um) and short (length of 0.4 um) object, a small and long (minimum of 49 pixels or 10 um) object, a thick (1 um) and short object, and a thick and long object, each dataset being replicated with increasing background noise values (Extended Data Fig. 2g). We calculate F1 score and IoU as before, and average the metrics across all resolutions and object sizes to generate the final quantifications (Extended Data Fig. 2e,f). As expected, a simple Otsu threshold performs best for single objects with a constant intensity against low background noise, but quickly decays, whereas Nelly performs steadily well across all noise values. MitoGraph performs consistently well, though not as well as Nellie, across all noises, except the highest in which it could not detect the object of interest. Nellie's F1 score was either better or comparable to other algorithms across all noise levels.

The final segmentation comparison tests each algorithm's ability to correctly detect and segment two adjacent objects, considering many datasets often have numerous objects of interest within closely confined regions (Extended Data Fig. 2h,i). Objects are simulated as in the second test, except no curvature is used. We again generate and compare objects at um/px resolutions of 0.04, 0.06, 0.08, 0.1, 0.15, 0.2, 0.25, 0.3, and for the 4 object dimensions as described in the second test. We additionally add in a second line during the initial skeleton population phase, separated in the X dimension by a separation distance of 2, 2.25, and 2.5 times the thickness (Extended Data Fig. 2j). We calculate F1 score and IoU as the second test (Extended Data Fig. 2h,i). IoU and F1 performance was roughly the same for all algorithms as the second test case, with Nellie again performing steadily well throughout, though Mitograph's F1 score is slightly higher at medium-high noise levels, but drops again at the highest noise level.

**Supplementary Note 5: The Swin UNETR Deep Learning Model Architecture for Organelle Segmentation in Fluorescence Microscopy Images**

Here, we provide a comprehensive description of our custom-trained **Swin U-Net Tr**ansformer (Swin UNETR) models used for organelle segmentation in fluorescence microscopy images. Originally created for medical image segmentation tasks, the model leverages the Swin Transformer to capture long-range dependencies and the U-Net architecture for precise localization, making it well-suited for both 2D and 3D fluorescence microscopy image segmentation tasks[49]. The primary goal of this section is to evaluate the generalization capabilities of deep learning based organelle segmentation models when trained on limited and specific datasets, such as a single organelle type, or a single microscopy experimental setup.

We utilized the public dataset from the Allen Institute for Cell Science (AICS), from their earlier publication on label-free prediction of three-dimensional fluorescence images, where all the images we included in our training were taken from the same microscope with the same imaging settings[66]. This dataset comprises a collection of high-quality fluorescence microscopy images, each focusing on a specific cellular structure or organelle. For our experiments, we trained models on individual organelle datasets, a mitochondria-only dataset and a desmosomes-only dataset, to generate organelle specific models, as well as a combined dataset including multiple structures such as actin, actomyosin, desmosomes, endoplasmic reticulum, Golgi apparatus, mitochondria, nucleoli, tight junctions, and tubulin, to generate a microscope specific, but structure agnostic model. Each organelle-specific dataset contains 80 images, providing a consistent set for training and validation.

Due to limited approaches for generating 3D manual segmentations at scale, and limited publicly available fluorescence-matched 3D segmentation datasets, we employed Nellie to generate ground truth segmentation masks for training (Supplementary Figure 4). While Nellie provides reliable organelle segmentation results, using its outputs as ground truth inherently limits the deep learning models to attempt to replicate Nellie's segmentation capabilities. However, as the objective of this study is specifically to evaluate the generalization capabilities of these models, rather than compare their outputs directly to Nellie, we deemed this limitation to be acceptable.

The Swin UNETR model combines the strengths of the Swin Transformer architecture with the U-Net segmentation framework, aiming to capture both global context and fine-grained details essential for accurate segmentation. Our model specifically processes 3D input patches (overlapping chunks of the original full data) of size (32, 128, 128) voxels, corresponding to the Z, Y, and X axes, and accepts single-channel grayscale images, outputting single-channel binary segmentation masks. A base feature size of 48 was selected to balance model complexity and computational efficiency.

Training was conducted separately for ten models: nine trained exclusively on individual structures–mitochondria, desmosomes, actin, actomyosin, endoplasmic reticulum, Golgi apparatus, nucleoli, tight junctions, and tubulin–two of which we showcase within this paper, and a "combo" model trained on the combined dataset of multiple organelles (Supplementary Figure 3a). Each model was trained for 100 epochs with a batch size of 1, due to memory constraints inherent in processing 3D data, and validation scores were calculated every 5 epochs (Supplementary Figure 3b). We used a composite loss function combining Dice Loss and Cross-Entropy Loss to address class imbalance and encourage accurate voxel-wise classification.

Data augmentation techniques are crucial in attempting to enhance the models' robustness and generalizability, given the limited size of each organelle dataset and confinement to a single microscopy setup. Augmentations included random flipping along spatial axes, random rotations, intensity normalization, padding, and cropping to standardize input dimensions.

Upon evaluation, all three models demonstrated strong performance on test data within the AICS dataset, accurately segmenting the structures they were trained on, as compared to Nellie's "ground truth" outputs (Supplementary Figure 4a). However, when applied to fluorescence microscopy images from different sources, microscopes, or resolutions not represented in the training data, including 2D images with padding, the models exhibited significant performance degradation and segmentation artifacts were prominent (Supplementary Figure 4b,c). Additionally, even within the AICS dataset, the mitochondria-trained model appeared to oversegment non-tubular structures such as desmosomes, and similarly, the desmosome-trained model appeared to undersegment non-spherical structures, such as mitochondria and golgi. The combo model showed some improvement across multiple structures within the combined dataset but still seemed to oversegment spherical structures (possibly due to a class imbalance issue), and struggled with data outside the AICS's microscopy setup.

These results highlight a critical limitation of deep learning models in the context of organelle segmentation: their generalization capabilities are heavily dependent on the quantity, diversity, range, and balance of the training data. When trained on datasets lacking sufficient variability in imaging conditions, organelle types, and resolutions, the models become specialized tools that perform well only within their narrow training domain.

In contrast, Nellie, being a traditional image processing algorithm, does not rely on learning from data to generalize. It applies consistent computational methods to enhance and segment structures across various datasets, imaging modalities, and resolutions. Our experiments quantitatively demonstrated that Nellie maintained robust segmentation performance when applied to simulation datasets of structures with various shapes and sizes, and qualitatively when applied to images from different microscopes, resolutions, cell types, and organelles. This consistency underscores the current advantage of traditional algorithms in scenarios where acquiring diverse and extensive training data is impractical.

The key takeaway from this study is the trade-off between specialized deep learning models and generalist algorithms like Nellie. While deep learning models have the potential to learn complex features and achieve high accuracy within their training domain, their effectiveness diminishes when faced with data that falls outside the scope of their training. This limitation poses a significant challenge in fields like fluorescence microscopy, where imaging conditions can vary widely, and obtaining large, diverse annotated datasets is often difficult. Nellie, on the other hand, provides reliable segmentation without the need for retraining or fine-tuning, making it a valuable tool for researchers working with varied and heterogeneous datasets, especially in 3D.

Importantly, Nellie may serve as an automated way of generating ground truth segmentation masks for training, either from its raw segmentation outputs as we do here, or from a manually reinforced version of its outputs. This will hopefully lead to a deep learning model that eventually outperforms Nellie in generalizability and segmentation accuracy.

**Swin UNETR Model Parameters and Training Details**

In this section, we provide a detailed description of the model parameters and training configurations used in our experiments, as implemented in our code. This detailed exposition aims to facilitate replication of our work and provide insights into the parameter choices made for the Swin UNETR model in the context of organelle segmentation in fluorescence microscopy images.

The Swin UNETR model was implemented using the SwinUNETR class from the MONAI library[67]. This model architecture effectively combines Swin Transformer blocks with a U-Net-like decoder, allowing it to capture both local and global contextual information, which is useful for accurate segmentation tasks in 3D images.

The model expects input data of any size, from 2D up to 5D data, so long as the ZYX axes are the last three dimensions. The code provided automatically selects the first slice of the non-ZYX dimensions.

The image is subsampled to chunks of dimension 32x128x128. This specific size was chosen based on the dimensions of the organelle structures in our fluorescence microscopy images and to ensure compatibility with the Swin Transformer architecture, which requires input dimensions to be divisible by the patch size and window size.

Since the fluorescence microscopy images are grayscale (single-channel), representing the intensity of fluorescence from the organelle-specific dyes, the input channel number is set to 1. The model outputs a single-channel probability map, indicating the likelihood of each voxel belonging to the organelle of interest, suitable for binary segmentation tasks, thus the output channel number is set to 1 as well. The base number of feature maps in the model is set to 48. This parameter controls the width of the network and was selected to balance between sufficient model capacity to capture complex features and computational efficiency to manage GPU memory constraints.

Other notable parameters of the SwinUNETR model were left at their default values, which are optimized for general medical image segmentation tasks. The parameters that were not explicitly modified include:

- Depths (depths): (2, 2, 2, 2)
    - Specifies the number of Swin Transformer layers at each stage of the encoder. Each number corresponds to a stage in the hierarchical architecture.
- Number of Heads (num_heads): (3, 6, 12, 24)
    - Defines the number of attention heads in the multi-head self-attention mechanism for each stage, corresponding to the depths.
- Patch Size (patch_size): (2, 4, 4)
    - Determines the size of the patches into which the input images are divided before being processed by the Swin Transformer blocks.
- Window Size (window_size): (7, 7, 7)
    - Sets the size of the local windows used in the self-attention computations within the Swin Transformer layers.

Although some of these parameters were considered for adjustment to potentially improve performance and reduce computational load, we retained the default settings after preliminary experiments indicated that they provided a good balance between accuracy and efficiency for our datasets.

We employed the DiceCELoss loss function from the MONAI library, which combines Dice Loss and Cross-Entropy Loss. Dice Loss measures the overlap between the predicted segmentation and the ground truth, which is particularly useful for addressing class imbalance in segmentation tasks. Cross-Entropy Loss evaluates the voxel-wise classification accuracy, penalizing incorrect predictions more heavily.

The combination of these two loss functions helps the model focus on both the overall shape and the precise boundaries of the organelles. The sigmoid=True parameter ensures that a sigmoid activation function is applied to the model's outputs before computing the loss, converting logits to probabilities suitable for binary classification.

We used AdamW optimizer from the PyTorch library for training, which is a variant of the Adam optimizer that includes weight decay regularization. A relatively low learning rate of 1e-4 was chosen to allow the model to converge steadily without overshooting minima. Due to the high memory requirements of processing 3D volumetric data and the complex architecture of the Swin UNETR model, we used a batch size of 1 to fit the data and model into the available GPU memory. Each model was trained for 100 epochs to ensure sufficient training iterations for the model to learn from the data. The model was evaluated on the validation set every 5 epochs to monitor its performance and to detect potential overfitting early, but as the validation process takes a substantial amount of time, we decided against a shorter validation period.

Data augmentation was performed using a series of transformations from the MONAI library to enhance the model's ability to generalize from limited data:

- Ensure Channel First (EnsureChannelFirstd)
    - Ensured that the input data has the channel dimension as the first dimension, as required by PyTorch models.
- Intensity Scaling (ScaleIntensityd)
    - Scaled the intensity values of the images to the range [0, 1], normalizing the data and helping the model to converge during training.
- Padding (SpatialPadd)
    - Applied symmetric padding to the images and labels to reach the desired spatial size (32, 128, 128). Padding was necessary because the images might not have uniform sizes and the Swin Transformer architecture requires inputs of specific dimensions.
- Random Spatial Cropping (RandSpatialCropd)
    - Extracted random crops of size (32, 128, 128) from the padded images and labels. This step helps the model learn from different regions of the images, increasing the diversity of the training data.
- Random Flipping (RandFlipd)
    - Randomly flipped the images and labels along each spatial axis (Z, Y, X) with a probability of 0.5. Flipping helps the model become invariant to the orientation of the organelles.
- Random Rotation (RandRotate90d)

- Applied random rotations of the images and labels by 90 degrees up to three times (max_k=3) with a probability of 0.5. Rotations further enhance the model's ability to recognize organelles in different orientations.
- Conversion to Tensors (ToTensord)
    - Converted the numpy arrays into PyTorch tensors, which are required for model input.

For validation and inference, we used the sliding_window_inference function from MONAI to handle large images that could not be processed in a single forward pass due to memory constraints
- ROI Size (roi_size): (32, 128, 128)
    - Matches the input size expected by the model.
- Sliding Window Batch Size (sw_batch_size): 1
    - Processes one window at a time during inference.
- Overlap (overlap): 0.25
    - Specifies the amount of overlap between adjacent windows to ensure seamless predictions across the entire volume.

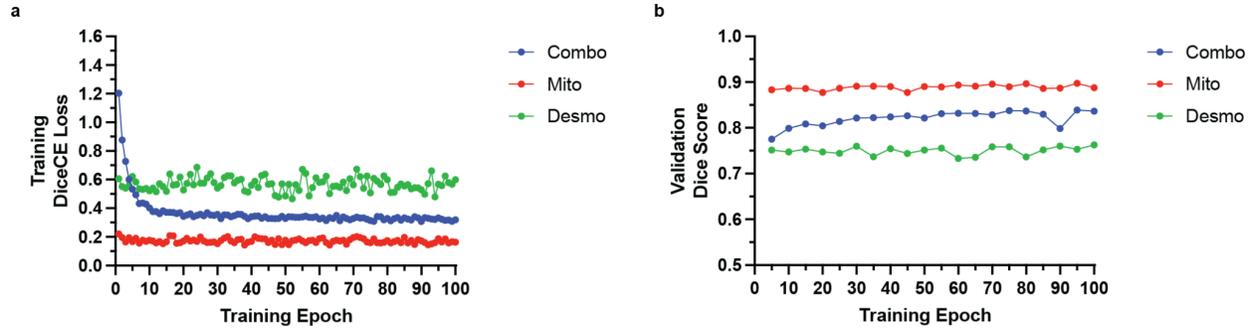

**Supplementary Data Fig. 3: Training and validation metrics for Swin UNETR segmentation models**
**a**, DiceCE loss values for training datasets of Swin UNETR models trained with datasets containing actin, actomyosin, desmosomes, endoplasmic reticulum, Golgi apparatus, mitochondria, nucleoli, tight junctions, and tubulin (blue), mitochondria alone (red) or desmosomes alone (green) at each epoch. **b**, Dice scores for training-excluded validation datasets of Swin UNETR models trained on the same respective organelle as in **a**.

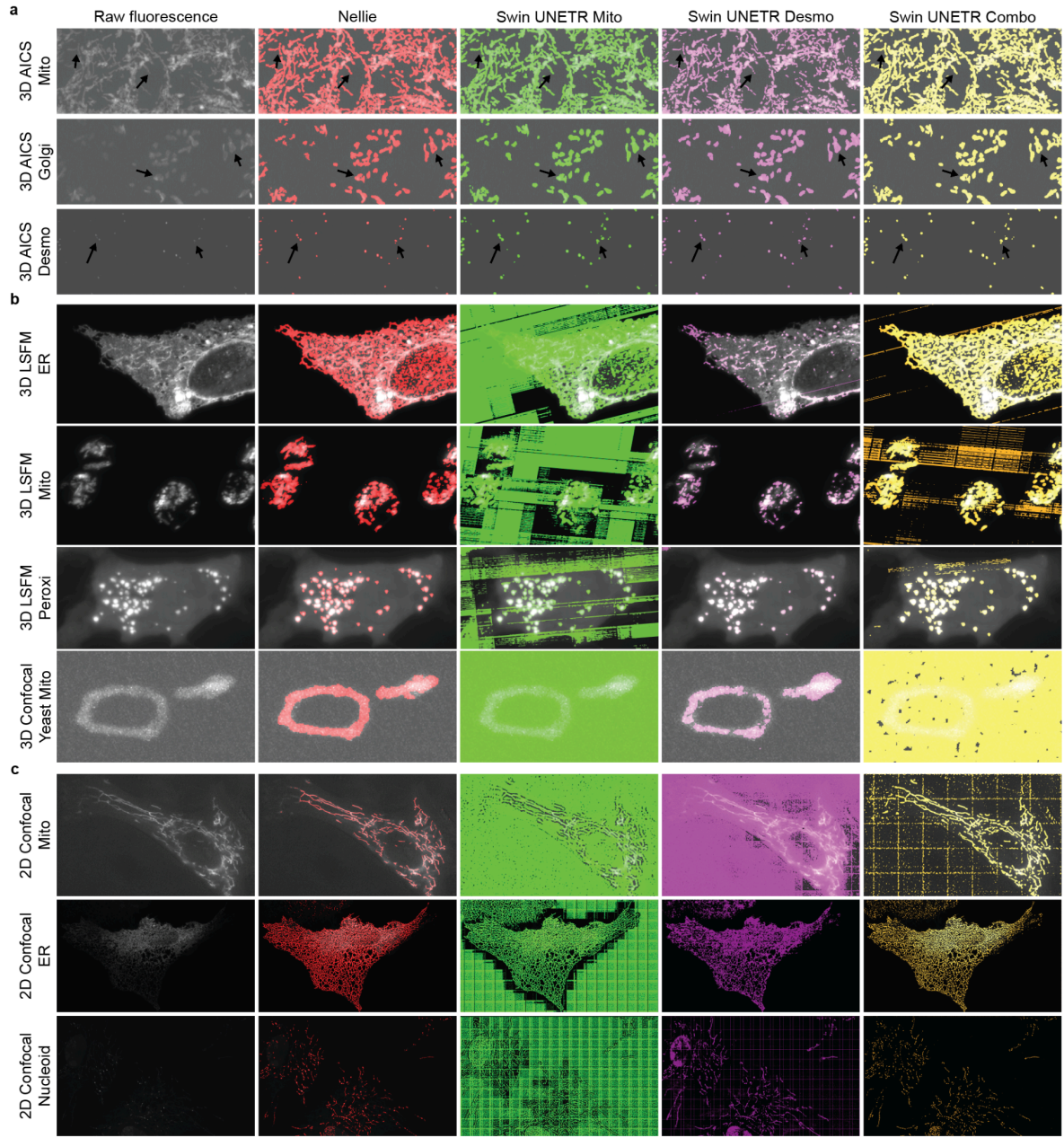

**Supplementary Data Fig. 4: Visualizations of segmentations from Nellie and custom trained Swin UNETR models**

Segmentation outputs from Nellie (red) and Swin UNETR models pre-trained on only mitochondria (green), only desmosomes (purple) and an aggregated dataset of actin, actomyosin, desmosomes, endoplasmic reticulum, Golgi apparatus, mitochondria, nucleoli, tight junctions, and tubulin (orange), with the raw fluorescence data shown in the left-most column. **a**, Output segmentations for the 3D dataset from the Allen Institute for Cell Science (AICS) for mitochondria, golgi, and desmosomes. Black arrows denote regions of notably different segmentation results between methods and models. **b**, Output segmentations for 3D datasets outside of the AICS dataset, including lightsheet fluorescence microscopy (LSFM) data of endoplasmic reticulum (ER), mitochondria, and peroxisomes, and confocal microscopy data of yeast cell mitochondria. **c**, Output segmentations for 2D datasets outside of the AICS dataset, including confocal microscopy datasets from mitochondria, ER, and nucleoids.

**Supplementary Note 6: Nellie's extended segmentation pipeline**

The segmentation process for organelles is a simple connected-components labeling scheme. Subsequently, Nellie incorporates a binary hole-filling algorithm exclusively in the analysis of 3D images. This decision is informed by distinct considerations regarding the structural characteristics and noise levels inherent to 2D versus 3D datasets.

One of the primary reasons for omitting the hole-filling algorithm in 2D segmentation is the inherently lower level of structural noise within these datasets. The simplified nature of 2D imaging, capturing a single plane of focus, generally results in clearer delineation of organelles. This clarity reduces the likelihood of generating the types of internal inconsistencies within segmented organelles that the hole-filling algorithm is designed to address.

The decision to apply the hole-filling algorithm selectively also stems from the geometric considerations of organelles across dimensions. In 2D images, closed circular organelles are common, and applying hole-filling indiscriminately would lead to the undesired effect of filling these circles, potentially obscuring meaningful biological cavities. While the algorithm aids in creating a more consistent representation of such organelles by addressing internal voids, it operates under the assumption that spaces within the segmented organelles are artifacts rather than biologically relevant features. A notable caveat in the application of the hole-filling algorithm within 3D segmentation concerns the potential for erroneous interpretation of certain organelle geometries. Specifically, if an organelle inherently forms a spherical shape with a central void

(resembling a ball with an internal hole), the algorithm may inappropriately fill this void. This limitation underscores a challenge in distinguishing between genuine biological internalized cavities and segmentation artifacts, particularly in the context of 3D structural complexity. If spherical shapes with central voids are important for the user's analysis, we suggest replacing this binary hole filling with a binary closing operation below the scale of the object of interest.

Following thresholding, a binary opening operation is conducted to smooth out noisy or irregular boundaries of the organelles and enhance the accuracy of the subsequent segmentation by removing weak, single-voxel connections. We then use an octree-based (or quadtree in 2D) skeletonization to reduce the semantic segmentation masks to single-voxel wide representations[68,69].

**Supplementary Note 7: Nellie's mocap marker generation pipeline via multi-scale and adaptive local maxima detection**

To generate mocap markers, Nellie first performs a computationally efficient distance transformation over our semantic segmentation mask[70] (Supplementary Data Fig. 5, Supplementary Note 8). The distance-transformed image undergoes filtering using a Laplacian of Gaussian (LoG) filter, incorporating a set of sigma values identical to those utilized in the preprocessing pipeline. Let $I(x, y, z)$ be the input image and $L_\sigma(x, y, z)$ be the LoG filtered image at scale $\sigma$:

$$L_\sigma(x, y, z) = -\sigma^2 \cdot \Delta\left[G_\sigma(x, y, z) * I(x, y, z)\right]$$

where $G_\sigma$ is the 3D Gaussian kernel with standard deviation $\sigma$, $*$ denotes convolution, and $\Delta$ is the Laplacian operator.

The LoG filter, by design, enhances regions of rapid intensity change, which are indicative of edges and, by extension, potential peak points within the organelles. The application of this filter across multiple scales allows for the detection of peaks that correspond to features of varying sizes, ensuring comprehensive coverage of the intracellular structural landscape (Supplementary Data Fig. 6). Let R(x,y,z) be the multi-scale LoG response.

$$R(x, y, z) = max\left[L_\sigma(x, y, z)\right]$$

$$\text{for all } \sigma \text{ in } S$$

where $S$ is the set of scale values used.

Following LoG filtering, a maximum filter is applied to the resultant multi-scale filtered images. This process involves comparing each voxel to its neighbors within a defined footprint, retaining only those voxels that represent the highest value. By conducting this operation across the stacked LoG filtered images, voxels that retain their maximal value in the corresponding volume slice of the LoG filtered image are classified as local maxima for that specific scale. Let P(x,y,z) be the binary peak map.

$$P(x, y, z) = 1 \text{ if } R(x, y, z) = max[R(x', y', z')]$$

$$\text{for all } (x', y', z') \text{ in } N(x, y, z)$$

$$P(x, y, z) = 0 \text{ otherwise}$$

where $N(x, y, z)$ is the neighborhood of voxel (x,y,z).

The identified local maxima across all scales are then compiled into a single stack of images. This stack is subsequently flattened, consolidating the peak points into a singular image that serves as the basis for motion capture marker identification. This flattened image, rich in peak coordinates, effectively highlights the most prominent features within the organelles, earmarking them for feature extraction and downstream temporal linkage.

Given the potential for closely situated peaks to complicate (due to redundancy) rather than enhance subsequent analyses, an additional refinement step is employed. This step involves assessing the intensity of each peak within the original image and applying a proximity-based selection criterion. Peaks deemed too close to one another are subjected to a comparison of their respective intensities, with preference given to the peak of higher intensity. Let Q be the set of refined peak coordinates.

$$Q = \{(x, y, z) \ in \ P : I(x, y, z) = max[I(x', y', z')] \ for \ all \ (x', y', z') \ in \ B_d(x, y, z)\}$$

where $B_d(x, y, z)$ is the set of peak coordinates within distance d of (x,y,z).

This is facilitated by the construction of a k-d tree from the sorted peak coordinates, enabling efficient nearest-neighbor searches to identify and eliminate proximal peaks that fall within a predefined minimum distance threshold.

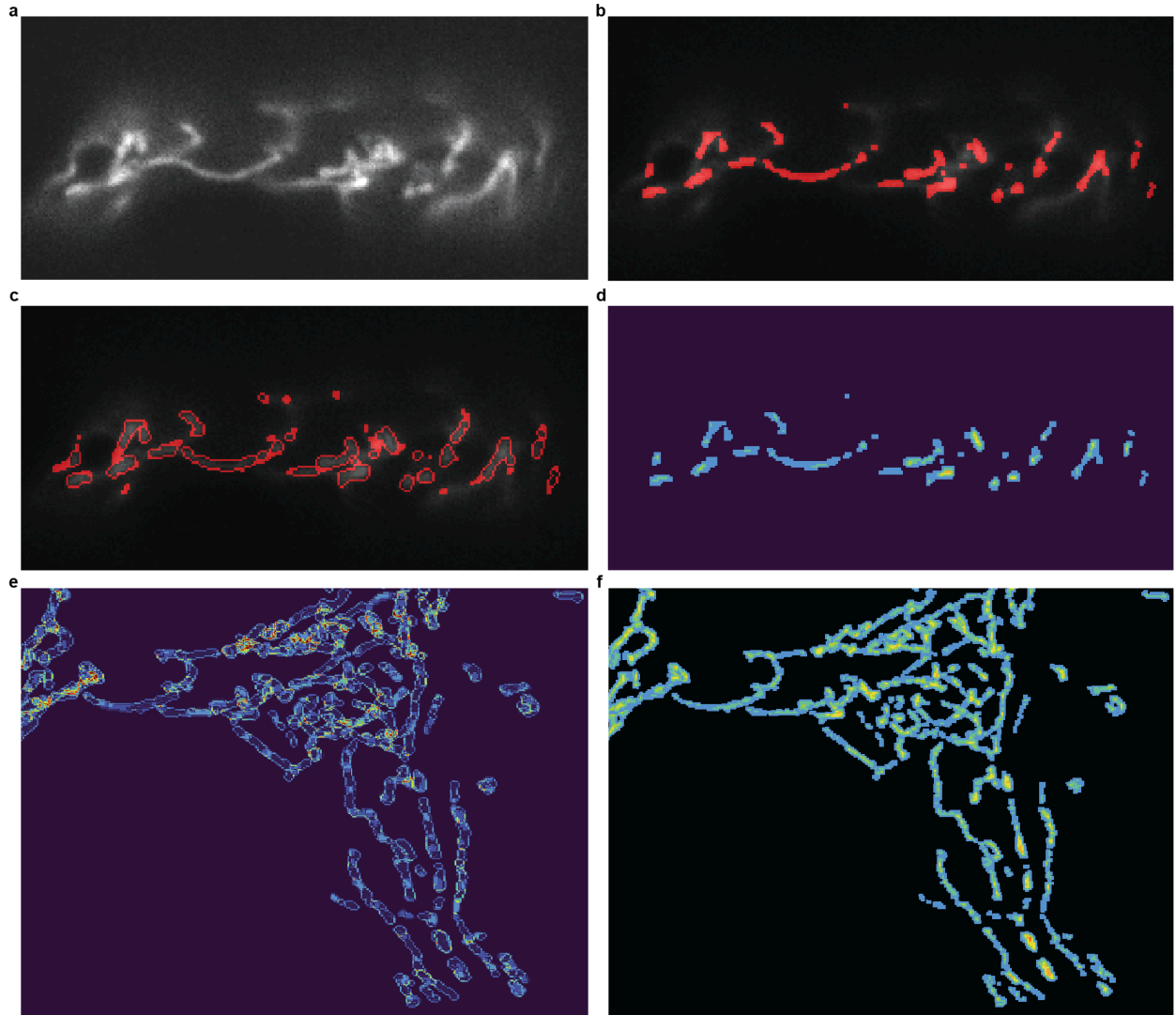

**Supplementary Data Fig. 5: Computationally efficient distance transformation using segmentation border coordinate k-d trees**

**a**, A 2D XY orthogonal slice of a 3D fluorescently mitochondrial-labeled cell from a single object lightsheet microscope (SOLS). **b**, The same 2D slice as in (**a**) of the instance segmentation of the 3D dataset. **c**, The border mask of the instance segmentation of (**b**) superimposed on (**a**), obtained via a one-voxel binary dilation of the instance segmentation mask, followed by subtraction of the instance segmentation mask. **d**, The same 2D slice as in (**a**) of the distance transformation image acquired by calculating the nearest neighbor distance of the voxel coordinates in (**b**) to the voxel coordinates in (**c**). **e**, The mean intensity projection of the 3D border mask of the instance segmentation. **f**, The max intensity projection of the 3D distance transformation image of the instance segmentation.

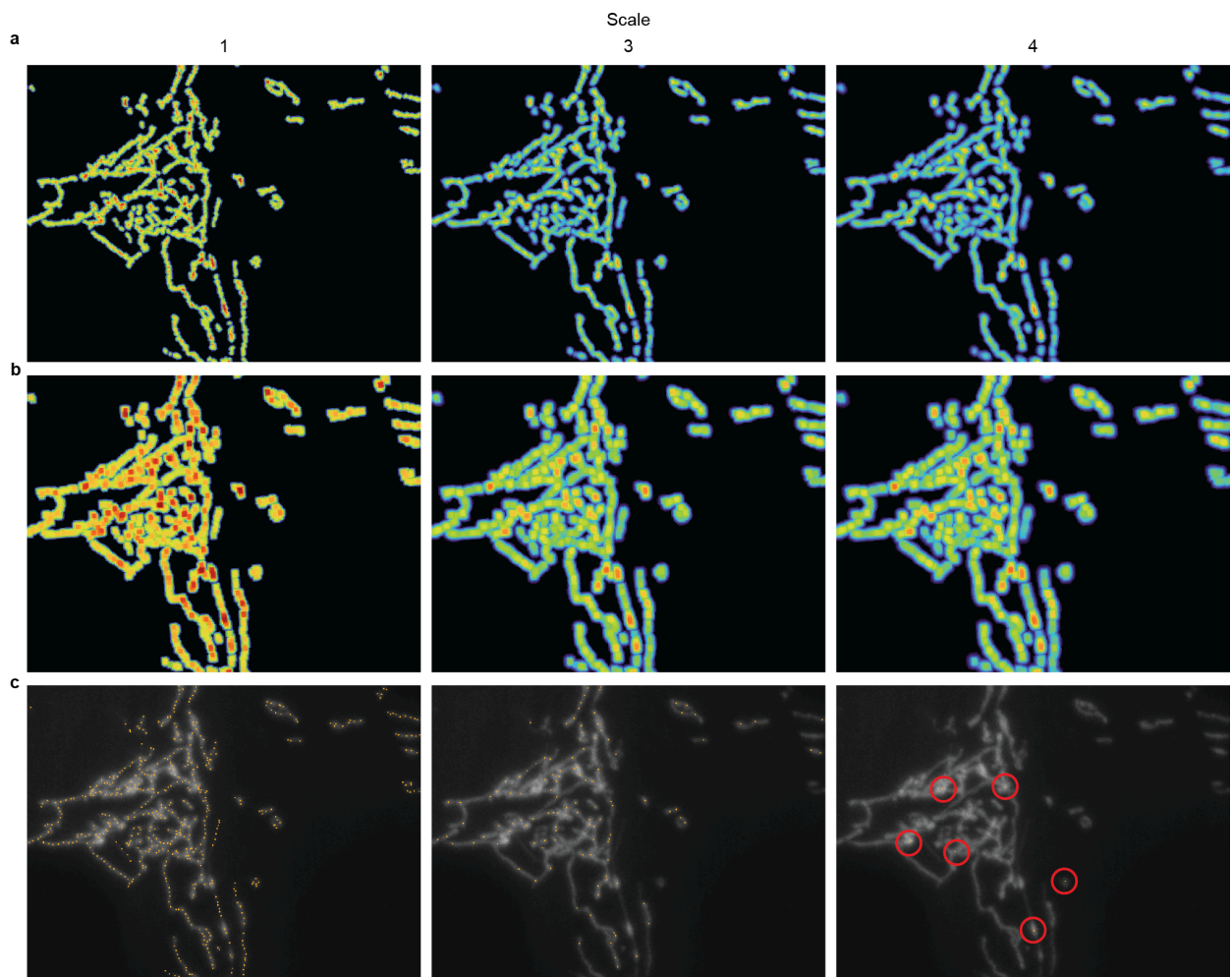

**Supplementary Data Fig. 6: Multi-scale local maxima peak detection for mocap marking**
**a**, A 3D fluorescently mitochondrial-labeled cell's post-processing volume is run through a Laplacian of Gaussian filter at different scales. Shown here is a filter sigma of 0.862, 1.67, and 2.07, (left, middle, right) corresponding to scales 1, 3, and 4, of 5 total scales. **b**, A maximum filter is run across the 4D stack of Laplacian of Gaussian filtered images (**a**). Shown here are again images corresponding to scales 1, 3, and 4, of 5 total scales. **c**, The raw fluorescence intensity image overlaid with single orange voxels corresponding to locations where the intensity value of the Laplacian of Gaussian image (**a**) is equal to the intensity value of the maximum filtered image (**b**) at different scales, again shown at scales 1, 3, and 4, of 5 total scales. Red circles are to indicate the few voxels at scale 4 where voxel intensity values match, indicating fewer (though still important) regions of larger scale structures.

**Supplementary Note 8: Efficient distance transformation using k-dimensional trees**

In this section, we introduce, to our knowledge, the first application of a k-dimensional (k-d) tree for a computationally efficient distance transformation of a binary mask. The process begins with the generation of a border mask, which delineates the periphery of the organelles within the binary mask. This is accomplished by subtracting the original binary mask from its one-voxel binary dilation. Binary dilation incrementally expands the boundaries of the organelle. Let M be the original binary mask and D be the dilated mask.

$$D = M \oplus S$$

where $\oplus$ denotes morphological dilation and S is the structuring element. Let B be the border mask.

$$B = D - M$$

where $-$ denotes set difference (subtraction).

The subtraction operation thus isolates a thin border that precisely maps the outermost edge of each organelle, effectively creating a mask that represents the interface between the organelle and its surrounding environment:

Following the creation of the border mask, the next step involves constructing a k-d tree from the coordinates of the border pixels. The k-d tree is a space-partitioning data structure optimized for organizing points in a k-d space, facilitating efficient queries for nearest neighbors. In the context of our method, the k-d tree serves as a computational framework that allows for rapid identification of the closest border pixel for any given voxel within the instance segmentation mask of organelles. Let T be the k-d tree and $C_B$ be the set of coordinates of border pixels.

$$T = KDTree(C_B)$$

where KDTree() is the function to construct a k-d tree.

With the k-d tree in place, the final step is to perform a distance transformation on the binary mask. This is achieved by querying the k-d tree with the coordinates of all voxels in the instance mask. For each voxel, the k-d tree identifies the nearest border pixel and calculates the Euclidean distance between these two points. This process is repeated for every voxel in the mask, resulting in a distance-transformed image where the value of each voxel reflects its minimum distance to the organelle's boundary. Let DT be the distance-transformed image and $C_M$ be the set of coordinates of mask pixels.

For each point p in $C_M$:

$$nearest\_border = T.query(p, k=1)$$

$$DT(p) = EuclideanDistance(p, nearest\_border)$$

where T.query() finds the nearest neighbor in the k-d tree, and EuclideanDistance() calculates the Euclidean distance between two points. For points p1(x1, y1, z1) and p2(x2, y2, z2):

$$EuclideanDistance(p1, p2) = \sqrt{(x_2 - x_1)^2 + (y_2 - y_1)^2 + (z_2 - z_1)^2}$$

By employing a k-d tree for the nearest-neighbor search, our approach minimizes computational overhead, enabling rapid processing of large datasets without sacrificing accuracy.

# Supplementary Note 9: Tracking comparison and benchmarking against state-of-the-art organelle tracking algorithms

|  | Nellie (ours) | MitoTNT | Mitometer |
|---|---|---|---|
| **2D** | Yes | No | Yes |
| **3D** | Yes | Yes | Yes |
| **Metadata detection** | Yes | No | No |
| **Automated parameters** | Yes | Yes | Yes |
| **Voxel motility metrics** | Yes | No | No |
| **Node motility metrics** | Yes | Limited | No |
| **Branch motility metrics** | Yes | Limited | No |
| **Organelle motility metrics** | Yes | Limited | Limited |
| **Image-wide motility metrics** | Yes | No | No |
| **Object permanence** | Yes | Limited | Limited |
| **Track output** | Yes | Yes | Limited |
| **Intermediates output** | Yes | No | Limited |
| **Visualization** | Yes | Limited | Yes |
| **GUI** | Yes | No | Yes |
| **Language** | Python | Python | MATLAB |
| **Compatible OS** | Mac, Windows, Linux | Mac | Mac, Windows, Linux |

**Nellie, 2024** (https://github.com/aelefebv/nellie)

The subject of this paper, Nellie, is an automated segmentation, tracking, and hierarchical feature extraction pipeline for organelles in both 2D and 3D live-cell microscopy. Nellie uses its segmentation pipelines' outputs and intermediates to perform tracking. Nellie automatically detects the tracking-based metadata of the image if it is present, but allows the user to modify the metadata if needed. Nellie returns motility metrics at the voxel, node, branch, organelle, and image-wide level in .csv format. Nellie allows for maintaining object permanence (down to the sub-voxel level) using flow interpolation, meaning one can keep track of a specific voxel, node, branch, or object starting from and up to any time point. Nellie also returns all intermediates, including the distance, border, mocap marker, and frame 1 object permanence images in .ome.tif format, as well as adjacency maps for connecting adjacent frame voxels, nodes, branches, and objects in .pkl format, and mocap marker flow vector arrays in .npy format. Nellie has a Napari-based GUI, is also fully written in Python, and is compatible with all major operating systems.

**MitoTNT, 2023** (https://github.com/pylattice/MitoTNT)

MitoTNT is an automated tracking, visualization, and analysis tool for 3D mitochondrial timelapse images, though it works well for other organelles as well. MitoTNT does not automatically detect the image's metadata, but does tune the tracking in response to the user-defined metadata. Though MitoTNT does not produce voxel-level outputs, it allows for the extraction of node, branch, and organelle-level motility metrics; however these outputs are limited to linear motion, such as diffusivity, using the mean squared displacement. There is limited object permanence via a sliding-window-based graph restructuring analysis for fission and fusion detection, allowing one to link nodes and objects between frames. MitoTNT outputs a .csv listing each nodes' tracks. There are also options of converting the node tracks into fragment and segment tracks using max pooling, though we had limited success in an errorless conversion. MitoTNT does not output intermediate validation files, but does output files that allow for track and dynamics visualization via ChimeraX, a free, open-source program from UCSF (https://github.com/RBVI/ChimeraX). MitoTNT itself does not have a GUI. It is written in python, with example Jupyter notebook use cases. MitoTNT is only compatible with MacOS. To allow for an automated experience using MitoTNT, we have generated scripts for the automated conversion of timelapse images, tracking, and fission and fusion detection adapted from MitoTNT's Jupyter notebooks. The scripts and methods are accessible in the supplemental Nellie repository (https://github.com/aelefebv/nellie-supplemental).

**Mitometer, 2021** (https://github.com/aelefebv/Mitometer)

Mitometer is an automated segmentation and tracking pipeline for mitochondria in both 2D and 3D live-cell microscopy, though still performs well for processing and analysis of other organelles. Mitometer uses its segmentation outputs to perform tracking. It requires the user to input temporal metadata and adjusts its tracking pipeline accordingly. Mitometer's tracking outputs are limited to connected-component center-of-mass analysis, and returns corresponding linear motion-based quantifications in a .txt file. Similar to MitoTNT, Mitometer has limited object permanence via fission and fusion detection and subsequent object-based linkage. Mitometer has a GUI for both single-file and batch-file processing, but uses MATLAB, which is not open source, but is compatible with all major operating systems. For the sake of reproducibility in running benchmarking and comparisons, we have rewritten Mitometer's MATLAB tracking and fission and fusion algorithms in Python. The rewritten code is accessible in the supplemental Nellie repository (https://github.com/aelefebv/nellie-supplemental).

For the benchmarking and comparisons of the aforementioned tracking algorithms, we chose to compare and contrast metrics such as linear and angular velocity, track length, and fission and fusion event detection via 5 different scenarios. As generating ground truth masks and tracks of 3D + T microscopy datasets is non-trivial and subject to user bias, we instead opt to build a suite of motility simulation tools to complement the object simulation tools in the segmentation comparison section to generate ground truth organelle-like objects with known motility features. All of the simulation tools are made available in the supplemental Nellie repository (https://github.com/aelefebv/nellie-supplemental).

The first two tracking simulations test each algorithm's ability to accurately track multiple objects' motions over a number of frames (Extended Data Fig. 4a-f). To do this, we simulate a dataset consisting of 3 objects with various lengths, widths, and intensities, each moving along an orthogonal axis (i.e. X only, Y only, and Z only) at a consistent rate of 1 pixel per frame in either the direction of its long axis for 26 frames (Extended Data Fig. 4a-c), of the direction of its short axis for 28 frames (Extended Data Fig. 4d-f).

The dataset is then reversed and appended to the end of the original dataset to create a back-and-forth motion of the objects. We set a pixel resolution of 0.2 um/px and a temporal resolution of 0.5 s/frame for the original dataset, and remove temporal frames in increasing powers of 2 to generate additional datasets with temporal resolutions of 1, 2, 4, and 8 s/frame, with objects moving at a consistent speed of 0.4 um/s. We use Nellie's features_branches.csv and im_skel_relabelled.ome.tif outputs to track individual branches over each time course, and monitor which branch we have tracked in each frame to measure how persistently each branch is tracked (Extended Data Fig. 4b,e). For MitoTNT, we attempted to use the node tracking outputs to generate fragment tracking outputs, but this process errored out in >50% of the samples. Instead, we simply use the node tracks' outputs to calculate displacement and monitor which node we have tracked in each frame to measure how persistently each node is tracked. For Mitometer, we use the track number and its centroid at each frame to calculate the velocity, and use the appearance of each track in each frame as the measure of persistence. For both long and short axis travel, Nellie performs essentially perfectly in both linear velocity calculation, and track persistence. Mitometer also performs close to perfectly, with some variability in both velocity and track length. MitoTNT performs decently for linear velocity calculation, but seems unable to consistently keep hold of a track for longer than about 20 frames. We also note that MitoTNT was unable to produce tracks for the 8 s/frame sample in the long-axis case.

The third tracking simulation tests each algorithm's ability to quantify angular velocity of a rotating object, considering organellar movement is not limited to linear velocity alone, but often comes about from angular movement about a pivot, such as an organellar or structural contact site. We generate a single, long object similar to how we generate an object in the first test case, but instead of linearly moving the object, we rotate the object about a single axis at a constant rate of 5 degrees per frame for 33 frames. We again set the pixel resolution to 0.2 um/px and the time interval to 0.5 s/frame, meaning the object is rotating at a constant speed of 10 deg/s about its center of mass (Extended Data Fig. 4g). We again produce additional datasets with frames at increasing powers of 2 removed, meaning the object will have rotated by 5, 10, 20, 40, and 80 degrees between each frame, for each respective dataset. We quantify both linear velocity, which should be near 0, as the object is rotating about its center of mass (though discretized to individual pixels, so it will never be exactly 0), and angular velocity, which should be a constant 10 deg/s. We use the same method as in the first test case to generate the linear velocity calculation for both Nellie and Mitometer. However, for MitoTNT, we instead use their script to convert the node tracking outputs to fragment tracks, which it was only able to do for the 1 and 2 s/frame datasets. For Nellie, we calculate angular velocity using the values in the features_nodes.csv output. Specifically, we get the mean ang_vel_mag_rel_mean, which is the mean angular velocity magnitude, relative to its pivot point, of all voxels that make up a node, and average this value over each frame. We then average each frames' average to get the final angular velocity in rad/s and convert this to deg/s. For MitoTNT, we use each node's track to calculate displacement of each node between frames, interpreted as rad/frame in this sample, and convert this to deg/s. Though this is not a perfectly accurate representation of angular velocity for MitoTNT's outputs, this was the best we could do with their limited outputs. Mitometer has no discernable way to calculate angular velocity from its outputs. Both Nellie and Mitometer perform well at correctly finding low linear velocity values in all datasets, while MitoTNT seems to capture a higher velocity than both other algorithms at the temporal resolutions it was able to convert for fragment tracking (Extended Data Fig. 4h). Nellie performs well at a resolution of 0.5 and 1 s/frame, but performs increasingly poorly, underestimating angular velocity at lower temporal resolutions. MitoTNT performs poorly at high temporal resolution, but seems to perform increasingly better at lower temporal resolutions.

The final two tracking simulations test the algorithms' abilities to maintain object permanence by quantifying its ability to maintain a link over objects after undergoing fragmentation (fission) or merging (fusion) events (Extended Data Fig. 4j-l). We simulate 2 objects in each orthogonal plane of our image, where we extend the object from the center point of the image along either the positive or negative direction of its axis, resulting in a 6 sided star-like object. Every 3 frames, we linearly begin moving an object away from the center, along its long axis, at a rate of 1 voxel/s, again with a pixel resolution of 0.2 um/px and a temporal resolution of 0.5 s/frame. After several frames, all objects have become disconnected from the center, resulting in a total of 5 fission events over the course of 33 frames. We generate another dataset by reversing the time series, where all objects begin disconnected, but end up reconnected by the final frame, resulting in 5 fusion events over the course of 33 frames. As previously, we remove temporal frames in increasing powers of 2 to generate additional datasets with lower temporal resolution. For Nellie, we use the organelle features' reassigned labels, and per-frame labels to track the objects of frame 1 over time. Specifically, for each frame we calculate the difference between the number of parent labels to the number of labels in that frame. If the difference in this difference value is positive between adjacent frames, this indicates a new label has appeared, whereas a negative value indicates the disappearance of a label. For MitoTNT, we run the tracking outputs through their provided code to detect fission and fusion, and sum up all rows with a fission type to count fission events, and all rows with a fusion type to count fusion events. For Mitometer, we sum up all raw detected fission events to count fission events, and all raw detected fusion events to count fusion events. Nellie perfectly detects all 5 fission and all 5 fusion events for all temporal resolutions (Extended Data Fig. 4k). MitoTNT detects at most 2 fission events and 1 fusion events, with no events detected at the lowest temporal resolutions. Mitometer picks up all 5 fission events for a few temporal resolutions, and detects no fusion events for a few temporal resolutions, detecting no events at all for the lowest temporal resolution.

**Supplementary Note 10: Creation of feature-based cost matrix for inter-frame mocap marker linkage**

To create the feature-based cost matrix used to link corresponding mocap markers between adjacent frames, a speed matrix is first created by calculating the displacement between all marker indices in adjacent frames (t and t+1) and dividing this by the corresponding time between frames. Let $P_t$ and $P_{t+1}$ be the marker positions at time t and t+1, respectively.

$$S = \frac{\|P_{t+1} - P_t\|}{t_{t+1} - t_t}$$

where $\|\cdot\|$ denotes the Euclidean norm.

The speeds are normalized to the maximum permissible speed, defaulting to 1 µm/sec. Let $S_{max}$ be the maximum permissible speed:

$$S_{norm} = \frac{S}{S_{max}}$$

Any linkages exceeding the maximum distance threshold are prohibited. The speed matrix is then z-score normalized for standardization. For a matrix M:

$$Z(M) = \frac{M - \overline{M}}{std(M)}$$

Absolute value differences of the stats vectors for markers between the two frames are computed and standardized. Let $V_t$ and $V_{t+1}$ be the stats vectors at time t and t+1, respectively.

$$\Delta V = |V_{t+1} - V_t|$$

The resulting matrix is divided by the number of feature columns in the stats vector to provide balanced weighting in the final cost matrix. Let $n_s$ be the number of features in the stats vector.

$$F_{norm} = \frac{Z(\Delta V)}{n_s}$$

Absolute value differences of the Hu vectors for markers between the two frames are computed and standardized. Let $H_t$ and $H_{t+1}$ be the Hu vectors at time t and t+1, respectively.

$$\Delta H = |H_{t+1} - H_t|$$

The Hu matrix, after normalization, is adjusted by dividing the matrix by the number of feature columns in the Hu vector, which varies depending on whether the dataset is 2D or 3D. Let $n_H$ be the number of features in the Hu vector.

$$H_{norm} = \frac{Z(\Delta H)}{n_H}$$

The standardized distance, stats, and Hu matrices are summed to form the final cost matrix, which will be used to temporally link mocap markers:

$$C = Z(S_{norm}) + F_{norm} + H_{norm}$$

**Supplementary Note 11: Temporal continuity in organelle tracking via forward and backward interpolation of semantic segmentations across frames**

Nellie includes an optional, but highly requested component for tracking instance segmentations of individual organelle objects between adjacent frames. This feature allows for the extraction of motility features and changes in morphology of a specific label across multiple temporal frames. Specifically, we maintain continuity in the identification of labeled organelle voxels across frames by sequentially relabelling adjacent frames via both a forward and backward interpolation scheme from a temporal frame T of interest (Extended Data Fig. 5).

For forward relabelling from frame t to t+1, forward interpolation of motion vectors for all semantic segmentation mask voxels in frame t is performed, yielding interpolated coordinates at frame t+1. Let $V_t$ be the set of voxel coordinates in frame t, and $F_t$ be the flow vectors at frame t.

$$C_{t+1} = V_t + F_t(V_t)$$

where $C_{t+1}$ are the interpolated coordinates at frame t+1.

Similarly, backward interpolation of motion vectors for all semantic segmentation mask voxels in frame t+1 is carried out, providing interpolated coordinates at frame t. Let $V_{t+1}$ be the set of voxel coordinates in frame t+1, and $B_{t+1}$ be the backward flow vectors at frame t+1.

$$C_t = V_{t+1} + B_{t+1}(V_{t+1})$$

where $C_t$ are the interpolated coordinates at frame t.

Next, frame t voxels are matched to the closest frame t interpolated coordinates derived from frame t+1 voxels, and vice versa. Matches are only considered if they are within the maximum travel distance threshold (1 um/s). For each $v$ in $V_t$ and $v'$ in $V_{t+1}$:

$$\text{Match}(v, v') \text{ if}$$

$$\|v - C_t(v')\| < d_{max}$$

$$\text{and}$$

$$\|'v - C_{t+1}(v)\| < d_{max}$$

where $d_{max}$ is the maximum travel distance threshold.

Unique matches are then assigned using a heap in-place priority queue based on the distance between the voxel and interpolated coordinate match. This prioritization ensures that the most accurate matches are assigned first, enhancing both computational efficiency and matching accuracy.

Let M be the set of all matches $(v, v', d)$ where $d$ is the distance between $v$ and $C_{t+1}(v)$:

$$\text{UniqueMatches} = \text{AssignUniqueMatches}(\text{SortByDistance}(M))$$

where AssignUniqueMatches assigns matches using a priority queue based on distance.

Any unlabelled voxel in frame t+1 is assigned the label of its nearest relabelled voxel, provided the distance is within the maximum travel distance threshold. For each unassigned voxel $u$ in $V_{t+1}$:

$$\text{Label}(u) = \text{Label}(argmin_v(\|u - v\|)$$

$$\text{if } min_v(\|u - v\|) < d_{max}$$

where $v$ are the assigned voxels in $V_{t+1}$.

This relabelling process is repeated until all voxels are labeled or until the number of unlabelled voxels stabilizes between iterations:

$$|Unassigned_i| - |Unassigned_{i-1}| = 0$$

or

$$|Unassigned_i| = 0$$

where $|Unassigned_i|$ is the number of unassigned voxels in iteration i.

The backward relabelling process from frame t to t-1 mirrors the forward relabelling approach. First, forward and backward interpolations are conducted for frame t-1 and frame t voxels, respectively. Voxels from frame t are matched to the closest interpolated coordinates derived from frame t-1, and vice versa, within the maximum travel distance threshold. The assignment of unique matches and relabelling of unassigned voxels follows the same procedure as in forward relabelling, ensuring consistency and accuracy across temporal frames.

**Supplementary Note 12: Spatial and temporal feature extraction of hierarchical organelle levels, extended**

Organelle features are extracted hierarchically, with voxels representing the smallest unit of measured space, followed by nodes, which are skeleton voxels plus their surrounding radius-dependent voxels, followed by branch components, which are semantic segmentations of skeleton branches and their surrounding voxels, followed by organelles, which are semantic segmentations of connected components, followed finally by the organellar landscape, representing the entire image. Each larger component in the hierarchy also gets aggregated statistics of features from the lower levels. We have described all of the extractable features and their respective variable names in Supplementary Table 1.

At the voxel level, local flow, structure, and fluorescence intensity information is captured. At frame T, a voxel's flow is interpolated from T-1 to T, and from T to T+1 to generate its independent linear and angular velocity and acceleration vectors, of which we can break down to its magnitude and normalized orientation vectors. We also choose, for each branch, a local pivot point, defined as the voxel within a branch that has the smallest velocity vector magnitude between the two frames of interest, as well as the entire organellar landscape's center of mass, representing two different points of reference for two additional sets of linear and angular motility features. We can then extract a directionality component by calculating the voxel's motion with respect to these points of reference.

a. Linear Velocity:

$$v_{lin}(t) = \frac{r(t+1) - r(t)}{\Delta t}$$

b. Angular Velocity (3D):

$$v_{ang}(t) = \frac{r(t) \times r(t+1)}{\Delta t ||r(t) \cdot r(t+1)||}$$

c. Linear Acceleration:

$$a_{lin}(t) = \frac{v_{lin}(t+1) - v_{lin}(t)}{\Delta t}$$

d. Angular Acceleration:

$$a_{ang}(t) = \frac{v_{ang}(t+1) - v_{ang}(t)}{\Delta t}$$

e. Directionality:

$$D = \frac{|r_{ref}(t+1)| - |r_{ref}(t)|}{|r_{ref}(t+1)| + |r_{ref}(t)|}$$

where $r_{ref}$ is the position relative to a reference point (local pivot or center of mass)

At the node level, we calculate neighborhood flow patterns, as well as the thickness of an organelle branch at that skeleton voxel. To extract flow patterns, we first retrieve all the flow vectors associated with the voxels of that node. To calculate the convergence of flow vectors to that node's center, we calculate the mean direction of magnitude of all the node's voxels with respect to the node center from T-1 to T. A positive number indicates voxels are flowing towards that node from T-1 to T. In contrast, divergence is calculated in the same way as convergence, but from T to T+1. A positive number indicates voxels are flowing away from that node from T to T+1. Vergere is simply the sum of divergence and convergence, where a positive number would indicate a local bottleneck of flow, and a negative number would indicate some flow repulsion from that node. The magnitude variability of flow vectors at that node is simply the standard deviation of the node's voxels' flow vector magnitudes. The direction uniformity is the mean value of the dot product similarity matrix of the node's voxels' orientation vectors.

  a. Convergence:

$$C = -\overline{\Sigma(-v_{in})d_{in}}$$

where $v_{in}$ is the inward velocity and $d_{in}$ is the direction to node center

  b. Divergence:

$$C = -\overline{\Sigma(-v_{out})d_{out}}$$

where $v_{out}$ is the outward velocity and $d_{out}$ is the direction from node center

  c. Vergere:

$$V = C + D$$

  d. Magnitude Variability:

$$MV = std(||v||)$$

where v are the velocities of node voxels

  e. Direction Uniformity:

$$DU = \overline{v_i v_j}$$

where $v_i$ and $v_j$ are normalized velocity vectors of node voxels

At the branch level, we calculate skeleton metrics such as length, thickness, aspect ratio, and tortuosity, as well as standard region properties such as area, axes lengths, and solidity and extent metrics. To calculate skeleton properties, we treat each network as a graph, composed of lone tip (no adjacent), tip (one adjacent), edge (two adjacent), and junction (>2 adjacent) voxels. We modify the skeleton graph by removing all junction voxels and replacing their neighboring edge voxels with tip voxels. Each tip-to-tip segment is considered one branch instance. To efficiently traverse this graph, we construct a distance matrix between all non-zero voxels in our skeleton and link voxels that are less than 2 voxels from one another, using only the lower diagonal to avoid duplicate matches. From these linkages, we can calculate the length of a branch as the sum distance of all linkages of that branches' voxels plus the voxel-to-border distance of the tip nodes of that branch. If the branch consists of only a single lone node, we instead calculate the length of that branch to be 2 times the voxel-to-border distance of the node. From this length measurement we calculate the tortuosity of a branch, defined as its length divided by the tip-to-tip spatial distance. We calculate the thickness of a branch as its mean voxel-to-border distance times two, and its aspect ratio as its length divided by its thickness. For calculating the region's properties, we use Scikit-Image's regionprops function.

a. Branch Length:

$$L = \Sigma(d_{ij}) + d_{tip1} + d_{tip2}$$

where $d_{ij}$ are distances between adjacent voxels and $d_{tip}$ are tip-to-border distances

b. Tortuosity:

$$T = \frac{L}{\|r_{tip1} - r_{tip2}\|}$$

c. Thickness:

$$Th = 2\overline{d_b}$$

where $d_b$ are voxel-to-border distances

    d. Aspect Ratio:

$$AR = \frac{L}{Th}$$

At the organelle level, we simply calculate the connected component's region properties, again via Scikit-Image's regionprops function.

Finally, at the organellar landscape (image) level, we simply allow all aggregate statistics of all other hierarchies to be stored for export.

**Supplementary Table 1: Table of Nellie's output features**

For each level of the organellar hierarchy, we extract a multitude of features, as described in Supplementary Note 12.

Voxels:

| Feature description (units) | Output name |
| --- | --- |
| The physical z location of that voxel (um) | z_raw |
| The physical y location of that voxel (um) | y_raw |
| The physical x location of that voxel (um) | x_raw |
| The fluorescence intensity image's value at that voxel (A.U.) | intensity_raw |
| The preprocessed image's intensity value at that voxel (A.U.) | structure_raw |
| Movement directionality wrt local pivot point (N/A) | directionality_rel_raw |
| Angular acceleration magnitude wrt local pivot point (rad/s$^2$) | ang_acc_mag_rel_raw |
| Angular acceleration magnitude, no reference point (rad/s$^2$) | ang_acc_mag_raw |
| Linear acceleration magnitude wrt local pivot point (um/s$^2$) | lin_acc_mag_rel_raw |
| Linear acceleration magnitude, no reference point (um/s$^2$) | lin_acc_mag_raw |
| Angular velocity magnitude, wrt local pivot point (rad/s) | ang_vel_mag_rel_raw |
| Angular velocity magnitude, no reference point (rad/s) | ang_vel_mag_raw |
| Linear velocity magnitude wrt local pivot point (um/s) | lin_vel_mag_rel_raw |
| Linear velocity magnitude, no reference point (um/s) | lin_vel_mag_raw |

Nodes:

| Feature description (units) | Output name |
| --- | --- |
| The physical z location of that node's centroid (um) | z_raw |
| The physical y location of that node's centroid (um) | y_raw |
| The physical x location of that node's centroid (um) | x_raw |
| Distance from the node center to the edge x2 (um) | node_thickness_raw |
| The sum of the convergence and divergence of vectors to the node center | vergere_raw |

| Feature description (units) | Output name |
|---|---|
| (um/s) | |
| The convergence of the flow vectors of the voxels in the node to the node center (um/s) | convergence_raw |
| The divergence of the flow vectors of the voxels in the node from the node center (um/s) | divergence_raw |
| Summary statistics of the node's voxels' metrics (see above) | voxel aggregate metrics |

Branches:

| Feature description (units) | Output name |
|---|---|
| The physical z location of that branch's centroid (um) | z_raw |
| The physical y location of that branch's centroid (um) | y_raw |
| The physical x location of that branch's centroid (um) | x_raw |
| Label corresponding to its original temporal frame 0 label, assigned during voxel reassignment, if selected (N/A) | reassigned_label_raw |
| The number of voxels in that branch divided by the number of voxels in that branch's convex hull (N/A) | branch_solidity_raw |
| The number of voxels in that branch divided by the number of voxels in that branch's bounding box (N/A) | branch_extent_raw |
| Minor axis length of an ellipse with the same normalized second central moments as the branch (um) | branch_axis_length_min_raw |
| Major axis length of an ellipse with the same normalized second central moments as the branch (um) | branch_axis_length_maj_raw |
| The number of voxels in the branch, scaled by the image resolution (2D: $um^2$  3D: $um^3$). In 3D this is the volume. | branch_area_raw |
| The branch's tips' euclidean distance divided by its length (N/A) | branch_tortuosity_raw |
| The length of the branch divided by its median thickness (N/A) | branch_aspect_ratio_raw |
| The mean thickness of the branch (um) | branch_thickness_raw |
| The length of the branch (um) | branch_length_raw |
| Summary statistics of the branch's voxels' metrics (see above) | voxel aggregate metrics |
| Summary statistics of the branch's nodes' metrics (see above) | node aggregate metrics |

Organelles:

| Feature description (units) | Output name |
|---|---|
| The physical z location of that organelle's centroid (um) | z_raw |
| The physical y location of that organelle's centroid (um) | y_raw |
| The physical x location of that organelle's centroid (um) | x_raw |
| Label corresponding to its original temporal frame 0 label, assigned during voxel reassignment, if selected (N/A) | reassigned_label_raw |
| The number of voxels in that organelle divided by the number of voxels in that organelle's convex hull (N/A) | organelle_solidity_raw |
| The number of voxels in that organelle divided by the number of voxels in that organelle's bounding box (N/A) | organelle_extent_raw |
| Minor axis length of an ellipse with the same normalized second central moments as the organelle, scaled by the image resolution (um) | organelle_axis_length_min_raw |
| Major axis length of an ellipse with the same normalized second central moments as the organelle, scaled by the image resolution (um) | organelle_axis_length_maj_raw |
| The number of voxels in the organelle, scaled by the image resolution (2D: $um^2$ 3D: $um^3$). In 3D this is the volume. | organelle_area_raw |
| Summary statistics of the organelle's voxels' metrics (see above) | voxel aggregate metrics |
| Summary statistics of the organelle's nodes' metrics (see above) | node aggregate metrics |
| Summary statistics of the organelle's branches' metrics (see above) | branch aggregate metrics |

Image

| Feature description (units) | Output name |
|---|---|
| Summary statistics of the image's voxels' metrics (see above) | voxel aggregate metrics |
| Summary statistics of the image's nodes' metrics (see above) | node aggregate metrics |
| Summary statistics of the image's branches' metrics (see above) | branch aggregate metrics |
| Summary statistics of the image's organelles' metrics (see above) | organelle aggregate metrics |

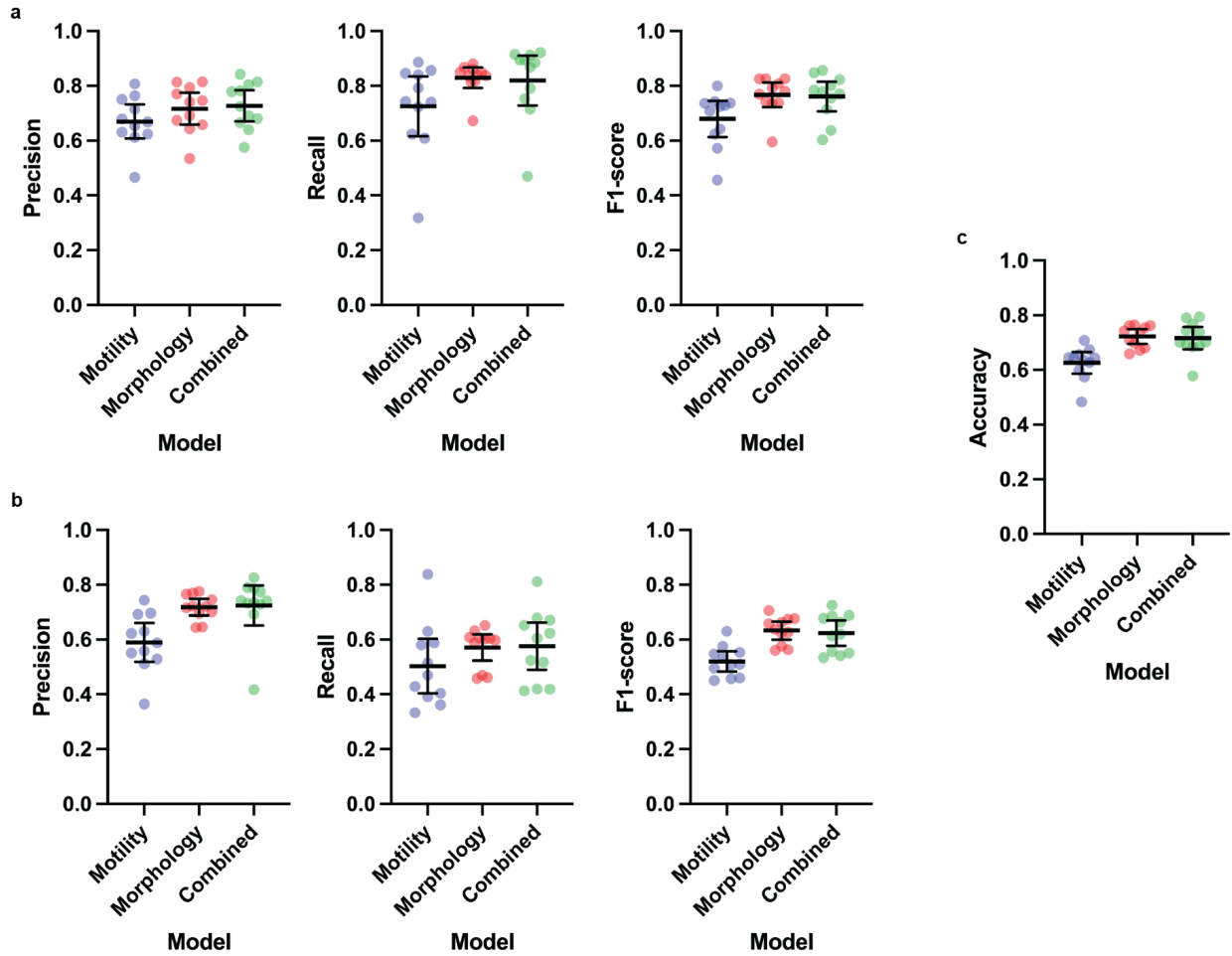

**Supplementary Data Fig. 7: Additional performance metrics from organelle unmixing random forest classifiers**
The precision, recall, and F1-score for Golgi (**a**) and mitochondria (**b**) classification of the motility-only, morphology-only and motility and morphology combined random forest models. **c**, The combined accuracy metric of the same three models for overall classification.

**Supplementary Note 13: Multi-mesh graph-based representation of cellular organelle networks**

In the construction of the multi-mesh network from organelle segmentation masks, the methodology employs a systematic approach to transform the intricate spatial relationships of organelles into a structured graph representation. This transformation is achieved through a series of computational steps designed to map the skeletonized representations of organelles onto a graph where nodes represent skeleton voxels that carry features of their surrounding voxels, and edges that delineate the spatial connections between these nodes, capturing the underlying organizational topology of the organelle network.

The initial step involves the utilization of a k-d tree for the efficient identification of immediately neighboring nodes. This radius is chosen to ensure that all immediate neighbors in a cubic voxel grid are considered, facilitating the accurate representation of spatial adjacencies within the graph. Each node, corresponding to a skeleton voxel, is then associated with a list of its neighbors, excluding self-references, to establish the basic graph structure. For each skeleton voxel v:

$$N(v) = \frac{\{u \epsilon V : ||v-u|| \leq 1.74\}}{v}$$

where $V$ is the set of all skeleton voxels, and $||\cdot||$ denotes Euclidean distance.

Subsequently, the algorithm identifies a starting node for the graph, preferentially selecting a node with a singular neighbor (a tip node) to anchor the construction process. In the absence of such nodes, indicating a closed network, the first node in the branch's list is selected by default.

start_node = first(n for n in nodes if len(neighbors(n)) == 1)

if not start_node:

start_node = nodes[0]

The construction of the multi-level mesh within the graph is then initiated by calculating the jump distances from the start node to all other nodes within the network:

$$J(v) = \min\{d : \exists \text{ path } p = (v_0, v_1, ... v_d) \text{ where } v_0 = s, v_d = v, \text{ and } (v_i, v_{i+1}) \in E$$

$$\text{for } 0 \leq i \leq d\}$$

where $s$ is the start node and $E$ is the set of edges between neighboring voxels $v$. This calculation employs a depth-first search algorithm, incrementally increasing the jump distance as the search progresses through the network. The jump distances serve as a basis for establishing a hierarchical structure within the graph, where connections between nodes are formed based on increasing powers of two. This hierarchical connectivity facilitates efficient message passing at multiple distances, enabling the graph to capture a wide range of spatial relationships within the organelle network.

For each scale within the multi-level mesh, the algorithm generates a list of nodes that are valid at that scale, determined by their jump distances:

$$V_k = \{v \in V : J(v) \% 2^{k-1} = 0\}$$

where k is the scale level ($k \geq 1$).

This selective inclusion of nodes at each scale allows for the dynamic adjustment of the graph's hierarchy, ensuring that both local details and global organizational patterns are adequately represented. The edges within the multi-scale edge list are then generated by direct accessibility queries within the graph, establishing connections between nodes based on their proximity and mutual accessibility across the different scales of the mesh:

$$E_k = \{(u, v) : u, v \in V_k \text{ and } d(u, v) \leq k^2 + 1\}$$

where $d(u, v)$ is the shortest path distance between u and v in the original graph.

This detailed and systematic approach to constructing the multi-mesh network from organelle segmentation masks and their respective skeletons encapsulates the complex spatial organization of organelles within a scalable and interpretable graph framework. By leveraging the principles of graph theory and computational geometry, the methodology provides a robust foundation for analyzing the dynamic alterations in organelle organization, offering insights into the functional implications of spatial arrangements within cellular environments.

**Supplementary Note 14: Graph autoencoder model construction for organelle multi-mesh network analysis**

The graph autoencoder model is structured to transform multi-dimensional data into a comparable latent space and subsequently reconstruct the original data, aiming to retain significant biological features and spatial relationships. This allows for latent space analysis of encoded features and complex interactions within organellar networks. The graph autoencoder comprises an encoder for data compression and a decoder for data reconstruction, each built with layers specifically chosen for their effectiveness in graph-structured data processing.

The encoder initiates with a Multilayer Perceptron (MLP) to project the initial feature set into a 512-dimensional embedding. Following the MLP, a Swish (SiLU) activation function introduces non-linearity, enhancing the model's ability to capture complex patterns. Layer normalization is applied to ensure consistent training dynamics across layers by normalizing the layer outputs:

$$E(x) = \mathrm{LN}(\mathrm{SiLU}(W_e x + b_e))$$

where LN is Layer Normalization, $W_e$ and $b_e$ are learnable parameters.

The core of the encoder consists of 16 graph neural network (GNN) layers. These layers are designed for effective feature transformation and aggregation within the graph, utilizing mean aggregation for message passing. The independent weighting of each GNN layer allows for flexibility in learning distinct aspects of the data. Residual connections are incorporated to facilitate deeper learning without the vanishing gradient problem:

$$h_i^{(I+1)} = \text{LN}\left(\text{SiLU}\left(W_I \cdot \overline{\{h_j^I : j \in N(i)\}} \, b_i\right) + h_i^I\right)$$

where $N(i)$ is the set of neighbors of node $i$, and $W_I$ and $b_i$ are layer-specific parameters. For the encoder output:

$$z = h^L$$

where L is the number of GNN layers (16 in this case).

The decoder mirrors the encoder's depth but switches to graph attention network (GAT) layers:

$$a_{ij} = \text{softmax}_j\left(a^T[W_d h_i \| W_d h_j]\right)$$

$$h_i^{I+1} = \text{LN}\left(\text{SiLU}\left(\Sigma_j \alpha_{ij} W_d h_j^I\right) + h_i^I\right)$$

where $a$ and $W_d$ are learnable parameters, and $\|$ denotes concatenation. For the decoder output:

$$x' = h^{L'}$$

where $L'$ is the number of GAT layers in the decoder.

These layers leverage attention mechanisms to prioritize the most relevant features from neighboring nodes, crucial for accurately reconstructing the original feature space from the compressed latent representation. Dropout (DO) is incorporated at a rate of 20% within the GAT layers to prevent overfitting by randomly omitting subsets of features during training:

$$h_{DO} = h \cdot mask, \; mask \sim Bernoulli(p = 0.8)$$

Applied in GAT layers with 20% dropout rate.

Similar to the encoder, the decoder employs Swish activation and layer normalization post each GAT layer, concluding with a transformation back to the original feature size. This design ensures that the decoder can effectively reconstruct the original data, highlighting changes in organelle configurations.

The initial embedding is essential for transitioning from the high-dimensional raw feature space to a more comparable representation without significant information loss. The MLP efficiently achieves this transformation, while the SiLU activation function is chosen for its property of smoothing nonlinearities, improving model performance over traditional ReLU functions. The choice of GNN layers for the encoder stems from their ability to capture the dependency and interaction between intra-organellar nodes in a graph. Mean aggregation is used for its simplicity and effectiveness in summarizing variable-range neighborhood information, essential for understanding the context of each organelle node within the network. The adoption of GAT layers for the decoder is motivated by the need for a nuanced reconstruction of the original data. The attention mechanism within GAT layers allows the model to focus on the most informative parts of the input graph, crucial for detailed and accurate data reconstruction. Layer normalization and residual connection components are integral to the model for maintaining stable learning rates across training and enabling the effective training of deep networks by allowing information and gradients to flow through the network without diminishment.

The model implementation leverages PyTorch and PyTorch Geometric, facilitating efficient graph data processing and neural network operations[71,72]. Training involves minimizing the Mean Squared Error (MSE) between the original and reconstructed datasets, with the Adam optimizer chosen for its adaptive learning rate capabilities, especially suited for sparse data commonly found in graphs:

$$\text{Loss} = \text{MSE}(x, x') = \frac{1}{n} \Sigma_i (x_i - x'_i)^2$$

$$\theta_{t+1} = \frac{\theta_t - \eta \cdot m_t}{\sqrt{v_t} + \varepsilon}$$

where $m_t$ and $v_t$ are the first and second moments of the gradients, computed by the Adam optimizer.

For training and validation of our model, we randomly assign 70% of the graphs in each dataset for training and 30% for validation, evaluating the model on unseen data. Training uses an Adam optimizer with a 0.01 learning rate and implements early stopping to prevent overfitting. We monitor validation loss and halt training if no improvement occurs for 10 consecutive epochs, with a 0.001 minimum improvement threshold. We plot and compare training and validation loss curves to visually identify divergence indicating overfitting. We prevent data leakage by strictly separating training and validation datasets, performing normalization and feature scaling independently on each set. We use a batch size of 1 to process each graph individually, avoiding information sharing between samples. The model's performance is continuously evaluated on the validation set, with the best-performing model saved based on lowest validation loss (Supplementary Data Fig. 8). This approach to training and validation, coupled with checks for overfitting and data leakage, ensures the integrity and reliability of our GNN Autoencoder model.

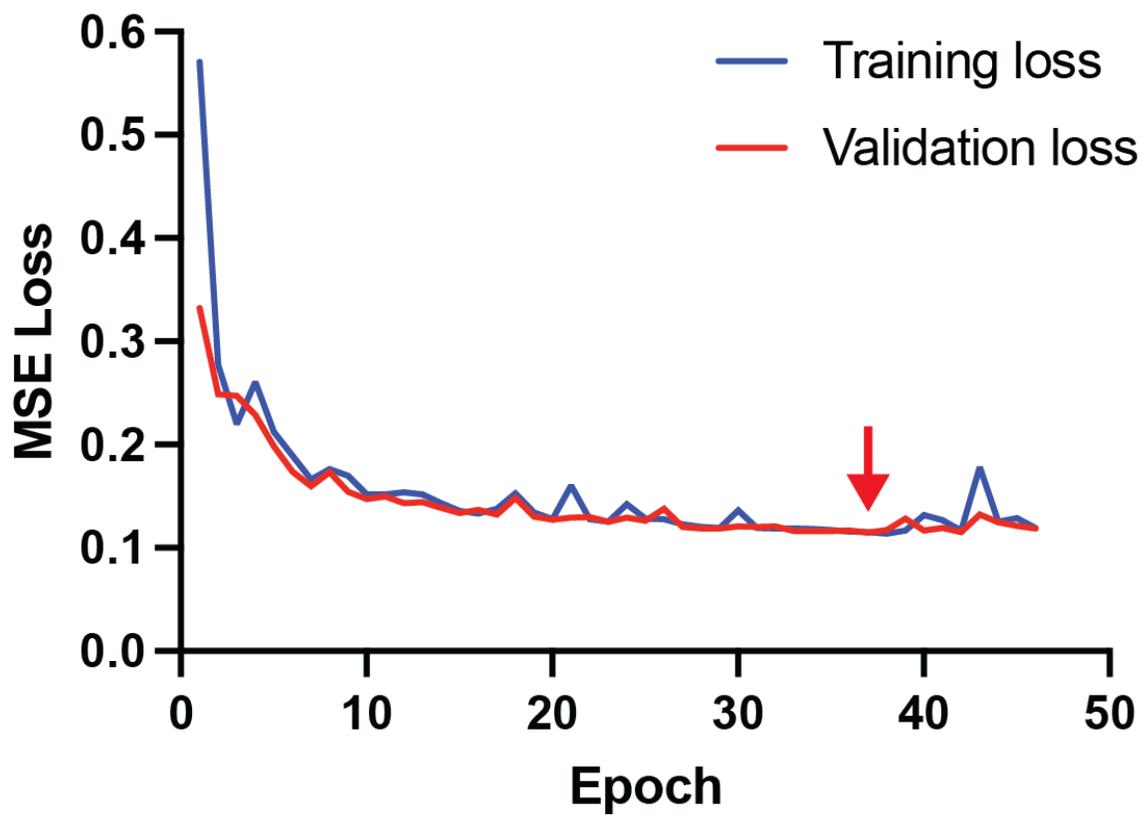

**Supplementary Data Fig. 8: Loss curves for the Multi-mesh GNN**
Loss curves for the multimesh GNN are shown for training (blue) and validation (red) datasets. The model was trained for a total of 47 epochs. The model was saved for the lowest validation loss epoch (37, red arrow).

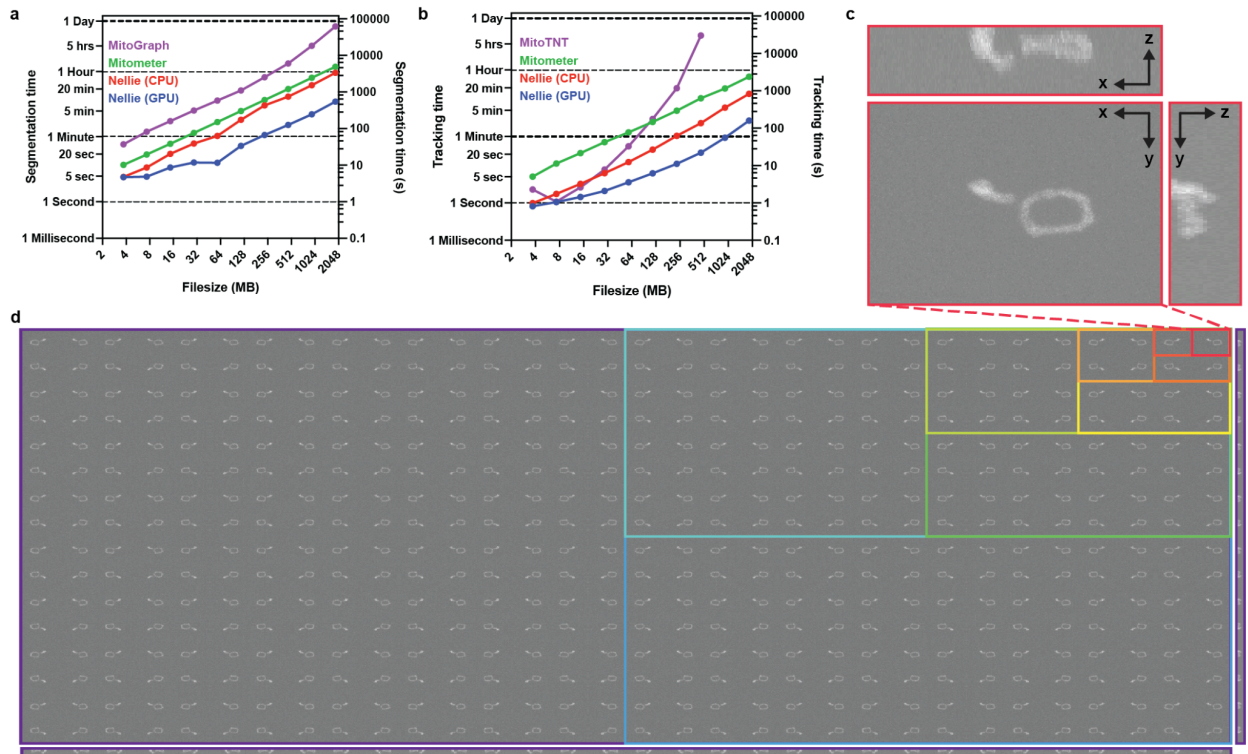

**Extended Data Fig. 1: Runtime comparisons**
**a**, Segmentation time comparisons for various tools with datasets of increasing size: MitoGraph (purple), Mitometer (green), Nellie running on CPU (red), and Nellie running on GPU (blue). **b**, Tracking time comparisons for different tools: MitoTNT (purple), Mitometer (green), Nellie on CPU (red), and Nellie on GPU (blue), across datasets of increasing size. **c**, Visualization of the first of two timepoints from the smallest dataset (3.7 MB). **d**, Visualization of the first of two timepoints from the largest dataset (1.87 GB), with intermediate dataset sizes marked by different colored borders.

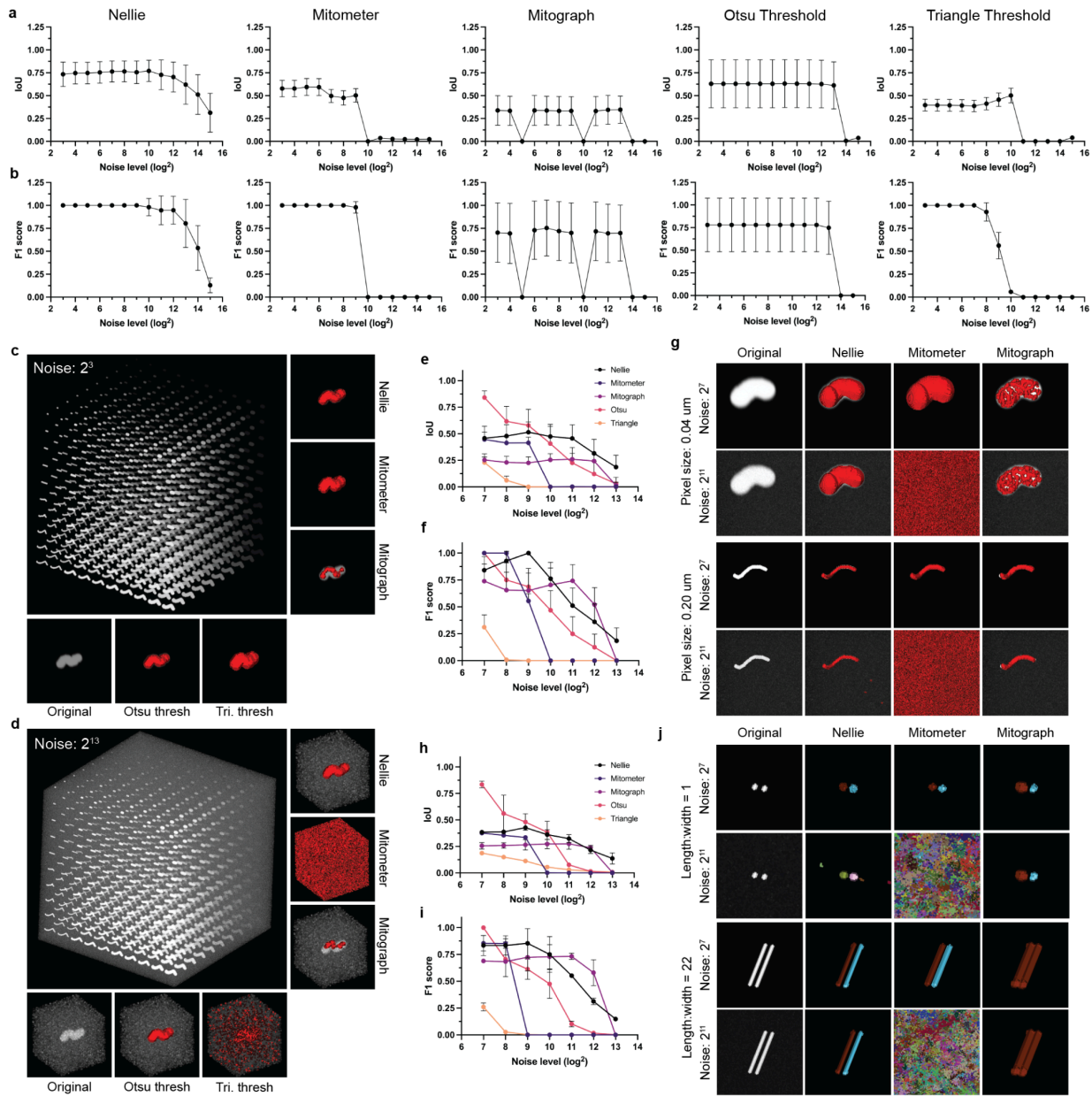

**Extended Data Fig. 2: Segmentation comparison and benchmarking against state-of-the-art organelle segmentation algorithms**
Quantification of segmentation quality of Nellie, Mitometer, Mitograph, a simple Otsu threshold, and a simple triangle threshold from intersection over the union (IoU) (**a**), and F1 scores (**b**) via comparisons of the algorithms' outputs against generated ground truth data. N=10, where each point is the mean score at a different intensity value. Generated ground truth consists of 1000 simulated organelle objects at low (**c**) to high (**d**) noise levels. Images of a single object's segmentation are shown for each method. **e, f**, IoU and F1 scores at increasing noise levels for objects of various lengths and thicknesses generated at varying pixel resolutions. N=32, where each point is the mean score at a different pixel resolution. **g**, Representative images of generated objects at low and high noise levels at high and low pixel resolutions. **h, i,** IoU and F1 scores at increasing noise levels for objects of various lengths and thicknesses generated at varying pixel resolutions at a separation distance of at least 2 times the objects' thicknesses. N=3, where each point is the mean score at a different distance. **j**, Representative images of generated objects at low and high noise levels at low and high length:width ratios. All data points are mean +/- 95% confidence intervals.

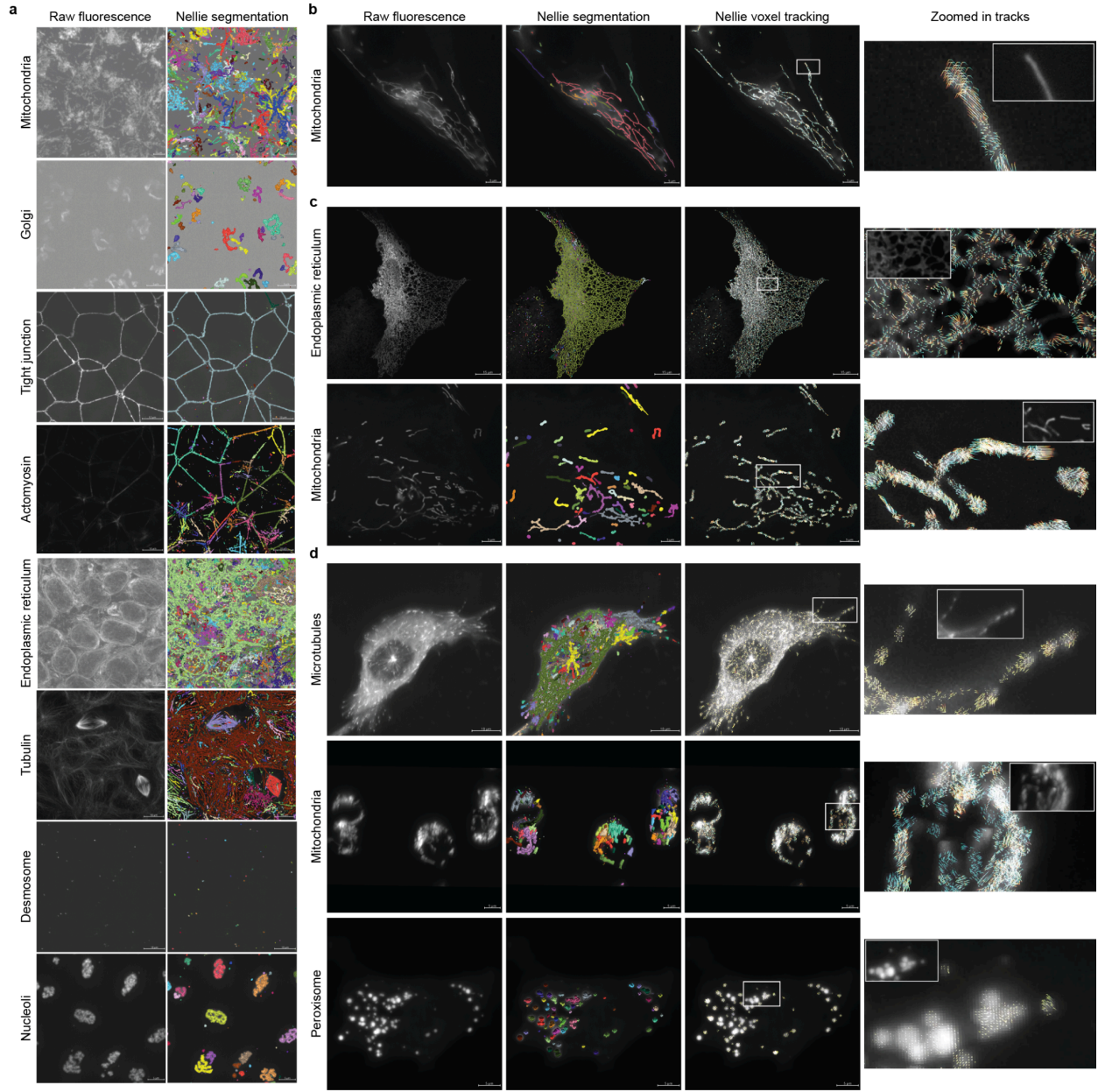

**Extended Data Fig. 3: Segmentation and tracking showcase across organelles and imaging modalities**

**a**, Examples of Nellie's segmentation algorithm outputs on the Allen Institute of Cell Science's 3D single-frame confocal microscopy datasets of various fluorescently labeled structures including their raw fluorescence intensity images on the left. **b**, Example of Nellie's segmentation and tracking algorithm outputs on a 2D timelapse widefield dataset of fluorescently labeled mitochondria, including its raw fluorescence intensity image on the left. **c**, Examples of Nellie's segmentation and tracking algorithm outputs on 2D timelapse confocal spinning disk microscopy datasets of fluorescently labeled endoplasmic reticulum and mitochondria, including their raw fluorescence intensity images on the left. **D**, Examples of Nellie's segmentation and tracking algorithm outputs on 3D timelapse lightsheet microscopy datasets of fluorescently labeled microtubules (EB2), mitochondria, and peroxisomes, including their raw fluorescence intensity images on the left. Zoomed in tracks correspond to supplementary videos.

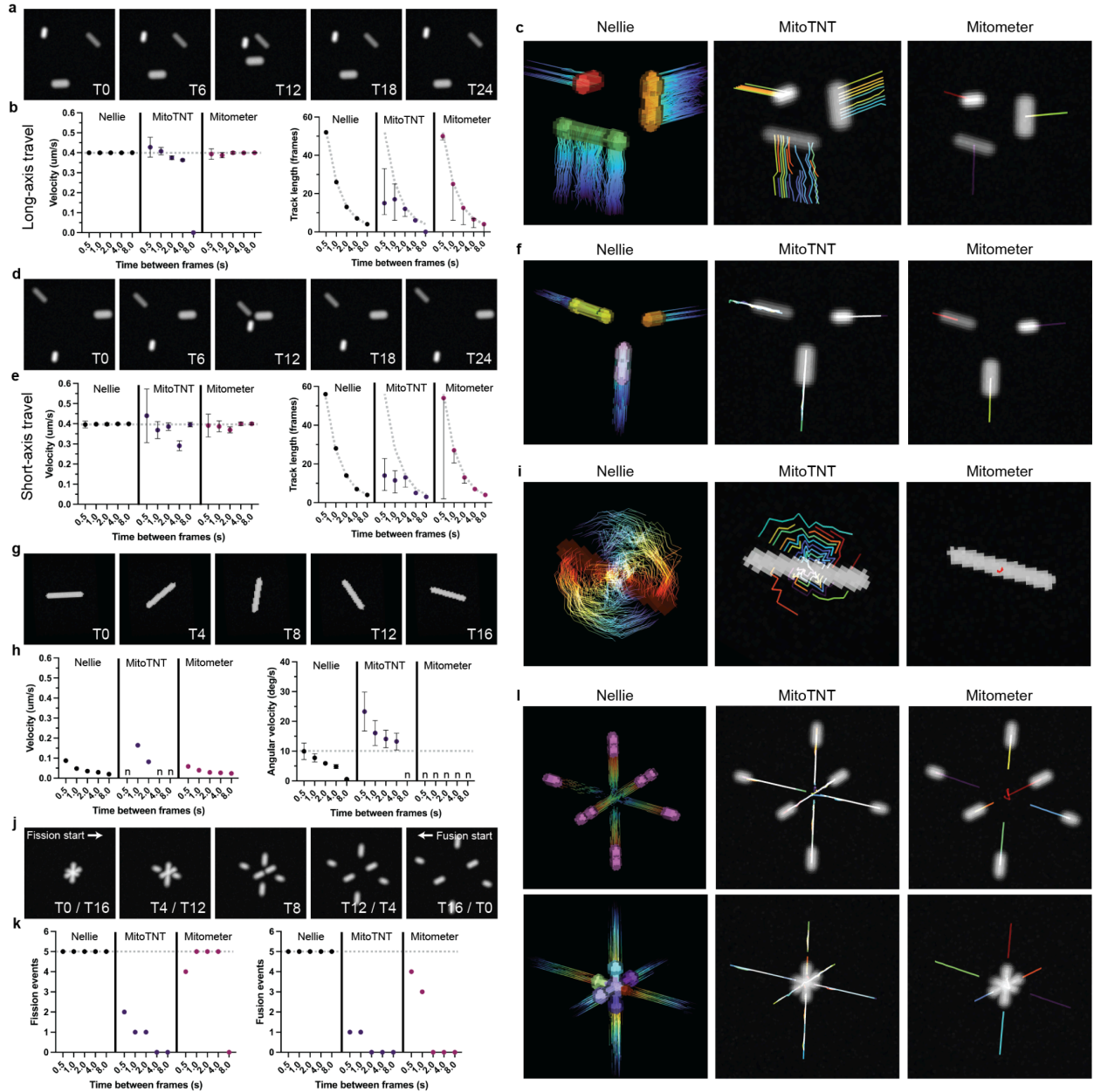

**Extended Data Fig. 4: Tracking comparison and benchmarking against state-of-the-art organelle tracking algorithms**

Simulated 3D organelle objects of varying characteristics traveling along their long (**a**) or short (**d**) axis, in three orthogonal directions. Linear velocity (left) and track length (right) quantification outputs (**b,e**), and representative branch segmentation (Nellie) and track visualizations for the 1 s/frame temporal resolution (**c, f**) for Nellie, MitoTNT, and Mitometer. Points are mean +/- 95% CI (left) and median +/- upper/lower quartiles (right). Dotted line is ground truth. **g**, Simulated 3D organelle object rotating along its non-short axis. Linear velocity (left) and angular velocity (right) quantification outputs (**h**), and branch segmentation (Nellie) and track visualizations for the 1 s/frame temporal resolution (**i**) for Nellie, MitoTNT, and Mitometer. Points are single values (left) and mean +/- 95% CI (right). Dotted line is ground truth. **j**, Simulated 3D organelle objects of varying lengths undergoing fission events (left to right) or fusion events (right to left). Number of quantified fission (left) and fusion (right) events (**k**) and branch segmentation (Nellie) and track visualizations for the 1 s/frame temporal resolution (**l**) for Nellie, MitoTNT, and Mitometer. Points are single values. Dotted line is ground truth.

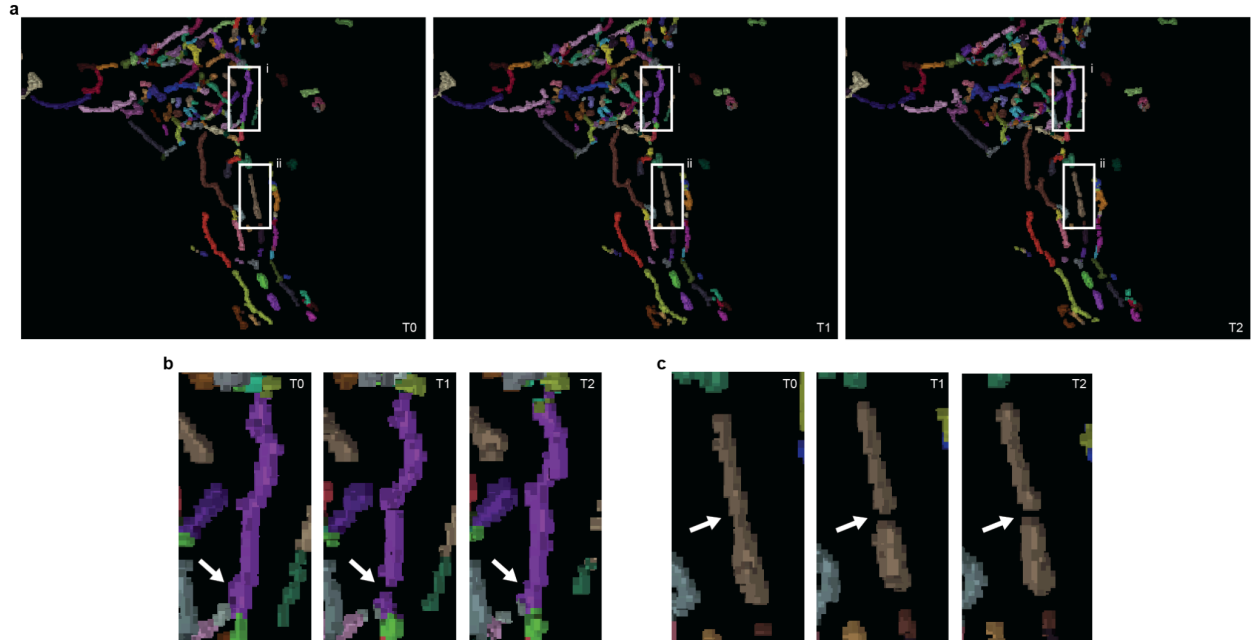

**Extended Data Fig. 5: Temporal continuity in organelle tracking via forward and backward interpolation of semantic segmentations across frames**

**a**, Three consecutive frames of labeled semantic segmentations of mitochondrial branch objects after undergoing forward and backward flow interpolation for voxel label reassignment based on T0's object labels. **b**, An example of an object (purple, **a** i.) undergoing a fission event in T1, and fusion event in T2, while keeping track of original T0 labeled voxels, despite the object's connectivity discontinuity between frames. **c**, Another example of an object (brown, **a** ii.) undergoing a fission event in T1, while keeping track of original T0 labeled voxels, despite the object's connectivity discontinuity in T1 and T2.

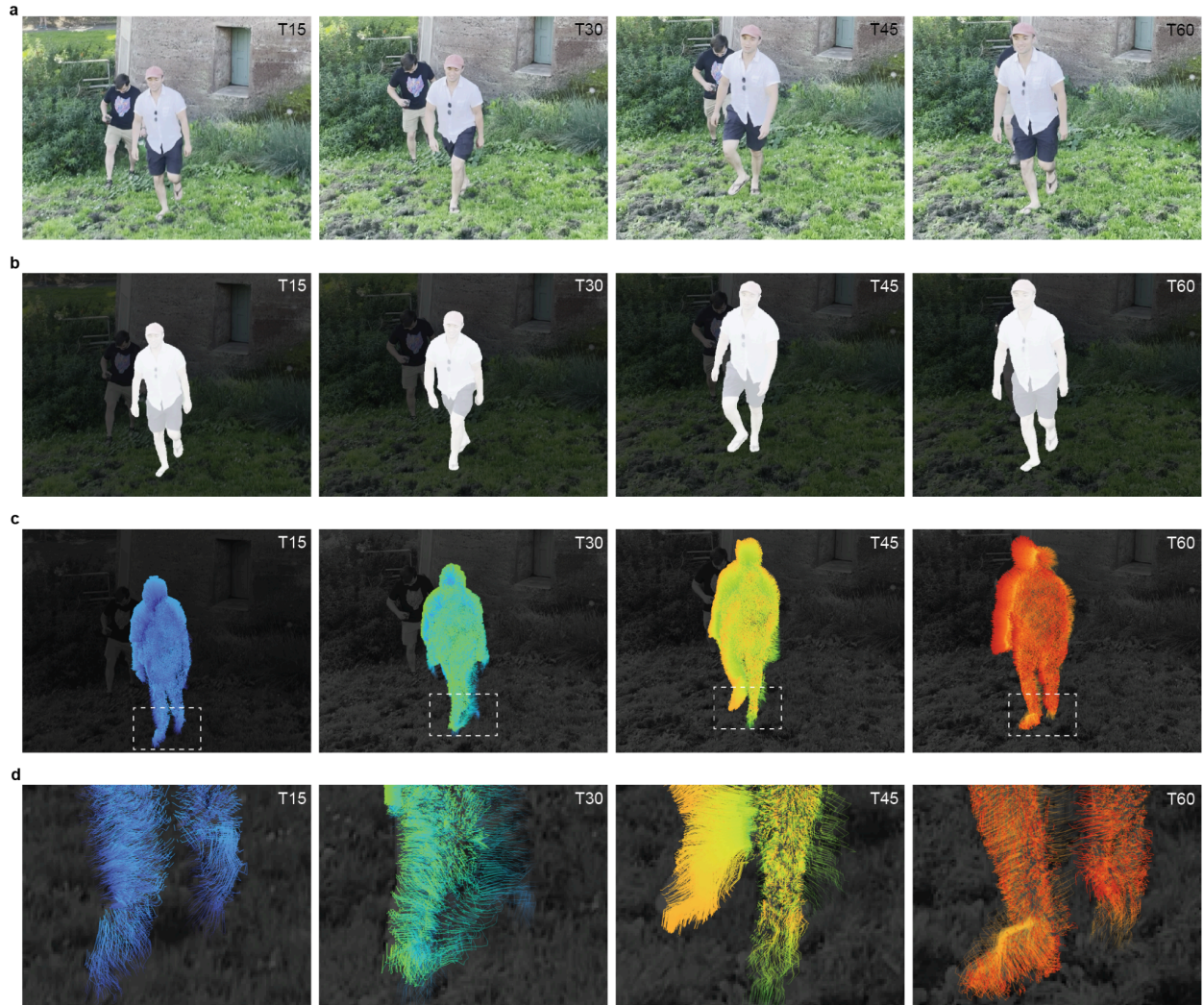

**Extended Data Fig. 6: Automated tracking and flow interpolation from iPhone camera video frames and corresponding semantic segmentation masks**
**a**, Four frames at T15, T30, T45, and T60 (left to right) of a 66 frame iPhone video taken at 29.98 fps using an iPhone 12 Pro, with a resolution of 1920x1080 pixels. **b**, The same corresponding frames as (**a**) after segmentation with Meta's Segment Anything model, and selection of semantic segmentations corresponding to the highlighted subject. **c**, The same corresponding frames as (**b**) after running the original video's frames and segmentation mask through Nellie's mocap marking, tracking, and flow interpolation pipelines, generating tracks for 5% of all pixels in the 3D (TYX) mask. **d**, A zoom in portion near a region of relatively abundant movement, the subject's feet, in the corresponding frames of (**c**). Tracks are colored by frame number, where T0 is dark blue and T66 is dark red based on a 'turbo' colormap, where each track is capped to a 5-frame tail.

**Code availability**

The Nellie pipeline and its Napari-based plugin is fully written in Python. The Python code and plugin are freely available online via Github at https://github.com/aelefebv/nellie. Supplemental materials for non pipeline-related code can be found via Github at https://github.com/aelefebv/nellie-supplemental. A template for creating Nellie plugins can be found via GitHub at https://github.com/aelefebv/nellie-plugin-example.

**Contributions**

A.E.Y.T.L. conceived of and designed Nellie and its pipeline, wrote the manuscript, created the figures, developed and wrote the code, performed the data analysis, and supervised the study. G.S. and K.H. provided experimental design ideas. A.E.Y.T.L. and G.S. alpha-tested the code, and performed cell-based maintenance and experiments, as well as microscopy experiments. T.Y.L., E.S., and M.P.L. provided cells and reagents. A.E.Y.T.L., G.S., K.H., and B.K.M provided helpful feedback on experimental design, data interpretation, and edited the manuscript. A.E.Y.T.L. and G.S. performed experiments for the revision of the manuscript.

**Acknowledgments**

The authors would like to thank Andrew G. York for helpful discussions on the pipeline, Alfred Millett-Sikking's advice and guidance on microscopy with the SOLS, and Evelia Salinas' insights on fluid dynamics and its applications to this pipeline. The authors also thank Brian Feng, Matthew Onsum, and Charlie Ledogar for reviewing the manuscript for its content, and Minna Kane and Colin Sanford for reviewing the manuscript for its wording and grammar. Thank you finally to Calico Life Sciences LLC for supporting this work.

**Ethics declarations**